%% file: main.tex
\documentclass[journal ]{new-aiaa}
\usepackage[utf8]{inputenc}
\usepackage{textcomp}

\usepackage{graphicx}
\usepackage[version=4]{mhchem}

\usepackage{bm, amsmath, siunitx, tabularx, tabularray, booktabs, subcaption, hyperref, cleveref, amsfonts, longtable, ragged2e}

\UseTblrLibrary{booktabs}
\usepackage[dvipsnames]{xcolor}
\usepackage[mode=buildnew]{standalone}
\usepackage[usestackEOL]{stackengine}
\hypersetup{hidelinks}

\title{Illumination-Robust Localization around Unknown Space Objects using Multispectral Image Fusion}

\author{Eric J. Elias \footnote{Draper Scholar, Department of Aeronautics and Astronautics, Massachusetts Institute of Technology, AIAA Student Member}}
\affil{Massachusetts Institute of Technology, Cambridge, MA, 02139, USA}
\author{Michael Esswein \footnote{Senior Member of Technical Staff, The Charles Stark Draper Laboratory Inc.}}
\affil{The Charles Stark Draper Laboratory Inc., Cambridge, MA 02139, USA}
\author{Jonathan P. How \footnote{Ford Professor of Engineering, Department of Aeronautics and Astronautics, Massachusetts Institute of Technology, AIAA Fellow} and David W. Miller \footnote{Professor Post Tenure, Department of Aeronautics and Astronautics, Massachusetts Institute of Technology, AIAA Fellow
\\ \\
\indent \indent The preliminary version of this work was presented at the 2026 AIAA SciTech Forum as manuscript AIAA 2026-2706~\cite{elias_navigation_2026}.
}}
\affil{Massachusetts Institute of Technology, Cambridge, MA 02139, USA}

\begin{document}

\maketitle

\begin{abstract}
As on-orbit operations grow, so does the need for precise navigation around unknown resident space objects (RSOs), such as spacecraft, debris, and asteroids. Simultaneous Localization and Mapping (SLAM) algorithms are often studied for mapping an RSO and estimating the inspector's relative pose, but conventional cameras struggle during eclipse, while lidar, though illumination-robust, has higher SWaP. Thermal-infrared cameras track targets through difficult lighting but lack the resolution and feature-richness of visible cameras. Images of a target satellite in low Earth orbit are simulated in both visible and thermal bands, and fused at the pixel level to combine each sensor's strengths for front-end tracking. Thermal imagery is further exploited by a novel silhouette-based visual place recognition (VPR) method that detects loop closures without training, runs entirely on CPU, and outperforms all baselines. Integrated SLAM performance is compared using visible, thermal-infrared, and fused imagery across lighting conditions, trajectories, and place recognition methods; using fused imagery for front-end tracking and our thermal-silhouette place recognition method performs best. The results show that visible and thermal modalities each provide unique information that is exploited differently at various stages of navigation. These low-SWaP sensors could ultimately enable reliable proximity operations on smaller, lower-cost spacecraft.
\end{abstract}





\input{1_intro}
\input{2_background}
\input{3_simulation}
\input{4_slam}
\input{5_loopClosure}
\input{6_results}
\input{7_conclusion}

\section*{Acknowledgments}
This work was supported by the Charles Stark Draper Laboratory, Inc. under the Draper Scholars Program. AI tools were used to improve clarity and grammar.

\bibliography{biblio}

\end{document}

%% file: 1_intro.tex
\section{Introduction}
\lettrine{O}{perating} in close proximity to resident space objects (RSOs)—such as asteroids, spacecraft, and orbital debris—is becoming increasingly common and essential. Close inspection and sample collection of asteroids may be required for scientific missions. Inspection of unknown spacecraft would support space situational awareness. Docking with the spacecraft may be needed for repair missions. Orbital debris may need to be intercepted and de-orbited. Given an RSO, whether natural or man-made, there is a wealth of information to gain from an inspection mission.

\begin{figure}[t]
\centering
\includegraphics[width=.6\textwidth]{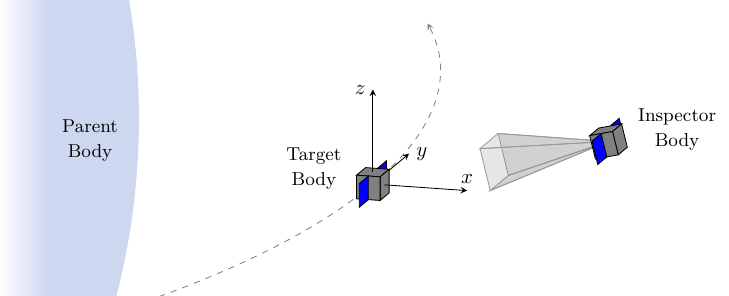}
\caption[An inspection mission around a parent body in the Hill frame]{A target and inspector orbiting around a parent body, with the inspector characterizing the target. The Hill coordinate frame is centered at the target, with the $x$~axis pointing radially outwards, the $y$~axis pointing in the target's direction of motion, and the $z$~axis pointing in the direction of the target's orbital angular momentum vector.}
\label{fig:hillFrame}
\end{figure}

The spacecraft performing the inspection (the inspector) requires precise knowledge of its position and attitude relative to the RSO of interest (the target). This navigation solution is used for avoiding collisions, predicting trajectories, planning maneuvers, characterizing the dynamics of the target, and creating a dense reconstruction of the target's shape and surface. This type of inspection scenario is illustrated in \autoref{fig:hillFrame}.

Target objects can be categorized into two classes: cooperative and uncooperative. Cooperative objects are designed to assist with the navigation of the inspector spacecraft in some way. As a result, they have navigation aids for inspectors to utilize, such as retroreflectors, and communicate their sensor measurements with the inspector.

This work focuses on uncooperative targets, which are not designed to interface with an inspector in any way. Uncooperative targets are considered ``known'' if a 3D model of the target's shape is available to the inspector prior to rendezvous. An inspector can estimate its position and attitude relative to the target (known as pose estimation) by matching 2D monocular images to the 3D model~\cite{sharma_comparative_2016,pauly_survey_2023} or by matching a 3D point cloud obtained from lidar to the 3D model~\cite{ruel_target_2008,opromolla_lidar_2014,liu_point_2016,woods_lidar_2016,perfetto_lidar_2019}. Cassinis et al. provide a general survey of these spacecraft pose estimation techniques~\cite{cassinis_review_2019}.

However, a target is not always ``known'' prior to the mission. These uncooperative and unknown targets are much more difficult to handle because a model of the target needs to be created while simultaneously being used to find the relative pose of the inspector. This class of algorithms is known as Simultaneous Localization and Mapping (SLAM) and is a prominent focus of research in robotics~\cite{slam_handbook_2026}. Lichter and Dubowsky~\cite{lichter_state_2004} first explored this problem in the context of space objects using collaborative 3D vision sensors (assumed to be uniformly distributed around the target), with a Kalman filter used to estimate the state, shape, and mass parameters of an unknown target. Sonnenburg et al.~\cite{sonnenburg_ekf_2010} used a set of stereo cameras to find the position of 3D features (high-contrast corners) on the target. These are inputs to an extended Kalman filter, which finds the relative pose of the inspector. A monocular camera can also be used to implement a similar system~\cite{augestein_improved_2011}. By tracking 2D features on the images of the target over multiple frames, the features can be triangulated and positioned in 3D space. This technique is known as structure from motion (SfM), and is particularly difficult in orbital domains due to the large operating ranges and slow relative motion as compared to ground-based SfM scenarios. Tweddle~\cite{tweddle_thesis_2013} implemented a SLAM algorithm assuming a stationary inspector that relies on a factor graph for mapping and pose estimation. Setterfield et al.~\cite{setterfield_mapping_2018} expanded on Tweddle's work so that it can handle a moving inspector. Orbital dynamics priors can also be incorporated into the SLAM estimate. Dor et al.~\cite{dor_astroslam_2024} formulated a SLAM system to navigate around an unknown small asteroid in orbit around the Sun using these priors. The effects of the Sun's gravity, the asteroid's gravity, and solar radiation pressure are used to improve the estimated trajectory of the inspector and the map of the asteroid. Learning-based methods have also been applied to this problem, where a single image is used to create a 3D reconstruction of the target spacecraft, which is then used to estimate the inspector's relative pose by matching 2D features on incoming images with 3D features on the 3D reconstruction~\cite{uwumukiza_dreamsatPose_2026}.

\begin{figure}[t]
\centering
\subcaptionbox{The Dragon spacecraft illuminated by the Sun. A large portion of the spacecraft is illuminated; however, the rest is in deep shadow and indistinguishable from the background of space.}[0.45\textwidth]{
    \centering
    \includegraphics[height=4cm]{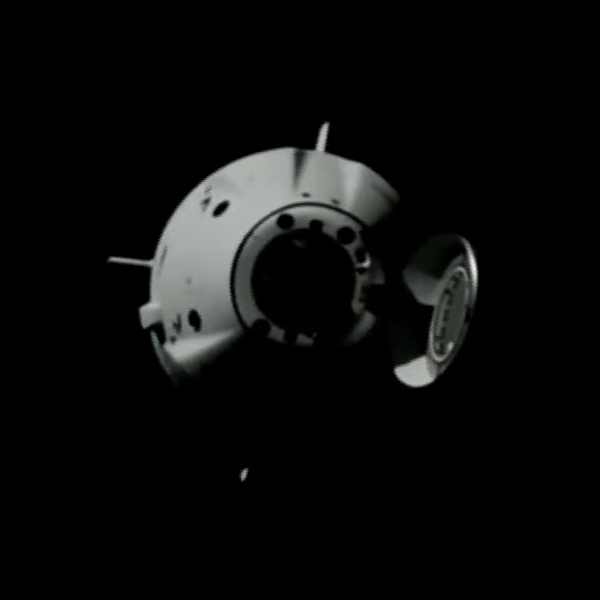}
}
\hspace{0.5cm}
\subcaptionbox{The Dragon spacecraft in eclipse. The only discernible parts of the spacecraft are the windows and the navigation lights, which are not present on most RSOs.}[0.45\textwidth]{
    \centering
    \includegraphics[height=4cm]{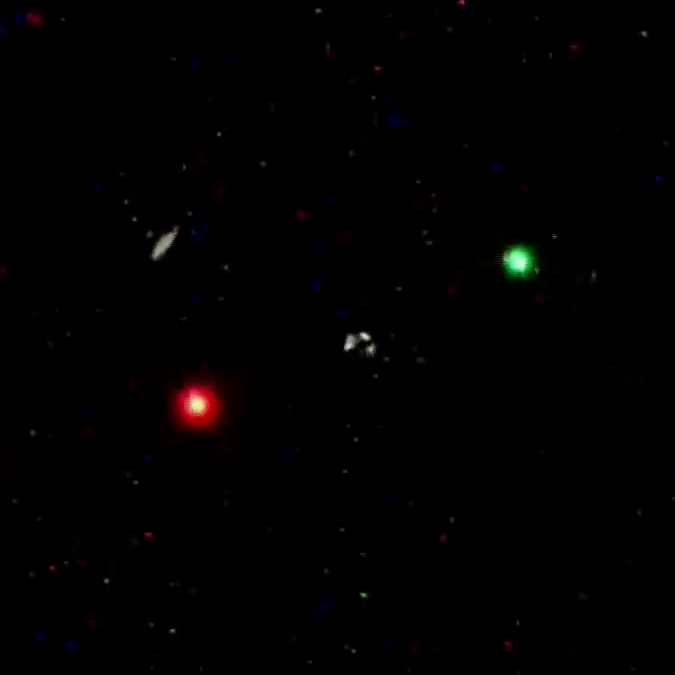}
}
\caption[Visible images of the SpaceX Dragon spacecraft during day and eclipse]{The SpaceX Dragon spacecraft seen during its approach to the ISS for the Crew-8 mission~\cite{ap_crew8_2024}. These two images show the visibility of the spacecraft during illumination by the Sun and during eclipse. These two images were taken approximately twenty minutes apart.}
\label{fig:dragon_iss}
\end{figure}

Illumination conditions are often extremely problematic for the inspection of an RSO. Depending on the alignment of the inspector, the target, and the Sun, the target may be fully illuminated, partially illuminated, or completely in shadow from the viewpoint of the inspector. More likely is the case in which the inspector sees a target with dynamic shadows that constantly move over the target's surface. Additionally, especially in low Earth orbit (LEO), it is common for the Sun to be completely eclipsed by Earth, as described in \autoref{section:eclipse}. An example of these shadowing and eclipse conditions from a real mission is shown in \autoref{fig:dragon_iss}. With dynamic shadows or full eclipse, a conventional camera is not able to track features on the target robustly.

Given these considerations, an additional sensor should be used by the inspector to improve navigation robustness. Lidar emits infrared lasers to measure the bearing and range of points on the target's surface. Unlike a camera, this active sensing method has the advantage of directly measuring the range to the target and is independent of illumination conditions. However, this also means that lidars use more power and are often significantly heavier and larger than cameras---constraints that are often difficult to accommodate in space vehicle design. Sufficiently powerful lidars also have the potential to damage fragile sensors on the target~\cite{bi_failure_2024}.

Thermal-infrared (TIR) cameras are robust to shadows and eclipses, while having a fraction of the footprint, weight, and power consumption compared to lidar. Instead of measuring visible light reflecting off of surfaces like visible (VIS) cameras do, TIR cameras detect thermal-infrared light being radiated or reflected from surfaces. They therefore measure the heat emitted by an object. For a given material, an object seen through a TIR camera is brighter when it is hot and dimmer when it is cold. In LEO, a spacecraft is heated from many sources, including internal heating, the Sun, and Earth. More details about thermal dynamics are discussed in \autoref{sec:thermal_dynamics}. Even in eclipse conditions, a spacecraft's internal heating and Earth's thermal radiation still allow the spacecraft to be visible from an inspector's TIR camera. Compared to VIS cameras, the advantages of TIR are countered by their low resolution, noise, and low feature-richness. For example, a surface covered with VIS features may appear completely featureless in the TIR band.

\begin{figure}[t]
\centering
\subcaptionbox{Range of 1.7 km while eclipsed from the Sun.}[0.4\textwidth]{
    \centering
    \includegraphics[height=4cm]{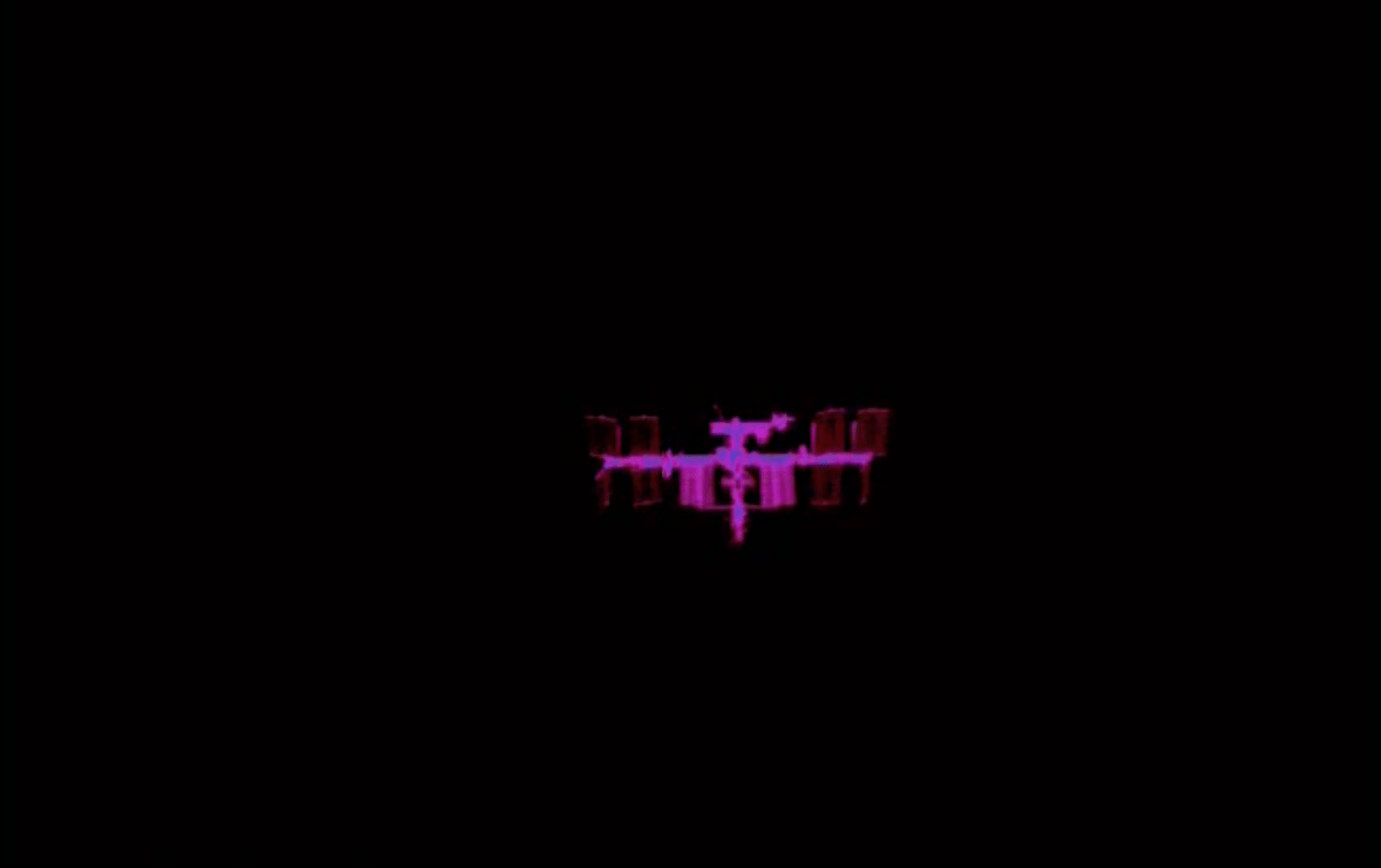}
}
\hspace{1cm}
\subcaptionbox{Range of 300 m while illuminated by the Sun.}[0.4\textwidth]{
    \centering
    \includegraphics[height=4cm]{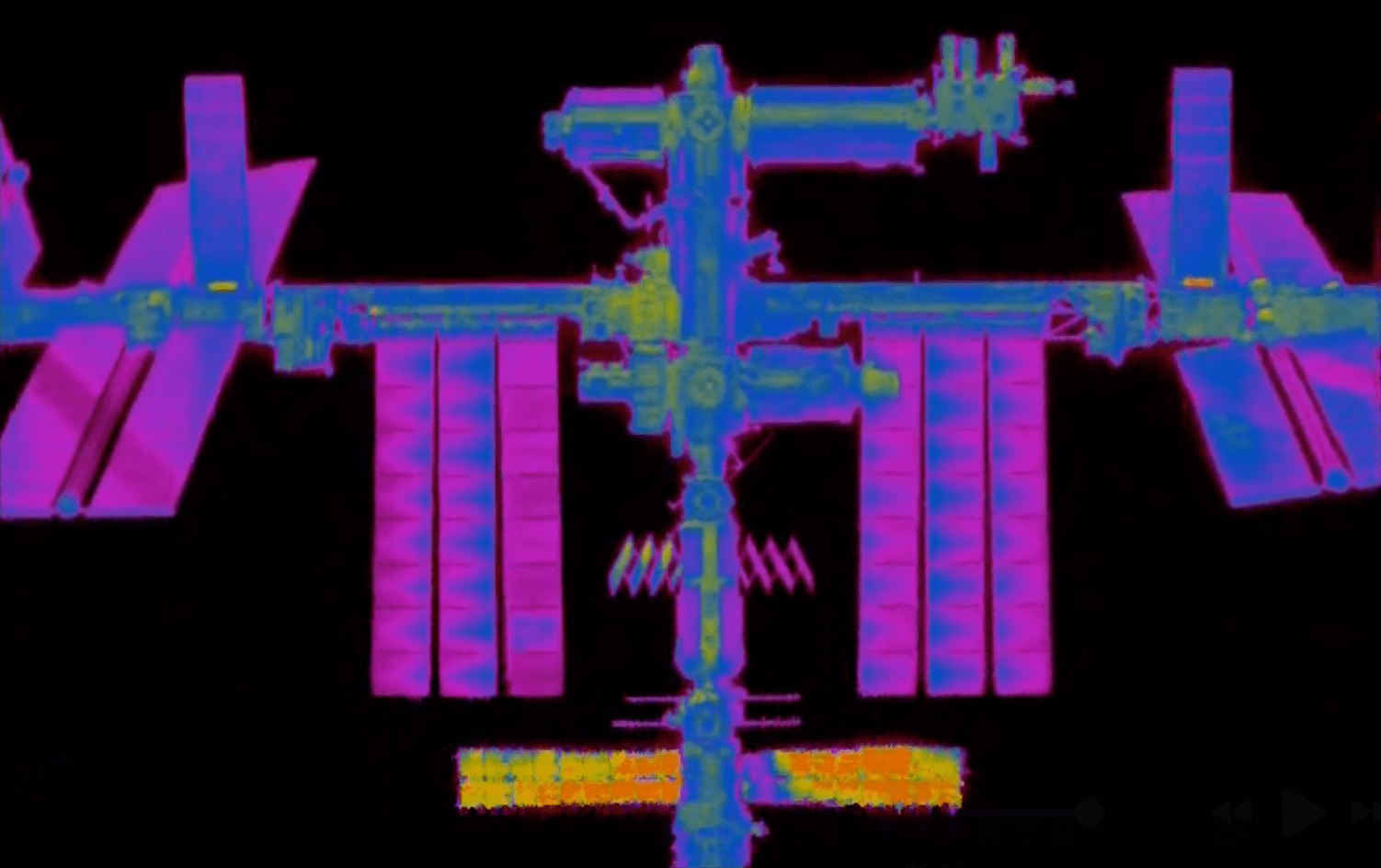}
}
\caption[Thermal images of the ISS during day and eclipse]{The ISS as seen from the TIR imager of the TriDAR vision system during rendezvous operations on STS-131~\cite{ruel_space_2012}.}
\label{fig:tir_iss}
\end{figure}

Imagery from TIR cameras has proven useful on flight demonstrations to the ISS~\cite{ruel_space_2012,cavrois_liris_2015}, as shown in \autoref{fig:tir_iss}. These missions demonstrate that, with a TIR imager, the ISS is clearly visible over the entire orbit, even when fully eclipsed. The use of TIR has been studied for relative navigation around known and unknown RSOs, with a focus on how to utilize information from both VIS and TIR cameras simultaneously. Combining the information from VIS and TIR images creates a solution that has the advantages of each while avoiding their disadvantages. Civardi et al.~\cite{civardi_small_2021} proposed a method to combine the VIS and TIR information to navigate around an unknown asteroid. SLAM is run separately for each sensor, and then the estimated relative pose from each SLAM solution is combined in an extended Kalman filter. While this method combines the estimation from the VIS and TIR sensors, it does so in a way that decouples the information between the two sensors. In other words, the features detected in the VIS image are not improving the navigation estimate obtained from the TIR image and vice versa. An approach that has been investigated for pose estimation around a known object~\cite{mcbryde_spacecraft_2018, civardi_generation_2024,civardi_uncooperative_2024} is the fusion of the VIS and TIR images at the pixel level before they are input into any navigation algorithm. Our approach uses the joint information in both sensors to create a navigation estimate, rather than fusing two decoupled pose estimates.

In addition to simultaneously using visible and thermal cameras for tracking features on the space object, we next focus on reducing the navigation errors even further. The core module of a SLAM system is visual-inertial odometry (VIO). This algorithm estimates the frame-to-frame motion of the camera, and integrates this motion to obtain a dense trajectory over time. This is often achieved by detecting and tracking high-contrast corners (features) on consecutive images, which then provides both the estimated trajectory of the camera and a map composed of the estimated location of each landmark (the feature in 3D space).

While modern VIO methods produce accurate frame-to-frame poses, any small errors in these estimates can integrate into significant errors over a long trajectory. Therefore, additional measurements must be included to minimize this drift. Each of these measurements relates two poses that are far apart in the trajectory, called a loop closure. A loop closure requires a visual place recognition (VPR) module, which detects when the current frame matches a similar perspective in the environment as that in a previous frame. After the relative pose between two matching frames is found, this constraint can be added to the estimator, minimizing the long-term drift in the trajectory estimate.

Existing VPR methods are designed with a focus on ground-based SLAM. When used to navigate around an RSO, however, they tend to perform poorly.  Illumination conditions are extreme and the target-pointing use case lies far outside of the terrestrial domain they were trained on. Additionally, most state-of-the-art VPR methods are designed to run on GPU, a platform that is typically not available onboard a spacecraft. We propose a CPU-only VPR method specifically for proximity navigation around an RSO, which performs better than all baselines, significantly reducing navigation errors.

This work investigates the use of VIS/TIR fused images compared to VIS-only or TIR-only images as inputs to a SLAM algorithm. To the best of our knowledge, this is the first time pixel-level VIS/TIR fusion has been used for navigation around a simulated unknown spacecraft in LEO. We also create a novel algorithm for identifying instances when a perspective of the unknown spacecraft is revisited. Specifically, the main contributions are:
\begin{itemize}
    \item Pixel-level fusion and image processing techniques are applied to create illumination-robust visible-thermal composite images of a photorealistic simulated spacecraft inspection.
    \item A novel visual place recognition method is developed that efficiently matches the thermal-infrared silhouette of a target to identify loop closure candidates. Place recognition performance is shown to be better than state-of-the-art baselines, with faster CPU-only runtime.
    \item Integrated navigation performance is demonstrated by feeding simulated inspection imagery into a monocular SLAM algorithm that estimates the inspector's trajectory without any prior knowledge of the target's shape. The results show that both front-end tracking and loop closure performance are significantly improved with the use of thermal imagery.
\end{itemize}
This paper incorporates and extends our previous work~\cite{elias_navigation_2026}, most notably adding the design of our silhouette-based VPR method as well as the analysis comparing it to baseline VPR methods. By including this new algorithm, we further improve navigation performance by incorporating thermal-infrared imagery into an additional part of the SLAM pipeline. This work also adds a scenario with a CubeSat spacecraft to provide a challenging test case for VPR methods because of its high degree of symmetry.

The remainder of this paper is organized into six sections. \Cref{sec:background} gives an overview of relevant background in SLAM, VPR, eclipse conditions, thermal dynamics, and VIS/TIR fusion. \Cref{sec:simulation_setup} discusses the details of the VIS and TIR simulator that is created. \Cref{sec:slam_around_RSO} reviews the setup and integration of this simulated imagery with a SLAM algorithm. \Cref{sec:silhouette_loop_closure} discusses a novel method to improve loop closure detection based on the matching of RSO silhouettes and shows its effect on navigation performance. \Cref{sec:results} compares the performance of our place recognition algorithm to baseline methods, followed by the analysis integrated SLAM navigation performance between VIS-only, TIR-only, and VIS/TIR fused methods, as well as between different place recognition techniques. \Cref{sec:conclusion} concludes the paper.

%% file: 2_background.tex
\section{Background} \label{sec:background}

\subsection{Kimera}
In this work, Kimera~\cite{kimera,kimera2} is used as the SLAM algorithm of choice. Kimera is a feature-based SLAM framework that distills priors, measurements, and dynamics into a factor graph to jointly optimize both the inspector's relative motion and the 3D positions of detected feature points on the target. In monocular mode, Kimera requires only monocular camera images and IMU measurements. We use two modules from Kimera: Kimera-VIO and Kimera-RPGO, implementing Visual-Inertial Odometry and Robust Pose Graph Optimization, respectively.

Kimera-VIO detects and tracks high-contrast points in the image (features) to estimate the motion of the inspector. This information is combined with the IMU measurements in a factor graph to estimate the states of the inspector and the 3D positions of each feature point (turning the 2D features on the image into 3D landmarks) over the entire smoothing window. Kimera uses a fixed-lag smoother in this case, meaning inspector states and landmark positions are repeatedly optimized within a recent time horizon. This visual-inertial odometry effectively adds together many visual and inertial measurements, each consisting of small changes in relative pose of the inspector. As a result, the error in the state grows without bound.

Kimera-RPGO bounds this error using loop closures, which occur when the inspector re-visits the same view of the target. When a loop closure is detected, the relative pose between the current frame and the corresponding previous frame is found. Kimera-RPGO uses a separate factor graph from Kimera-VIO to set up the estimated full trajectory of the inspector. Every time a loop closure is detected, a new constraint is added to the factor graph and the full inspector trajectory is re-optimized.
It is important to minimize the number of false loop closures, as even a single incorrect measurement can completely corrupt the entire estimated trajectory. A specific problem to avoid, known as perceptual aliasing, is when two places in the environment look very similar even though they are at very different locations. As a result, many verification and outlier rejection steps are often included in SLAM pipelines to throw out these false measurements. The Kimera pipeline is summarized in the following list of steps:
\bigskip
\begin{description}
    \item[Kimera-VIO:] Creates a visual-inertial odometry estimate of the inspector's motion.
        \begin{enumerate}
            \item Detect new features in the current frame and track them from the detected features in the previous frame.
            \item Preintegrate IMU measurements between the last frame and the current frame.
            \item Use visual and inertial measurements to jointly optimize for inspector states and landmark positions that fall within a recent time horizon (bundle adjustment).
        \end{enumerate}    
    \item[Kimera-RPGO:] Uses information from Kimera-VIO along with loop closure detections to create a Robust Pose Graph Optimization (RPGO) estimate.
        \begin{enumerate}
            \item At each frame, check for matching frames using VPR. If no match is detected, finish processing.
            \item Find the relative pose between the current frame and the matching previous frame (also known as \textit{pose recovery} or \textit{geometric verification}) by first detecting features in each image and then using RANSAC~\cite{fischler_ransac_1981} to find inlier matches that solve for a relative pose.
            \item Add the loop closure constraint to the current trajectory estimate (which may already include previous loop closure constraints).
            \item Re-optimize the trajectory estimate, removing outliers that are inconsistent with the VIO estimate using Graduated Non-Convexity (GNC)~\cite{yang_graduated_2020}.
        \end{enumerate}
\end{description}

\subsection{Visual Place Recognition (VPR)}
Loop closures are extremely important in SLAM, as they correct for any drift in the trajectory estimate. The first and most difficult step in loop closure detection is place recognition. When each query frame is read, it must be compared to potentially tens of thousands of previous frames in real time. This comparison checks if the frames have similar views of the environment. VPR algorithms fall into three classes: (1) ones that extract notable patches within each frame (local feature methods), (2) ones that describe the entire frame (global methods), and (3) ones that use a pretrained neural network to compute descriptors (learning-based methods).

Local feature methods work by extracting features from a frame and computing descriptors (long vectors of numbers) based on a small patch of pixels around the feature. Matching scenes are found by identifying frames with many matching local features (descriptors). Feature detectors and descriptors such as SIFT~\cite{lowe_sift_2004}, SURF~\cite{bay_surf_2008}, BRIEF~\cite{colander_brief_2010}, ORB~\cite{rublee_orb_2011}, and BRISK~\cite{leutenegger_brisk_2011} are commonly used. However, each image may contain hundreds of local features, so directly comparing frames is highly inefficient. Inspired by text retrieval methods~\cite{baeza_modern_1999}, an efficient image retrieval algorithm, Bag of Words~\cite{sivic_video_2003}, was developed. This method has often been integrated into SLAM systems to enable reliable loop closures~\cite{lopez_bags_2012}.
The Bag of Words method finds all the feature descriptors in each image. These descriptors are then classified using a vocabulary into ``visual words''. The vocabulary is a classification method trained offline using a large representative dataset, which may contain tens of thousands or even hundreds of thousands of words. The distribution of visual words in each image can then be assembled into a histogram. Now that each image can be encoded into a histogram, an efficient comparison can be made between them, especially with the use of vocabulary trees~\cite{nister_scalable_2006}. In practice, the most recent frame from the camera can be used to query a database of all previous frames using this method. The closest image match, if the descriptor distance is below a threshold, becomes a loop closure candidate.
In contrast, global methods directly compute a descriptor based on the edges, corners, and/or color within an entire image. While global methods are less sensitive to lighting changes than local feature methods, they are more sensitive to camera viewpoint changes~\cite{lowry_visual_2015}. An example of a popular global descriptor method is Gist~\cite{oliva_gist_2006}.

The third class of VPR methods is learning-based methods. These methods use neural networks trained with very large numbers of images to ``learn'' image descriptors such that similar places have descriptors that are close together and dissimilar places have descriptors that are far apart.
One of the first learning-based VPR methods, NetVLAD~\cite{arandjelovic_netvlad_2016}, uses a convolutional neural network (CNN) with a novel aggregation layer to generate the image descriptors. The CNN was trained using labeled data aggregated from Google Street View Time Machine. This method was shown to be far more robust to lighting and viewpoint changes than classical techniques.
CosPlace~\cite{Berton_CosPlace_2022} also uses a CNN, but instead treats the VPR problem as a classification problem, where images are sorted into classes, each representing part of the geographical area of interest. This method shows faster training and stronger performance than NetVLAD. MixVPR~\cite{ali_mixvpr_2023} proposed a feature mixer on the output of the CNN, which is able to preserve the spatial relationships between features in an image. EigenPlaces~\cite{Berton_EigenPlaces_2023} created a model robust to viewpoint changes by training with many viewpoints of the same environment. AnyLoc~\cite{keetha_anyloc_2023} pivoted away from CNNs and instead used a vision transformer foundation model as its backbone. MegaLoc~\cite{Berton_MegaLoc_2025} built on this by combining landmark retrieval, visual place recognition, and visual localization tasks into one unified model.

\subsection{Illumination in Orbit}  \label{section:eclipse}
\begin{figure}[btp]
\centering
\subcaptionbox{$\beta$ is defined as the angle between the solar vector $\vec{s}$ and its projection onto the orbital plane.\label{fig:beta}}[0.45\textwidth]{
    \centering
    \includegraphics[width=.45\textwidth]{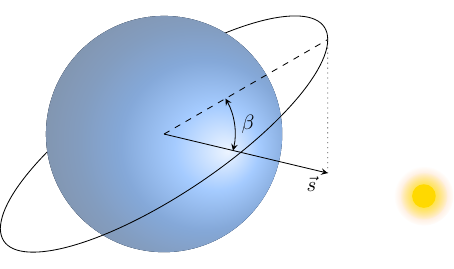}
}
\hfill
\subcaptionbox{The fraction of a 400 km circular Earth orbit spent in eclipse and sunlight, depending on $\beta$.\label{fig:betaVsEclipse}}[0.45\textwidth]{
    \centering
    \includegraphics[width=.45\textwidth]{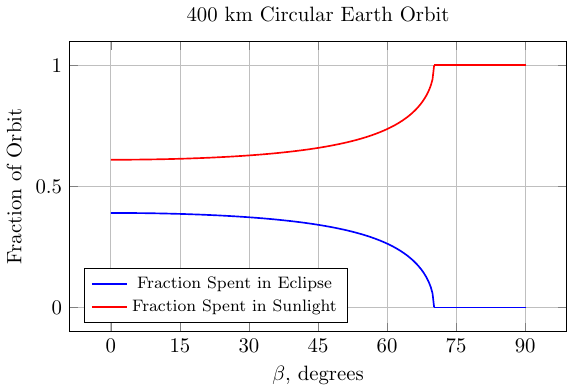}
}
\caption{Visualization of the $\beta$ angle and its effect on illumination.}
\end{figure}

While cameras are a commonly-used sensor for SLAM systems, a significant issue with visible cameras is their inability to see parts of the target that are not illuminated by the Sun. In an on-orbit scenario, this occurs when the target is in between the inspector and the Sun, such that the target is backlit. However, a far more significant and common scenario is when Earth is eclipsing the target. To get a sense of how significant this effect is, we define the beta angle, $\beta$, as the angle between the solar vector and its projection onto the orbital plane~\cite{gilmore_spacecraft_2002}. This relationship is shown in \autoref{fig:beta}.

When $\beta$ is small, the amount of time spent in eclipse is maximized. When $\beta$ is at its maximum value of \ang{90}, the orbital plane is normal to the solar vector, so the target is always exposed to the Sun. $\beta$ varies over time as an orbit precesses due to non-spherical gravity, as well as due to the solar vector changing as the Earth orbits the Sun. An example of how $\beta$ affects the time spent in sunlight can be seen in \autoref{fig:betaVsEclipse}. In LEO, it is common for objects to spend a significant amount of time in eclipse, sometimes exceeding 40\% of the orbital period.

\subsection{Spacecraft Thermal Dynamics} \label{sec:thermal_dynamics}
In order to overcome the illumination limitations of visible imagery, we combine it with thermal imagery. Before simulating a spacecraft in the thermal band, however, it is important to first understand the sources of thermal radiation in LEO~\cite{gilmore_spacecraft_2002}. The sources we consider are internal heating, direct solar, Earth albedo, and Earth IR. 

\paragraph{Internal Heating.} A spacecraft has many sources of heat generated internally. These sources may include computers, communication systems, actuators, and heaters. The heat from these components conducts throughout the spacecraft and reaches the surface, where it is radiated away. 
\paragraph{Direct Solar.} The most significant source of spacecraft heating is by far direct sunlight. As long as the Sun is in view, it provides a source of radiation to the spacecraft with a constant intensity.
\paragraph{Earth Albedo.} Sunlight reflects off Earth and back onto the spacecraft. This albedo is usually higher over land than ocean, and also has some dependence on latitude. For our simulation, we assume a constant albedo over Earth's surface.
\paragraph{Earth IR.} Sunlight is absorbed by Earth and then radiated back into space as IR emission. Similar to Earth albedo, the intensity of this radiation is dependent on the location above Earth's surface. Again, we assume a constant Earth IR intensity over the surface.

%% file: 3_simulation.tex
\section{Visible/Thermal-Infrared Simulation} \label{sec:simulation_setup}

\subsection{Dynamics Propagation}
MATLAB's high-fidelity numerical propagator is used to propagate the orbits of the inspector and the target, taking into account high-order gravity and atmospheric drag. For each simulated scenario, the orbital parameters of the inspector are adjusted to achieve the desired relative trajectory of a natural motion circumnavigation.

\subsection{Rendering}
\begin{figure}[t]
\centering
\includegraphics[width=.8\textwidth]{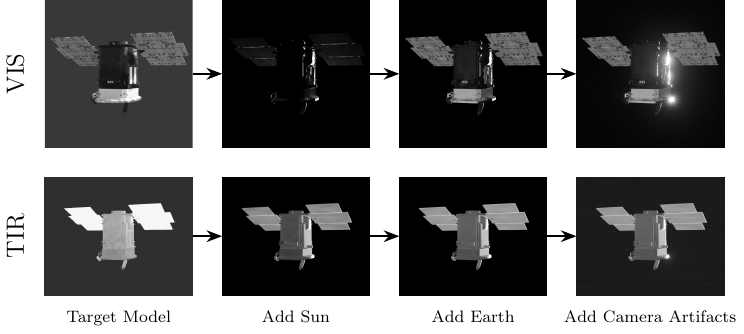}
\caption{The rendering framework, showing the effects of each simulation component on the CloudSat model. Camera artifacts include lens glare, bloom, and noise.}
\label{fig:renderingPipeline}
\end{figure}

Blender is used as the rendering software of choice because of its photorealistic rendering performance. The propagated dynamics of the target and inspector from MATLAB are imported into Blender. The target and a LEO environment are simulated to be as realistic as reasonably possible in both the VIS and TIR bands. This process starts with the creation of a target spacecraft, after which an environment is built around it. Since the simulation is in LEO, the environment includes Earth and the Sun. The VIS and TIR cameras also have unique properties that need to be included. These steps in the rendering pipeline are illustrated in \autoref{fig:renderingPipeline}. It is important to note that every element in the rendering pipeline is adjusted such that the rendered images match real reference images in both VIS and TIR bands as closely as possible.

\subsection{Target Spacecraft}
3D models of CloudSat and a 1U CubeSat from NASA's 3D model repository are imported into Blender as the target vehicles. These models include the mesh of each satellite, as well as the relevant colors, textures, and material properties. For the VIS renderings, the imported 3D models remain mostly unchanged. However, significant changes are made to duplicated 3D models to simulate the appearance of the spacecraft in the TIR band.

To emulate internal heating of the spacecraft, a randomized emission texture is applied over the surface of each model. This causes some parts of the spacecraft to be slightly brighter (warmer) and others to be slightly dimmer (cooler). The simulation of the surface heating due to external sources is achieved with a grey-colored texture, so when the thermal emission from the Sun and Earth is stronger, the surface is brighter and when it is weaker, the surface is dimmer. However, this ignores the conduction of heat towards the unlit sides of the spacecraft, as well as the storage of heat, especially after orbital sunset. To address this, the emission of each part of the spacecraft is separately hand-animated to reproduce the thermal conduction behavior seen on orbit. Larger and more thermally massive components such as the spacecraft bus are animated to experience slow temperature changes, while low-mass components such as the solar array are animated with fast temperature changes, particularly during transitions between day and night. As the CubeSat is significantly smaller than CloudSat, it is animated with faster temperature changes.

\subsection{Environment}
In the VIS simulation, the two significant illumination sources are the Sun and Earth albedo. The Sun is simulated as an inertially fixed point source infinitely far away. Earth is simulated photorealistically as a sphere, with land, water, and cloud textures applied from satellite imagery. Atmospheric scattering is also included in this simulation. The rendering of Earth has a significant effect on the lighting of the target spacecraft, often filling in the harsh shadows left by the Sun. This work does not study the effect of Earth in the background on tracking performance, so the direct visibility of Earth is excluded from the renderings. However, it has been shown that---compared to VIS imagery---the use of TIR imagery is far more robust to the clutter of Earth in the background~\cite{fiengo_multispectral_2019}.

\begin{figure}[t]
\centering
\subcaptionbox{VIS}[0.23\textwidth]{
    \centering
    \includegraphics[width=0.23\textwidth]{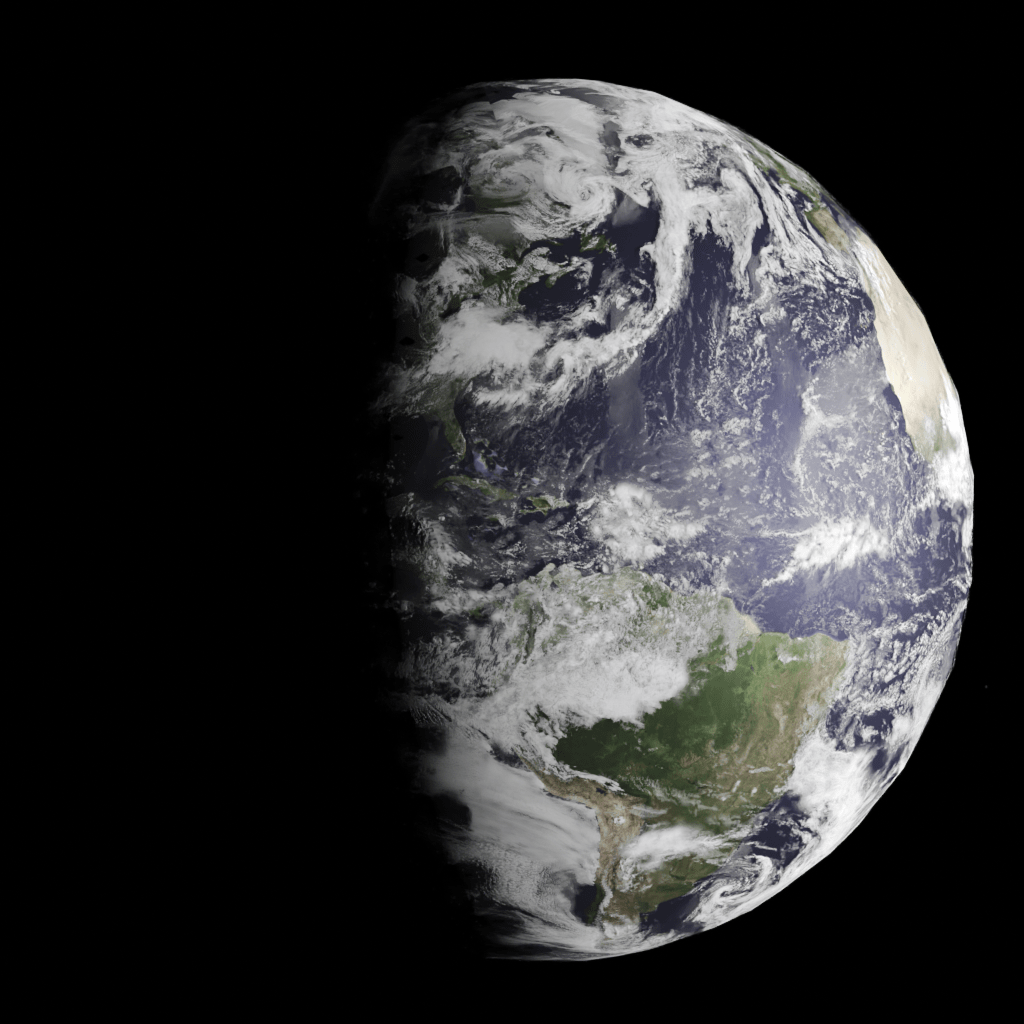}
}
\hspace{1cm}
\subcaptionbox{TIR}[0.23\textwidth]{
    \centering
    \includegraphics[width=0.23\textwidth]{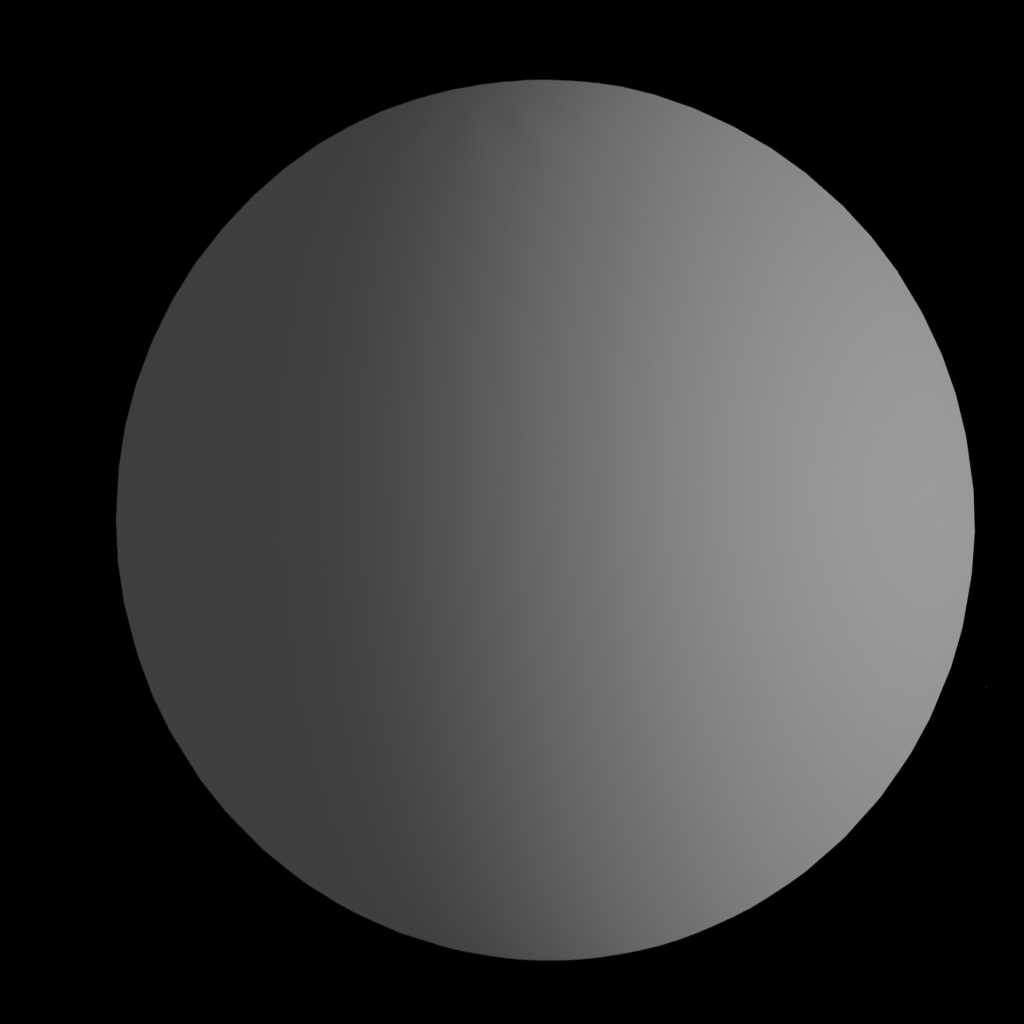}
}
\caption{Earth as simulated using Blender in the VIS and TIR bands.}
\label{fig:blenderEnvironment}
\end{figure}

The TIR simulation replaces Earth with a grey sphere. The grey color emulates the Earth albedo, such that the incident solar light is reflected. An emission property is also added, causing the Earth model to radiate light, emulating TIR radiation from Earth's surface and atmosphere. The Sun is modeled as a large, inertially fixed light source in order to avoid any harsh shadowing on the target spacecraft in the TIR. The simulation of Earth---including Earth albedo and Earth IR---is shown in \autoref{fig:blenderEnvironment}.

\subsection{Camera}
For each type of imager, commonly used camera resolutions are chosen with full specifications laid out in \autoref{tab:cam_comparison}. Narrower camera fields of view are used in some scenarios to ensure the target is large enough in the cameras' frames for effective tracking.

\begin{table}[t]
\centering
\caption{VIS and TIR camera properties used in the simulations. The field of view is changed depending on the scenario to keep the target large in the frame.}
\begin{tabular}{lcc}
\toprule
 & VIS Camera & TIR Camera \\
\midrule
Resolution, pixels & $1024 \times 1024$ & $640 \times 512$ \\
FoV (S1-S4) & $25^{\circ} \times 25^{\circ}$ & $31.25^{\circ} \times 25^{\circ}$ \\
FoV (S5) & $8^{\circ} \times 8^{\circ}$ & $10^{\circ} \times 8^{\circ}$ \\
FoV (S6) & $2^{\circ} \times 2^{\circ}$ & $2.5^{\circ} \times 2^{\circ}$ \\
\bottomrule
\end{tabular}
\label{tab:cam_comparison}
\end{table}

Lens glare and bloom are added to the rendered images to replicate the artifacts seen through real cameras. These effects are often seen in the VIS camera due to reflections of the Sun off the target, which cause a large part of the image to be overexposed. The same glare and bloom effects are applied to the TIR images; however, the limited strength of specular reflections causes these effects to be minimal.

Another significant camera artifact is noise. A small amount of Gaussian noise is added to the VIS camera to simulate the electronic noise inherent to the camera. Most TIR cameras use an uncooled microbolometer imager which, unlike other types of TIR imagers, does not require a cryogenic cooler. However, the lack of cooling creates significant noise in the form of fixed-pattern noise (FPN) consisting of vertical and horizontal streaks across the image~\cite{Barral_FPN_2024}. This FPN is added to the rendered TIR images, along with standard Gaussian noise. Because few TIR reference images are available, the fidelity of our TIR simulation is uncertain relative to the VIS imagery. To verify that the observed trends in our results are insensitive to this uncertainty, we conduct a sweep of the TIR noise parameters, confirming that each set of results matches the trends seen in~\autoref{subsec:integrated_performance}.

\subsection{Scenarios}
Across all scenarios, the orbital parameters for the target remain the same. A circular orbit with a semi-major axis of 6,778 km and an inclination of 51.6\textdegree\ is used, representing an altitude of approximately 400 km. No translational maneuvers are simulated in any case. The Sun is placed such that $\beta$ is close to zero, creating a worst-case scenario for eclipse conditions where approximately 40\% of each scenario has the Sun eclipsed by Earth.

The attitude of the inspector is set such that the camera boresight is always pointed at the center of the target, and that the target's vertical axis is aligned vertically in the inspector's image frame.

Each scenario is propagated for three orbital periods to simulate an inspection mission. A single 400 km orbit has a period of 1.5 hours, so each scenario has a duration of 4.5 hours. Six scenarios are simulated, each representing unique trajectory, illumination, and attitude configurations. The details of each scenario are found in \autoref{tab:scenarios}.

\newcolumntype{Z}[1]{>{\hsize=#1\hsize\linewidth=\hsize\RaggedRight\arraybackslash}X}

\begin{table}[t]
\footnotesize
\centering
\caption{The six simulated scenarios used to render visible and thermal-infrared images.}
\begin{tabularx}{\textwidth}{@{} Z{0.65} Z{0.2} Z{0.9} Z{1.25} @{}}
\toprule
Scenario & Range & Illumination & Target Attitude \\
\midrule
S1: CloudSat 15 m & 15 m & Nominal illumination with eclipse & Inertially fixed \\
S2: CloudSat 15 m, Fully Lit & 15 m & Perfectly illuminated from the inspector & Inertially fixed \\
S3: CloudSat 15 m, Rotating & 15 m & Nominal illumination with eclipse & Spins opposite the inspector's motion about the axis parallel to Earth's rotation axis at \qty{5e-4}{rad/s} \\
S4: CloudSat 15--25 m & 15--25 m & Nominal illumination with eclipse & Inertially fixed \\
S5: CloudSat 50 m & 50 m & Nominal illumination with eclipse & Inertially fixed \\
S6: CubeSat 15 m, Rotating & 15 m & Nominal illumination with eclipse & Spins opposite the inspector's motion about the axis parallel to Earth rotation axis at \qty{5e-4}{rad/s} \\
\bottomrule
\end{tabularx}
\label{tab:scenarios}
\end{table}

\begin{figure}[tp]
\centering
\subcaptionbox{15 m Range}[0.3\textwidth]{
    \centering
    \includegraphics[width=0.3\textwidth]{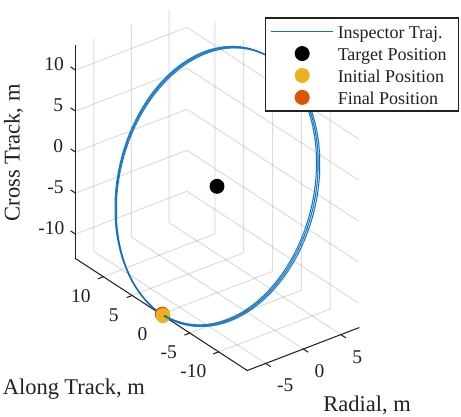}
}
\hfill
\subcaptionbox{15--25 m Range}[0.3\textwidth]{
    \centering
    \includegraphics[width=0.3\textwidth]{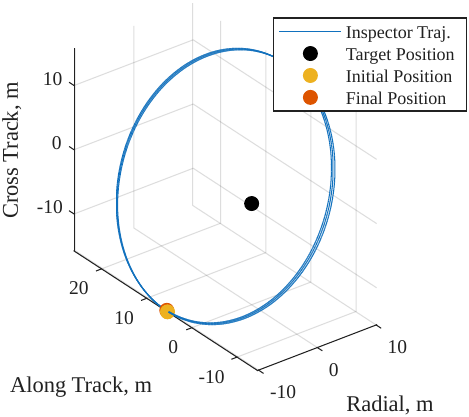}
}
\hfill
\subcaptionbox{50 m Range}[0.3\textwidth]{
    \centering
    \includegraphics[width=0.3\textwidth]{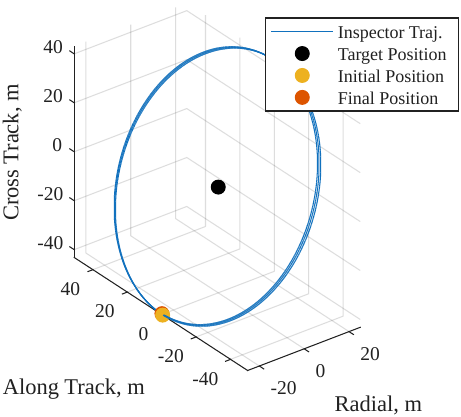}
}
\caption{Three unique relative trajectories are generated to create the scenarios. All trajectories are shown in the Hill frame over three orbits.}
\label{fig:scenarios}
%
\centering
\includegraphics[width=0.4\textwidth]{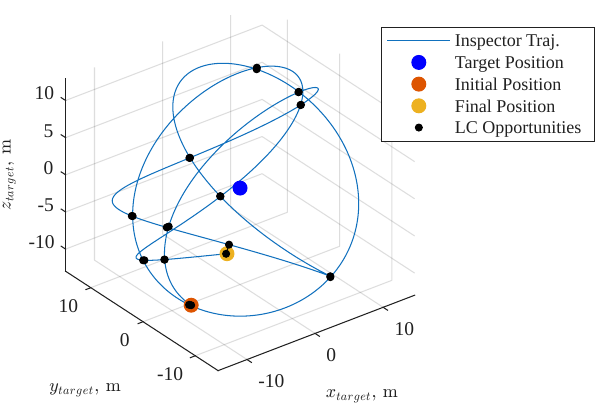}
\caption{The relative inspector trajectory for Scenarios 3 and 6 in the target body frame with the 12 loop closure opportunities shown.}
\label{fig:rotScenario}
\end{figure}

Out of these six scenarios, three unique relative inspector trajectories are used, as shown in \autoref{fig:scenarios}. Scenarios 1, 2, and 3 all follow the same trajectory in the Hill frame, repeated three times. However, unlike Scenario 1, Scenario 2 has the target perfectly illuminated from the inspector’s position, similar to placing a large light source on the inspector. Comparing Scenarios 1 and 2 isolates how eclipse and shadowing affect navigation performance. The TIR renderings are exactly the same between these two cases.

Scenario 3 is the same as Scenario 1, except that a constant spin is applied to the target. The spin creates constantly moving shadows across the surface of the target. Kimera assumes a static environment, so for Scenario 3, Kimera must perform estimation in the body frame of the target instead of the Hill frame. This trajectory becomes more complex in the fixed body frame of the target. While the trajectory in the Hill frame is the same as in Scenarios 1 and 2, the target is rotating under the trajectory, causing the inspector to see a unique view of the target at almost all points in this scenario. As a result, there are only 12 self-intersections of the trajectory in the target body frame over the entire 4.5-hour scenario. The trajectory and each of the 12 loop closure opportunities are shown in \autoref{fig:rotScenario}. At each of these self-intersections, the inspector sees the same view of the target. Compared to the scenarios with an inertially fixed target attitude, where the imagery in the second and third circumnavigations is virtually identical to that in the first, this scenario is far more dependent on the performance of the VPR method, as missing these limited loop closure opportunities has a significant effect on localization performance.

Scenarios 4 and 5 have the same illumination and target attitude conditions as Scenario 1. However, in Scenario 4, the inspector-target range varies between 15 and 25 m, while Scenario 5 has a constant range of 50 m. Scenario 6 is identical to Scenario 3, except that the CloudSat model is replaced with the CubeSat. The CubeSat is highly symmetric, providing a particularly difficult test case for VPR algorithms.

\subsection{Relative Dynamics Measurements}  \label{section:rel_dynamics}
Kimera requires translational acceleration and rotational velocity measurements in the sensor/body frame, relative to an inertially fixed environment. These measurements are combined with visual data to create the estimated state of the inspector. In most terrestrial SLAM use cases, these measurements come from an accelerometer and gyroscope mounted alongside the camera.

The dynamics of an on-orbit inspector behave very differently from the dynamics of a ground robot or UAV, however. Contrary to Kimera's assumptions, the environment being mapped (the target RSO in this case) may not have an attitude that is inertially fixed if it has some rotational velocity. Similarly, the target's inertial position is certainly not fixed as it is orbiting Earth. As a result, the SLAM formulation needs to be adjusted to estimate the rotational velocity of the target and to incorporate the relative orbital dynamics between the target and inspector. Dor et al. formulate a SLAM framework that incorporates these dynamics~\cite{dor_astroslam_2024}.

In order to simplify the problems arising from a non-inertially-fixed target to instead focus on the problem of visible-thermal fusion, this work assumes we have some noisy translational acceleration and rotational velocity measurements relative to the body frame of the target. Rotational velocity measurements are simulated using the ground truth relative trajectory with added noise, random walk, and bias instability to match measurement errors seen from a gyroscope. 

Relative acceleration measurements are also simulated using the ground-truth relative trajectory. In practice, these estimates could be obtained from relative orbital dynamics models~\cite{dor_astroslam_2024}. Analytical propagation models can exhibit position errors on the order of hundreds of meters over a 24-hour period~\cite{sullivan_comprehensive_2017}. Rather than explicitly incorporating an orbital dynamics model into Kimera, we add errors to the simulated relative acceleration measurements, calibrated to produce position drift of a comparable magnitude when integrated twice. This added error also factors in any uncertainty about the target's spin or maneuvering.

\begin{table}[t]
\centering
\caption{Error parameters used for the generation of simulated relative acceleration and relative angular velocity measurements. These are generated at a rate of 10 Hz.}
\begin{tabular}{@{}l c l@{}}
\toprule
 & \multicolumn{1}{c}{\shortstack{Relative \\ Acceleration}}
 & \multicolumn{1}{c}{\shortstack{Relative \\ Angular Velocity}} \\
\midrule
Noise Density       & 0                    & \qty{1e-6}{rad/s/\sqrt{Hz}} \\
Bias Instability    & \qty{1e-5}{m/s^2}    & \qty{1e-6}{rad/s} \\
Random Walk         & 0                    & \qty{1e-6}{rad/s \cdot \sqrt{Hz}} \\
\bottomrule
\end{tabular}
\label{tab:msmtParams}
\end{table}

%% file: 4_slam.tex
\section{Visible-Thermal SLAM} \label{sec:slam_around_RSO}

\subsection{Image Processing \& Fusion}
After images are generated in Blender, preprocessing is applied to increase contrast and reduce noise in the images before they are fed to Kimera. For each VIS and TIR frame, a threshold mask is used to separate the target from the dark background to minimize amplification of noise. Contrast Limited Adaptive Histogram Equalization (CLAHE)~\cite{zuiderveld_clahe_1994} is then applied to each VIS and TIR frame to improve local contrasts over the image and enhance the definition of edges; this is a common technique used to improve feature detection performance.

The input VIS and TIR images are often not the same resolution, with TIR usually a substantially lower resolution than VIS. However, the two images need to have matching resolution for the fusion methods to combine them. Civardi et al.~\cite{civardi_generation_2024} show that upscaling the TIR images to match the resolution of the VIS creates a fused image that is less sensitive to noise in comparison to downscaling the VIS images to match the TIR resolution. This approach is not only preferred because it is more robust to noise, but also because the higher resolution of the fused image allows for more precise placement of feature points, creating a more accurate navigation solution.

Following this approach, we upscale the TIR images using bicubic interpolation and add padding to match the VIS resolution. The TIR image is then registered to the VIS image using a transformation matrix that can be pre-computed during imager calibration. After contrast enhancement, upscaling, and registration, the VIS and TIR images are fused. 

Ref.~\cite{civardi_generation_2024} presents an analysis of different image processing and fusion techniques to combine two separate VIS and TIR images into a fused image in the context of uncooperative spacecraft. Their analysis compares how closely the fused images retain the information from the input VIS and TIR images in the presence of noise, as well as the runtime of each algorithm. They find that two image fusion methods create the best resultant composites with a fast runtime (< 60 ms): Anisotropic Diffusion-Based Fusion (ADF)~\cite{bavirisetti_fusion_2016} and Two-scale Image Fusion using Saliency Detection (TSIFSD)~\cite{bavirisetti_twoscale_2016}. For this study, we choose TSIFSD over ADF due to its faster runtime, and use it in the context of SLAM around an unknown spacecraft. An example of the renders and fusion results is shown in \autoref{fig:renders}.

\input{imgSeqTable}

\subsection{Kimera Setup}
For each scenario and each sensing modality, the corresponding monocular image sequences are input into Kimera. It should be noted that significant changes to Kimera's default parameters had to be made for successful tracking and mapping, since Kimera is usually used for non-object-centric SLAM with wide field-of-view cameras. Parameters are changed for Kimera's feature tracker, optimizer, and loop closure detector modules. Additionally, some modifications had to be implemented in the software, such as handling of blank images (which the VIS camera sees during eclipse). All scenarios and sensing types use the same parameter set other than changes in camera intrinsics for the different camera resolutions and fields of view. Kimera's noise parameters are also set to match the noise in the relative dynamics measurements (as described in \autoref{section:rel_dynamics}).

The estimated trajectory output by Kimera is relative to the target body frame, so in the absence of a relevant prior, an infinite family of valid solutions exists, related by an arbitrary rigid transformation of both the target's landmark positions and the inspector's trajectory. As a result, an initial estimate of the inspector's relative pose is input to anchor its trajectory estimate and the position of the target in Kimera's coordinate system. In practice, this initial relative pose can be arbitrarily chosen.

The estimated trajectory can be created in two ways. The first method evaluates the estimated pose as each measurement and image comes in over the course of the scenario. The second method uses the batch-optimized estimate solved at the end of the trajectory, assuming all measurements and images are available from the full scenario. We follow the first method when analyzing the performance of this system. Real-time navigation errors are important to analyze because they directly affect closed-loop guidance and control of the inspector. While the batch method is useful for post-inspection analysis, this work focuses on navigation during an active inspection.

\begin{figure}[ht]
\centering
\subcaptionbox{Features are tracked on a TIR image of the target spacecraft. Green points are features that are detected in the current frame and are being tracked from previous frames. Blue points are features that have been detected, but are not yet tracked. Red points are outlier features.}[0.45\textwidth]{
    \centering
    \includegraphics[height=4cm]{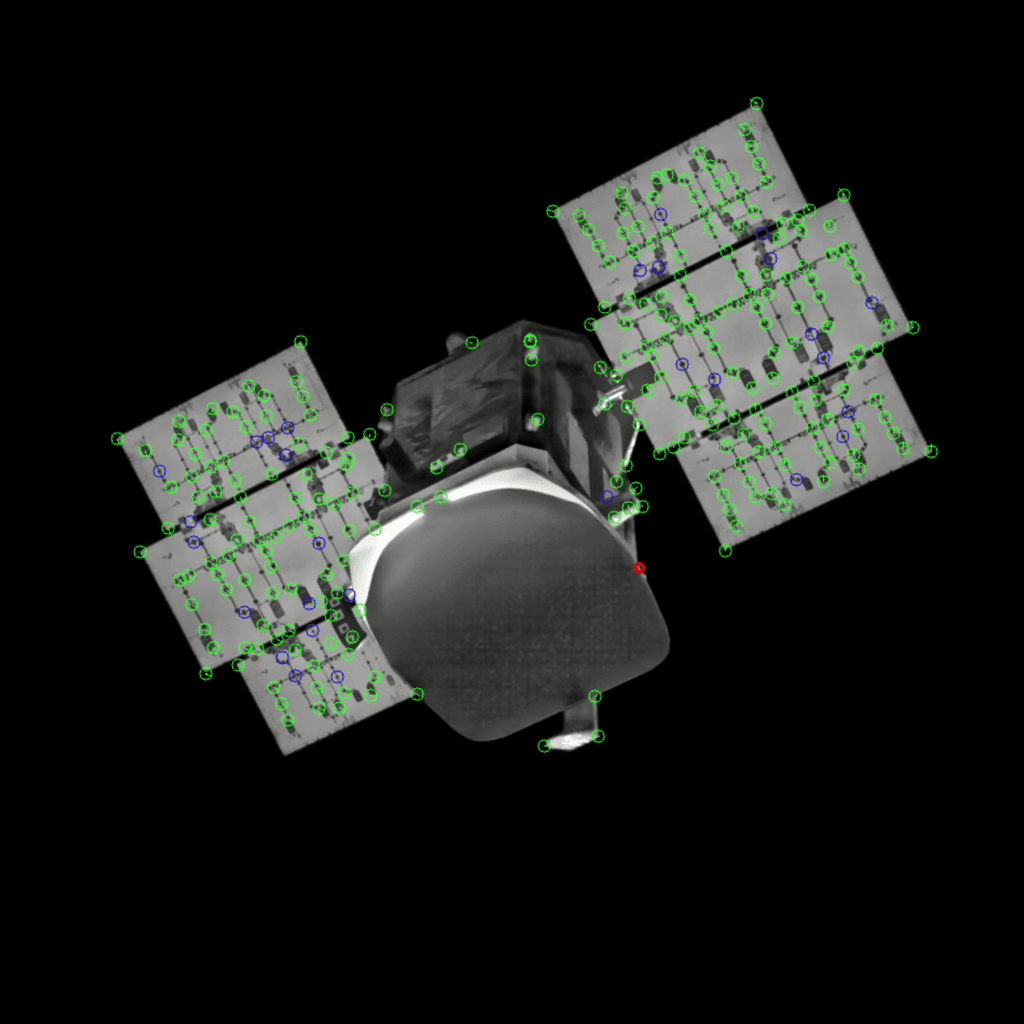}
}
\hspace{0.4cm}
\subcaptionbox{The estimated 3D positions of the landmarks tracked on the target (white points) and the estimated trajectory of the inspector (red curve).}[0.45\textwidth]{
    \centering
    \includegraphics[height=4cm]{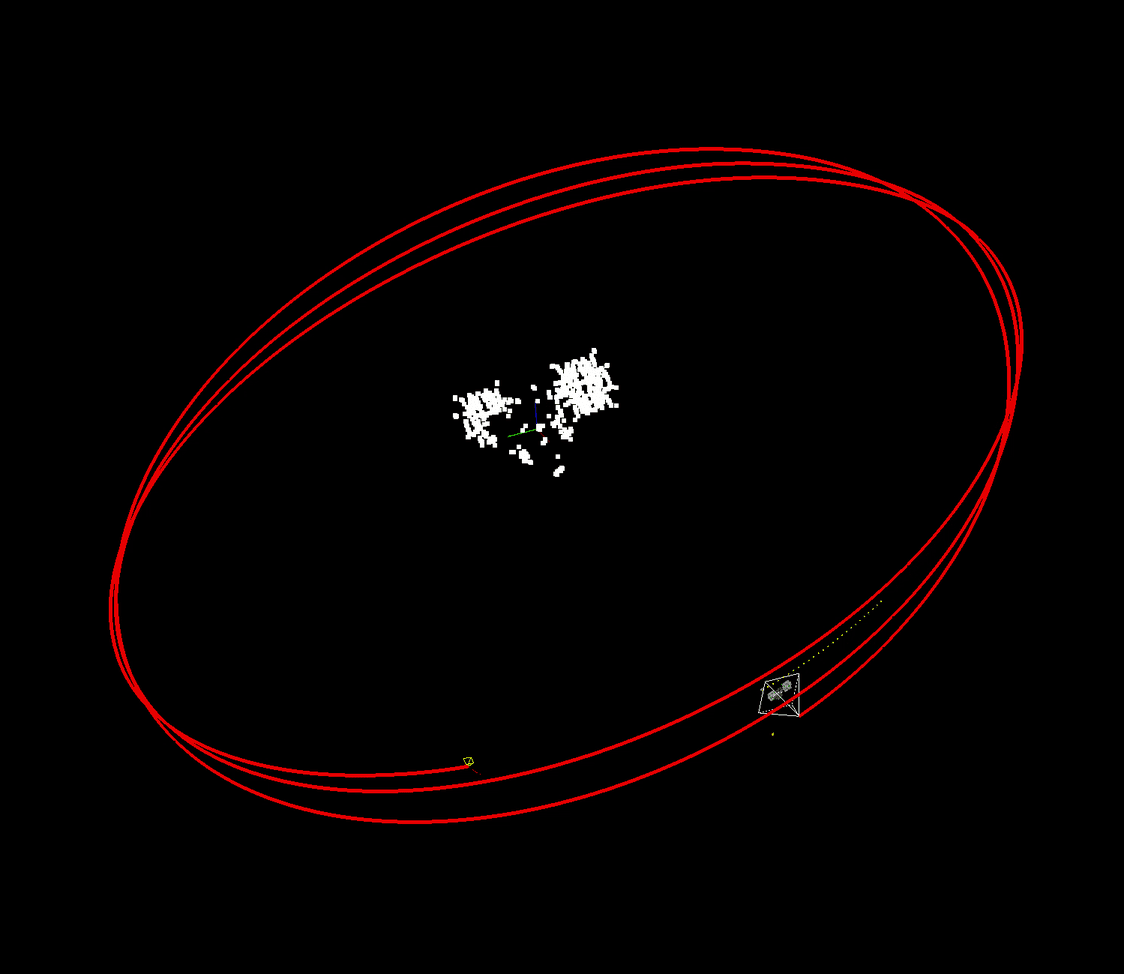}
}
\caption[Kimera feature tracking, trajectory estimation, and mapping]{Kimera tracks features on sequential images of the target spacecraft. The feature tracks are used to create a 3D map of the target and estimate the trajectory of the inspector.}
\label{fig:kimera}
\end{figure}

\begin{figure}[ht]
\centering
\bigskip
\includegraphics[width=\textwidth]{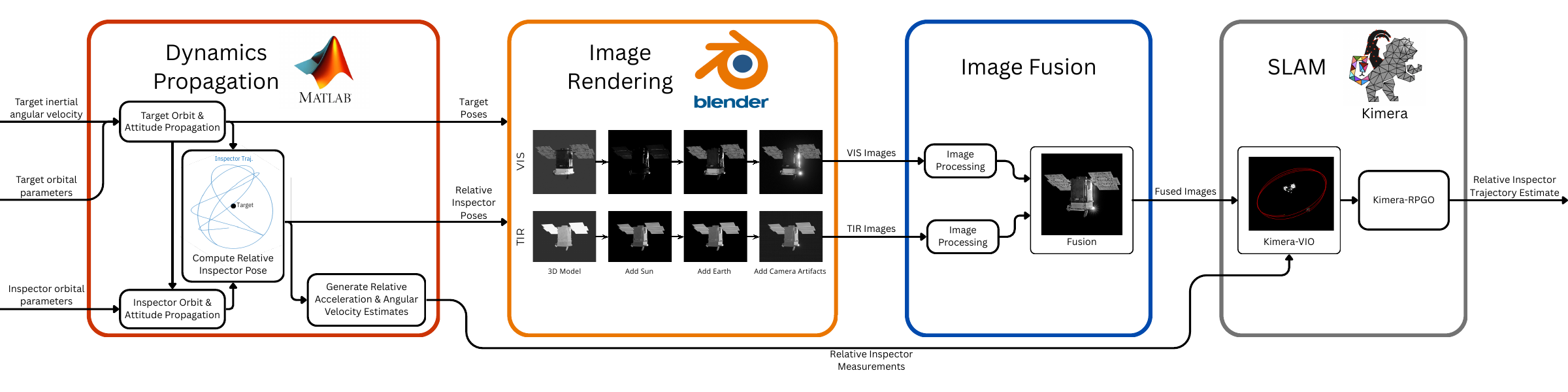}
\caption{The full pipeline used for dynamics propagation, rendering, fusion, and SLAM.}
\label{fig:fullPipeline}
\end{figure}

The camera images are input at 0.1 \si{\hertz} and the relative dynamics measurements are input at 10 \si{\hertz}. An example of Kimera's operation is seen in \autoref{fig:kimera} and a full system architecture is included in \autoref{fig:fullPipeline}.

%% file: imgSeqTable.tex
\def\imgWidth{1.7cm}
\newcommand{\visFig}[1]{\begin{minipage}[c][1.8cm][c]{\imgWidth}\includegraphics[width=\imgWidth]{#1}\end{minipage}}
\newcommand{\tirFig}[1]{\begin{minipage}[c][1.4cm][c]{\imgWidth}\includegraphics[width=\imgWidth]{#1}\end{minipage}}

\begin{figure}[tp]
\centering
\begin{tblr}{
  colspec = {Q[c,m]*{9}{Q[c,m]}},
  colsep = 0pt,
  column{2} = {leftsep=3pt},  
}
\hline \hline
\SetCell[c=9]{c} \textbf{S1: CloudSat 15 m} \\
\rotatebox[origin=c]{90}{VIS} 
  & \visFig{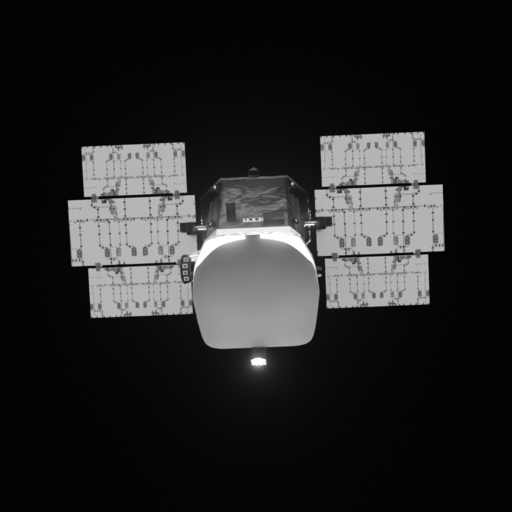} & \visFig{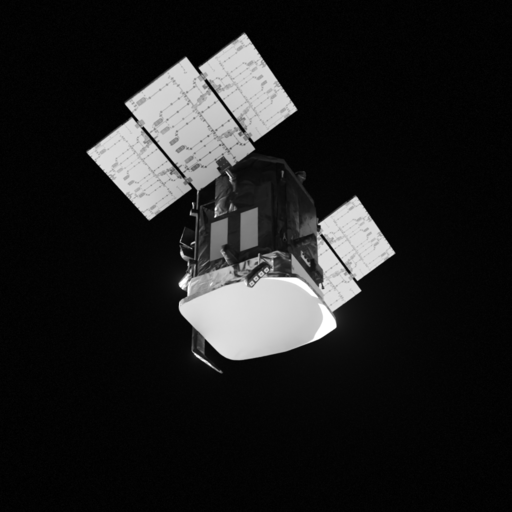} & \visFig{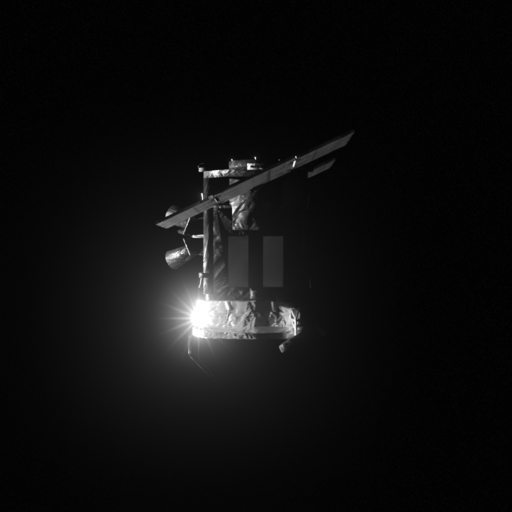} & \visFig{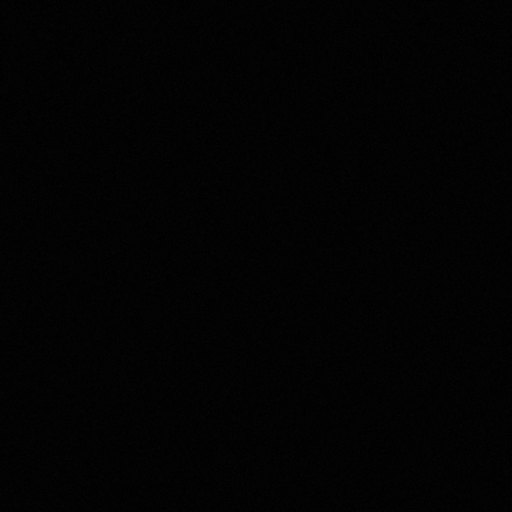} & \visFig{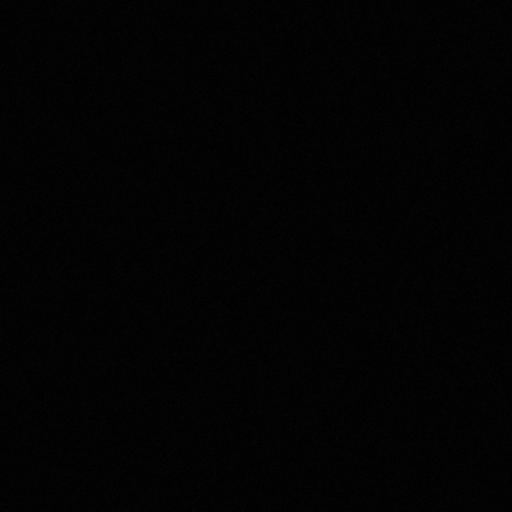} & \visFig{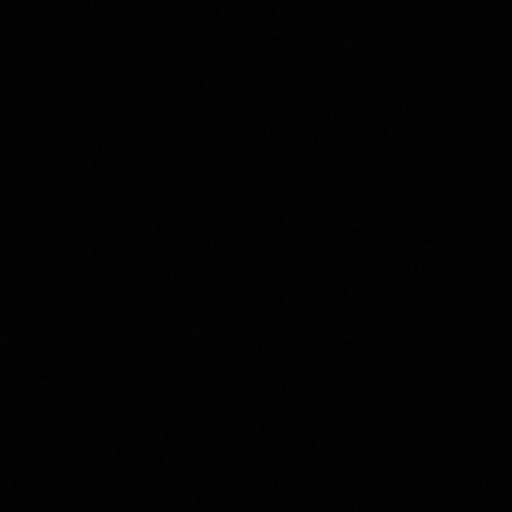} & \visFig{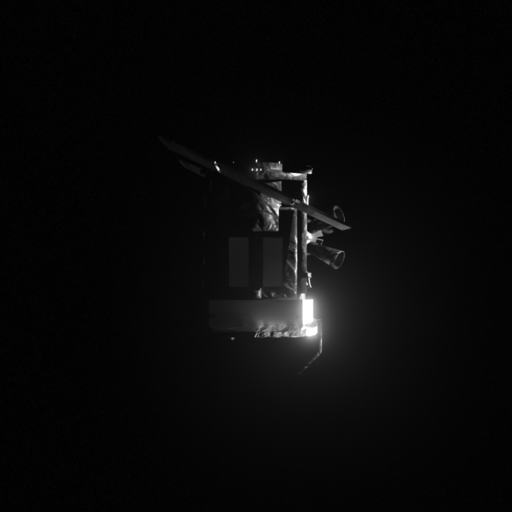} & \visFig{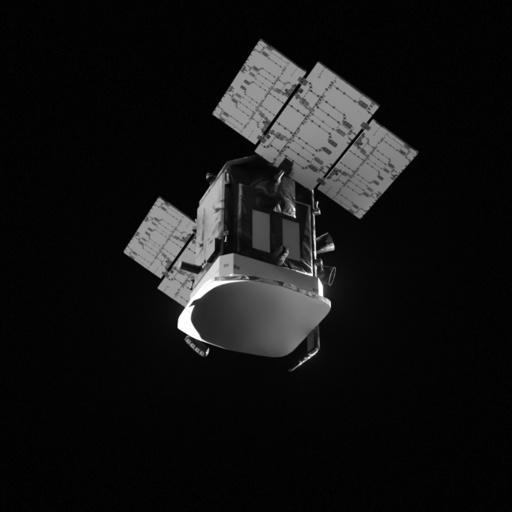} & \visFig{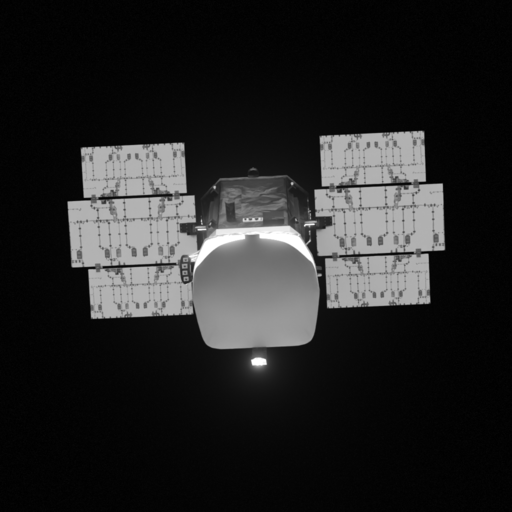} \\
\rotatebox[origin=c]{90}{TIR} 
  & \tirFig{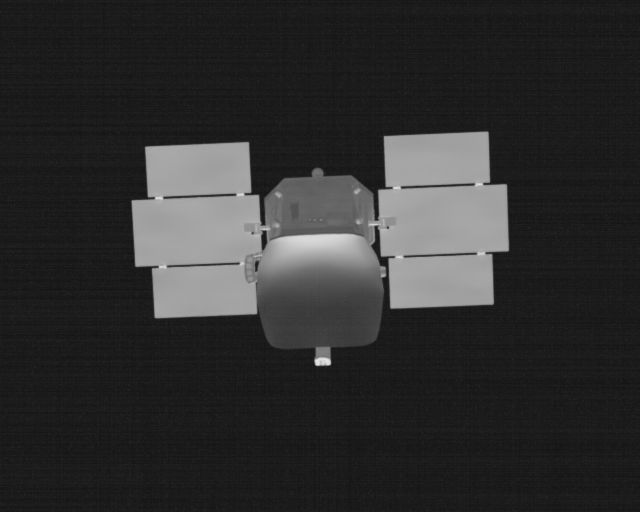} & \tirFig{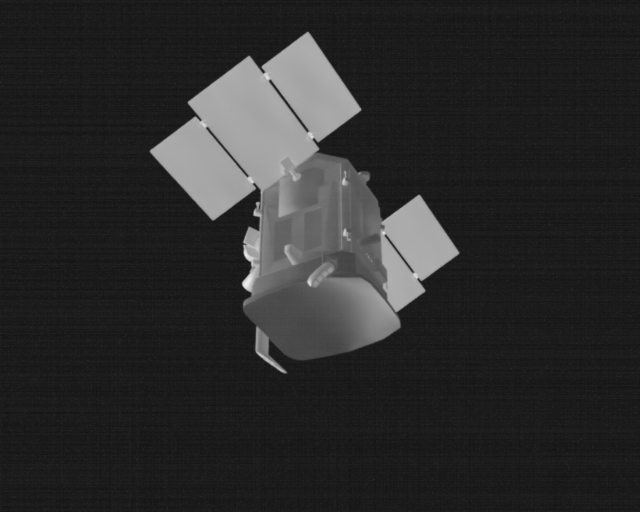} & \tirFig{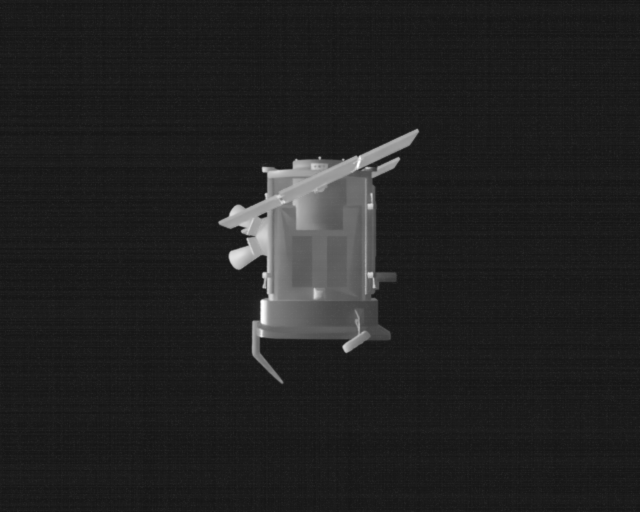} & \tirFig{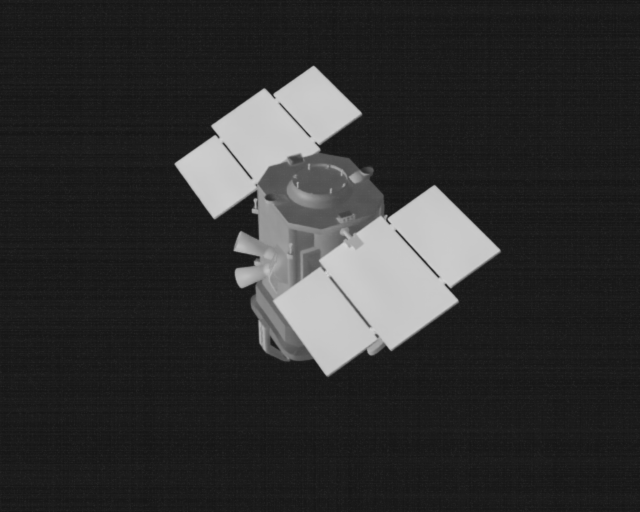} & \tirFig{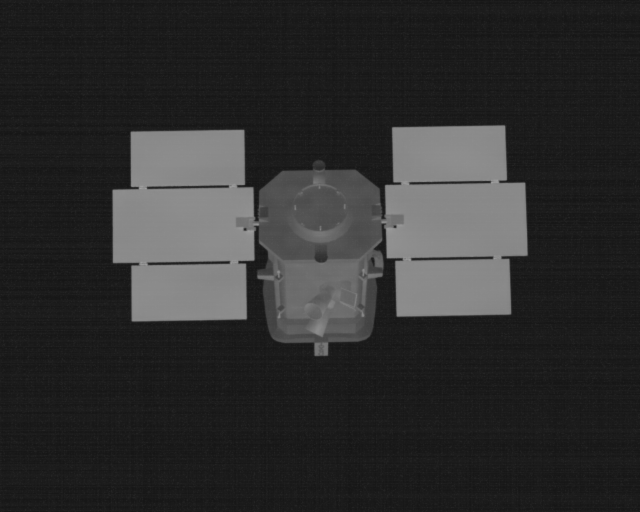} & \tirFig{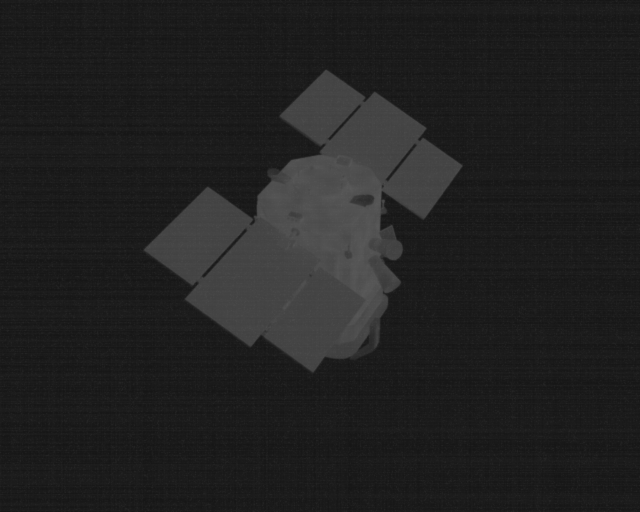} & \tirFig{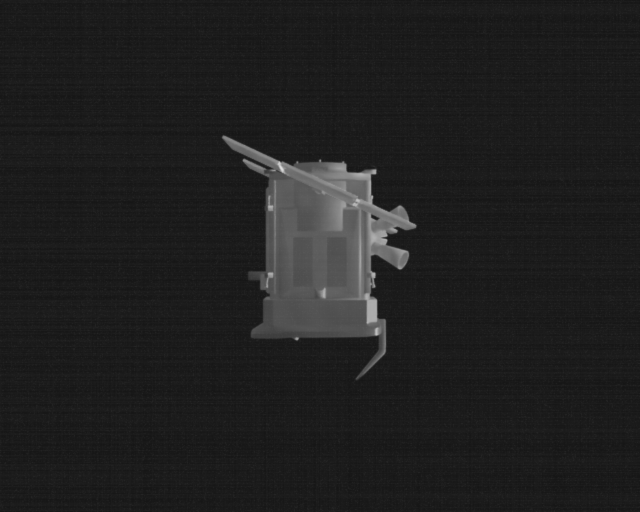} & \tirFig{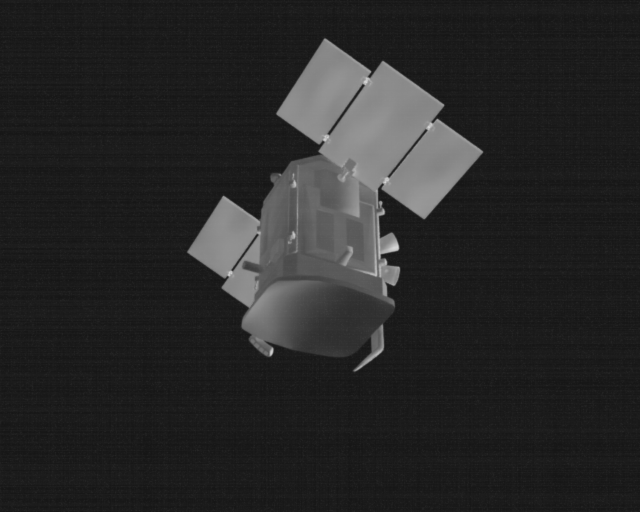} & \tirFig{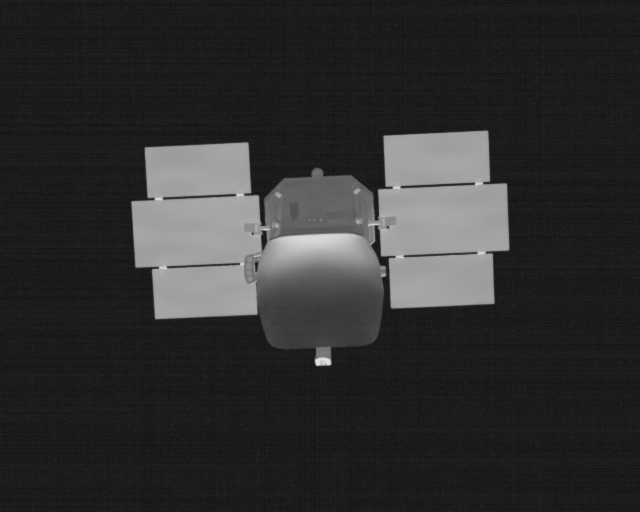} \\
\rotatebox[origin=c]{90}{Fused} 
  & \visFig{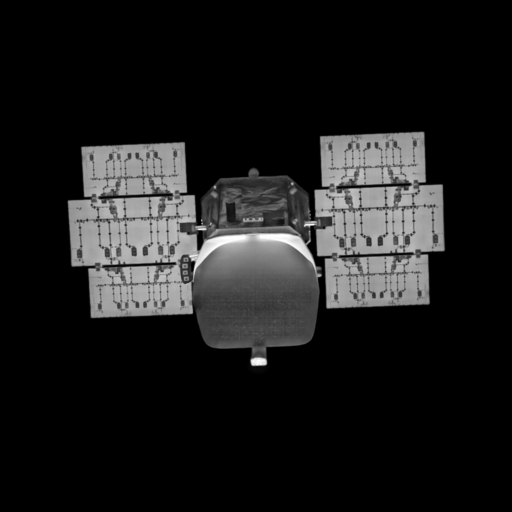} & \visFig{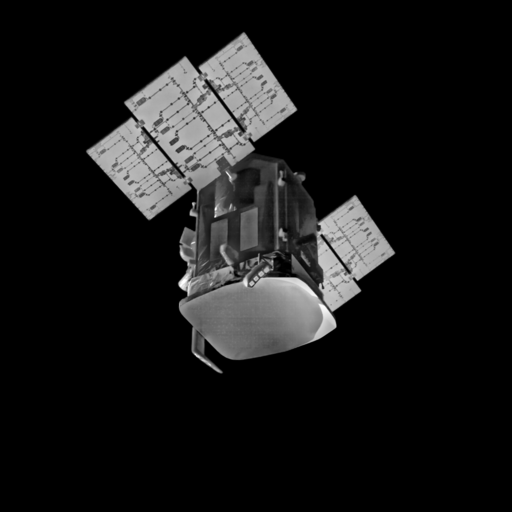} & \visFig{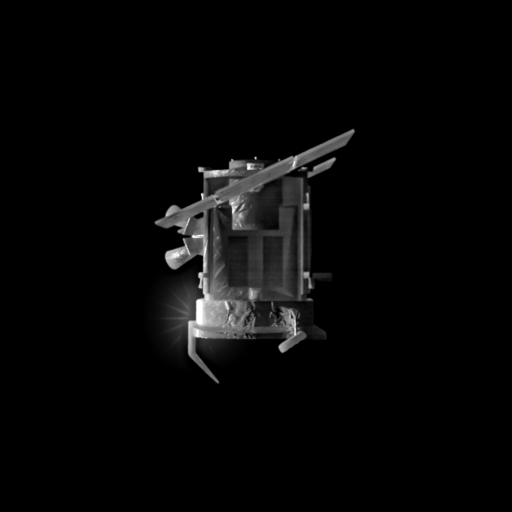} & \visFig{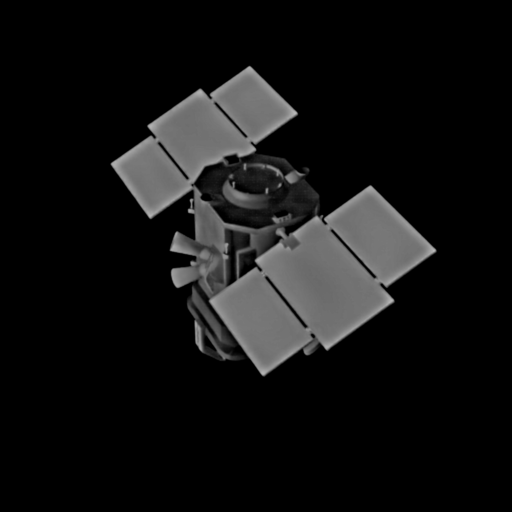} & \visFig{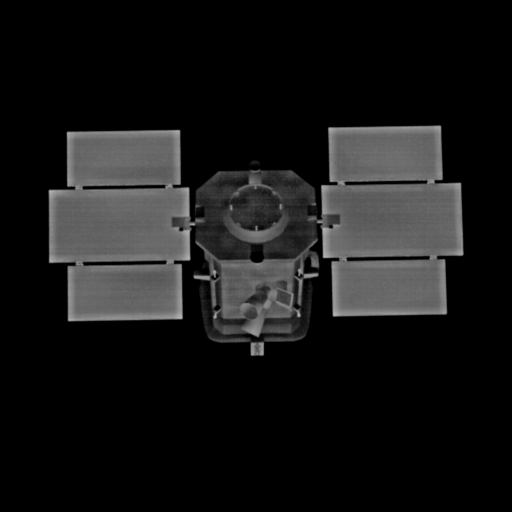} & \visFig{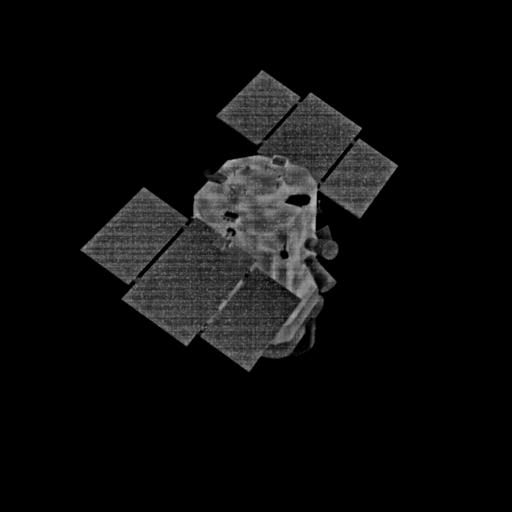} & \visFig{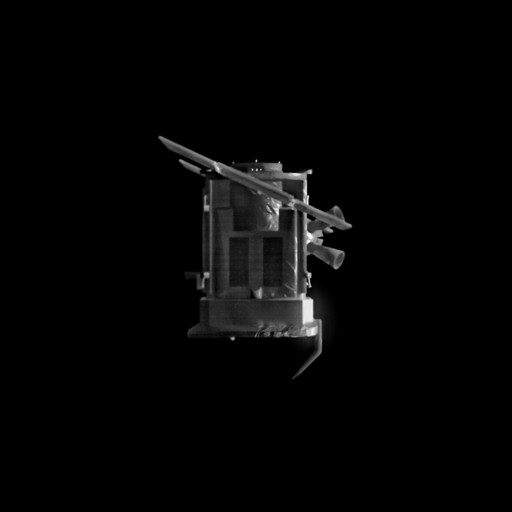} & \visFig{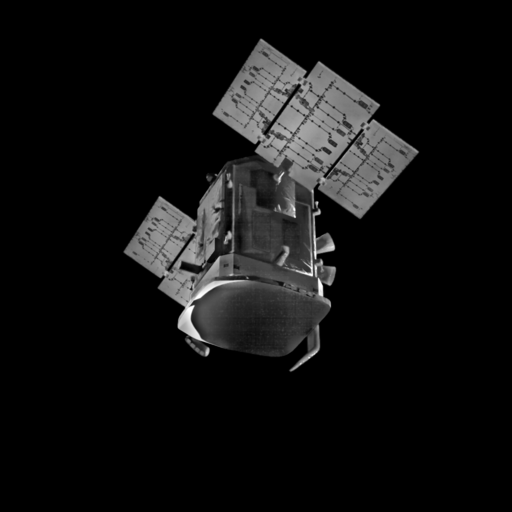} & \visFig{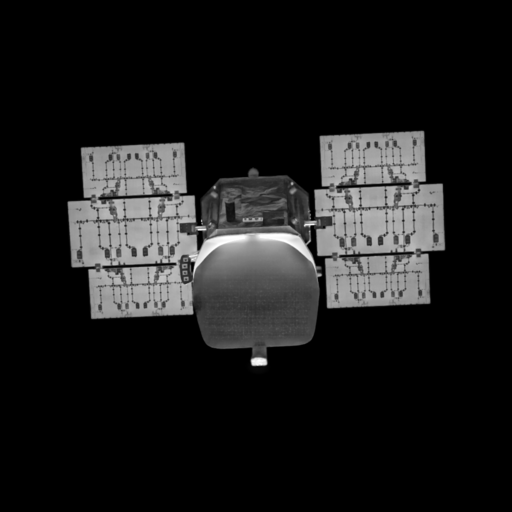} \\
\hline
\SetCell[c=9]{c} \textbf{S6: CubeSat 15 m, Rotating} \\
\rotatebox[origin=c]{90}{VIS} 
  & \visFig{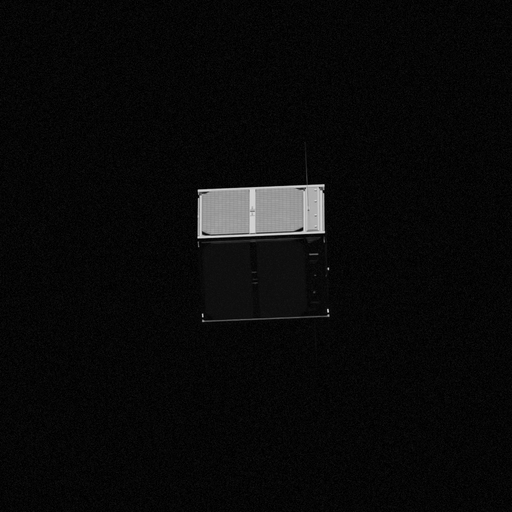} & \visFig{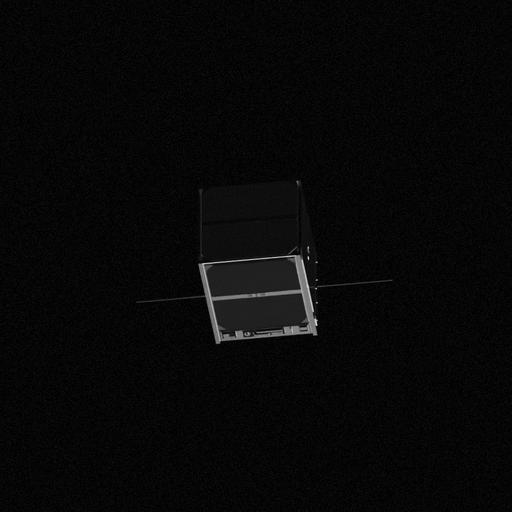} & \visFig{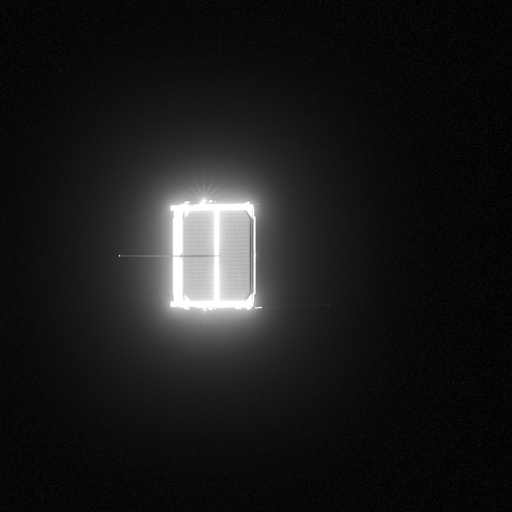} & \visFig{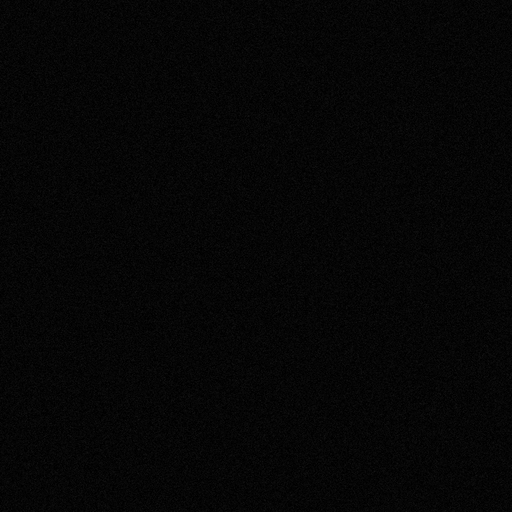} & \visFig{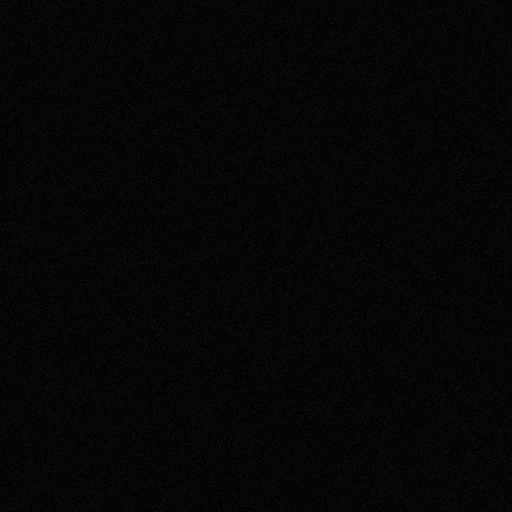} & \visFig{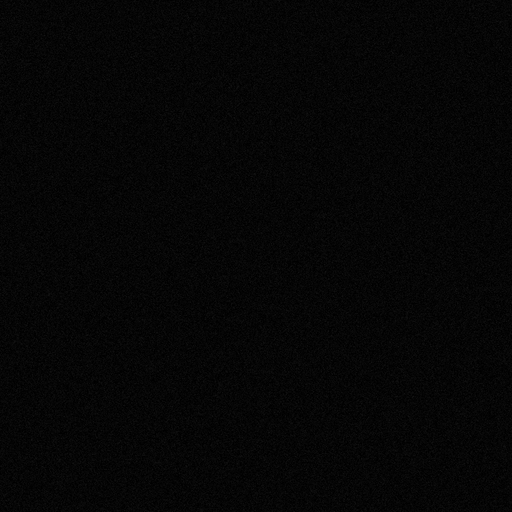} & \visFig{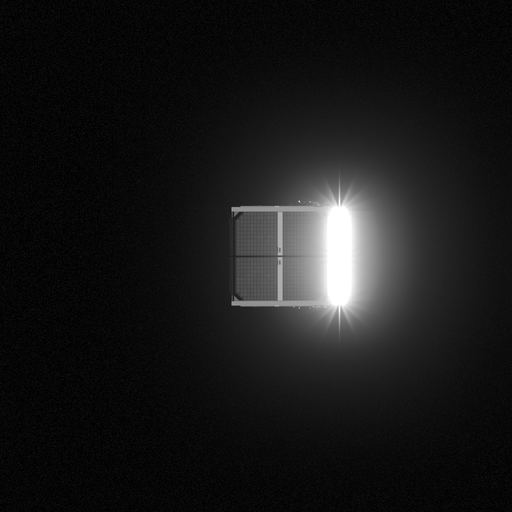} & \visFig{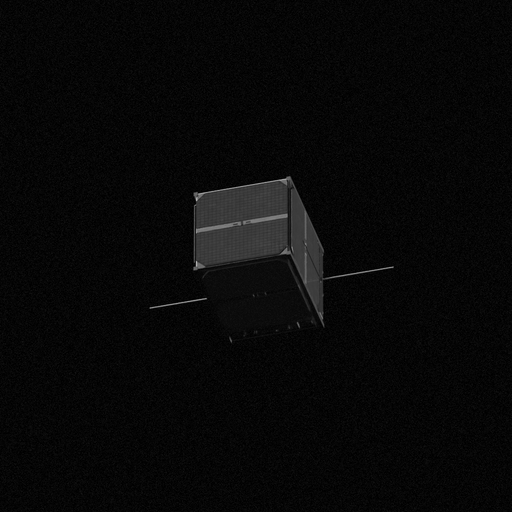} & \visFig{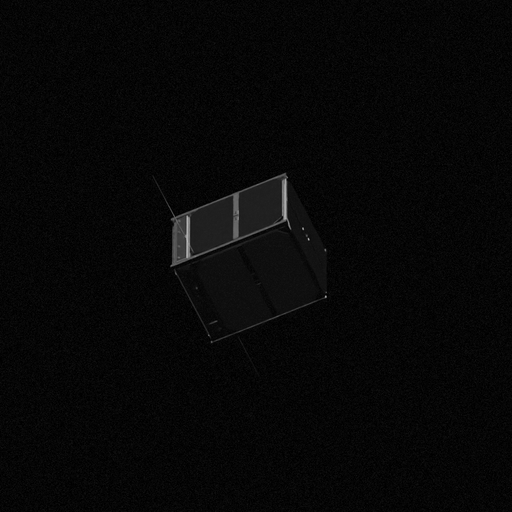} \\
\rotatebox[origin=c]{90}{TIR} 
  & \tirFig{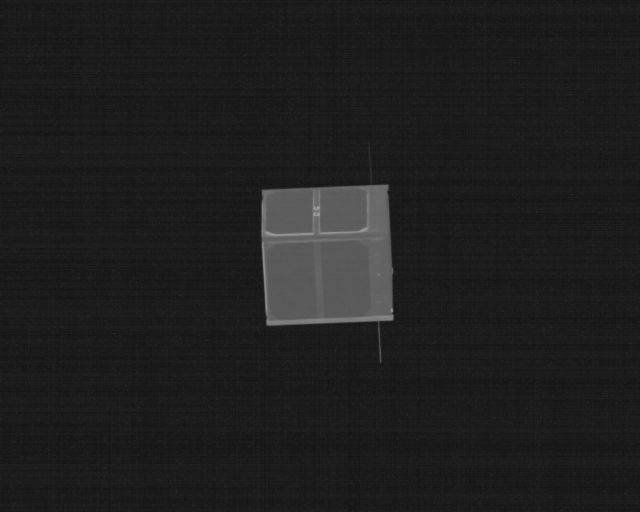} & \tirFig{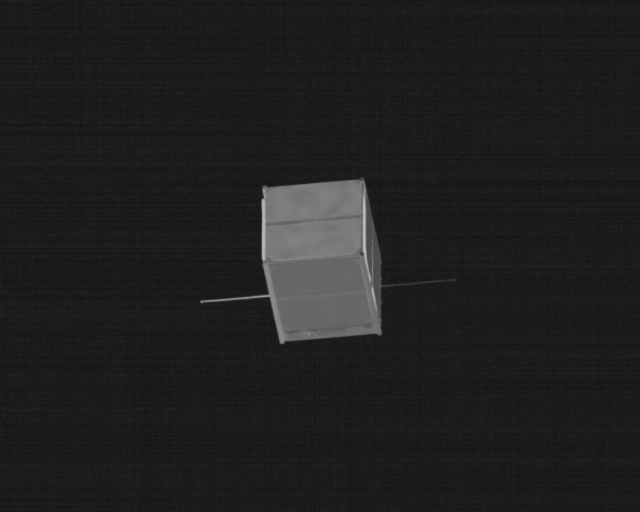} & \tirFig{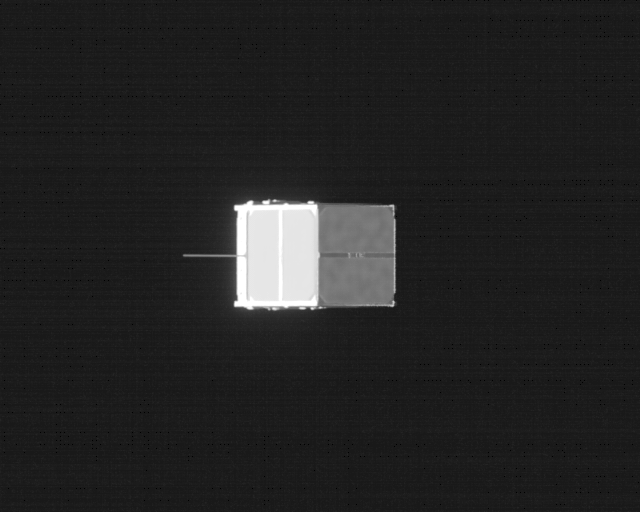} & \tirFig{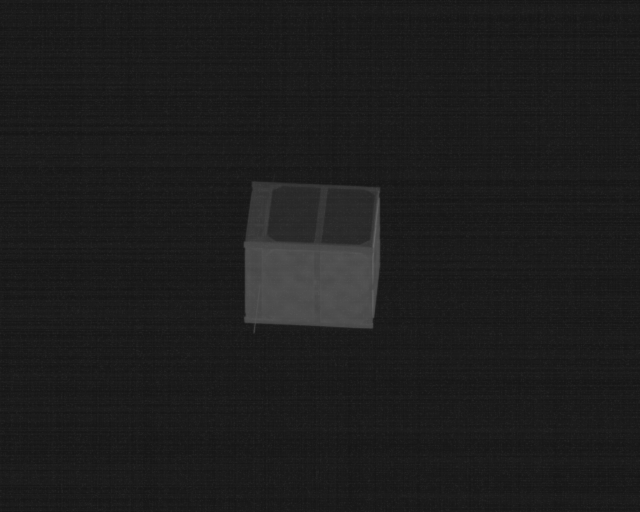} & \tirFig{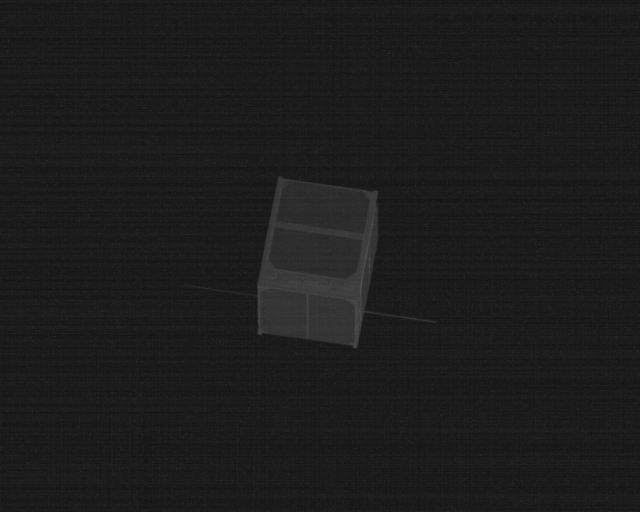} & \tirFig{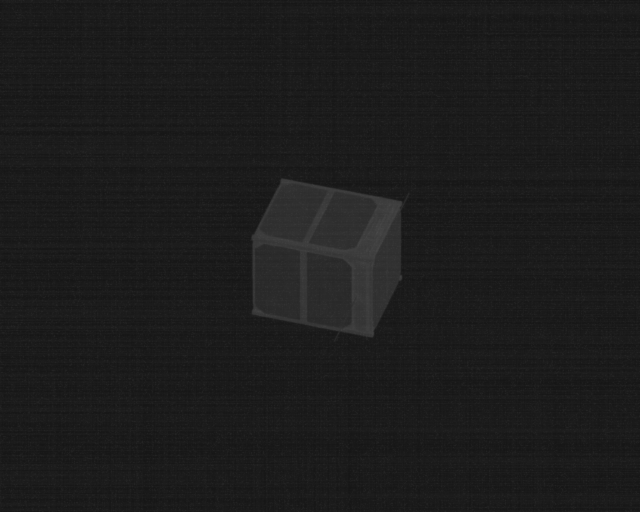} & \tirFig{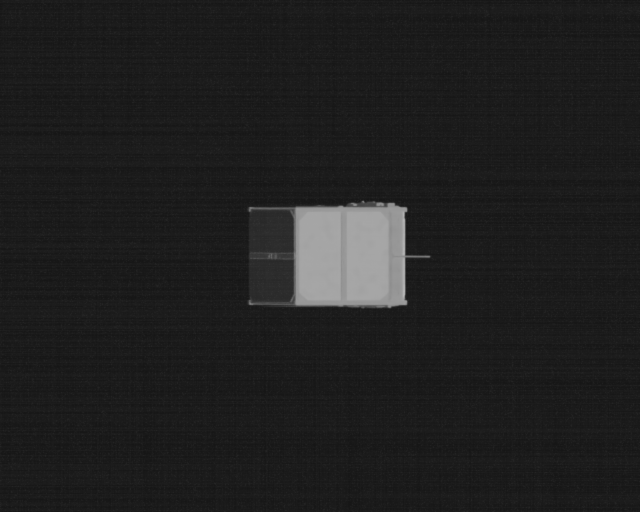} & \tirFig{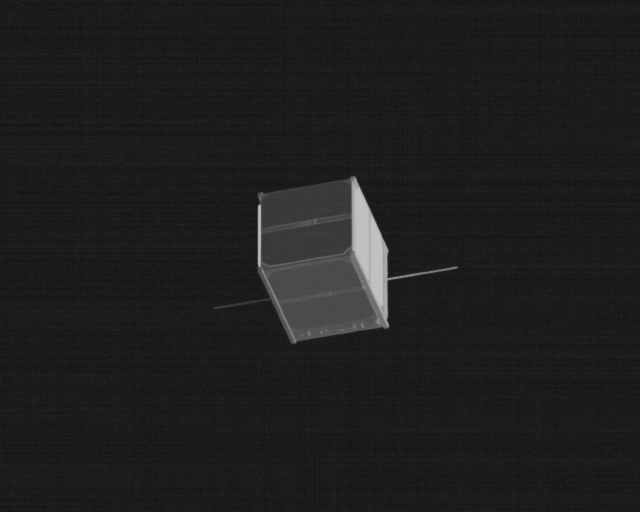} & \tirFig{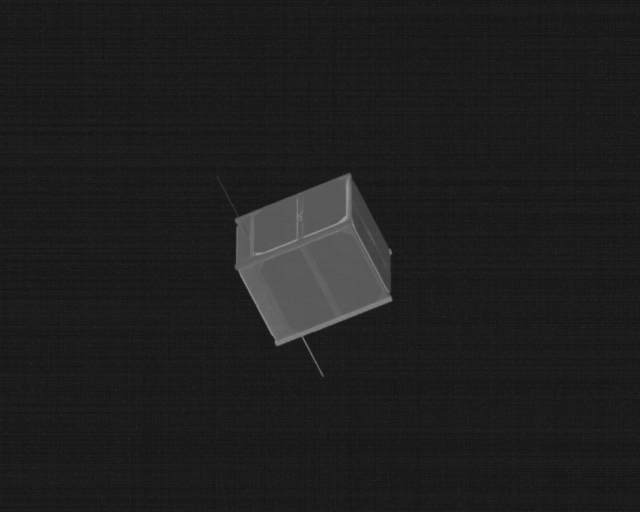} \\
\rotatebox[origin=c]{90}{Fused} 
  & \visFig{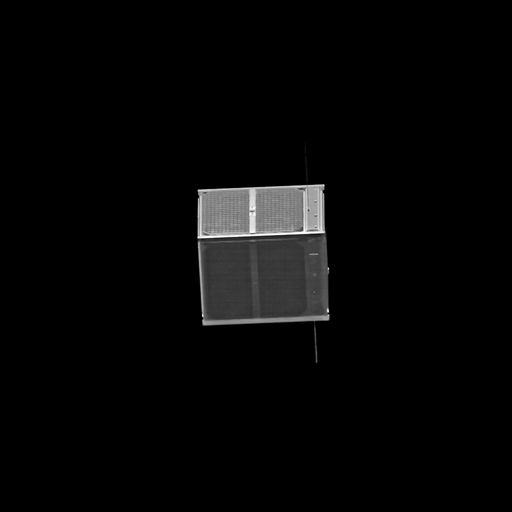} & \visFig{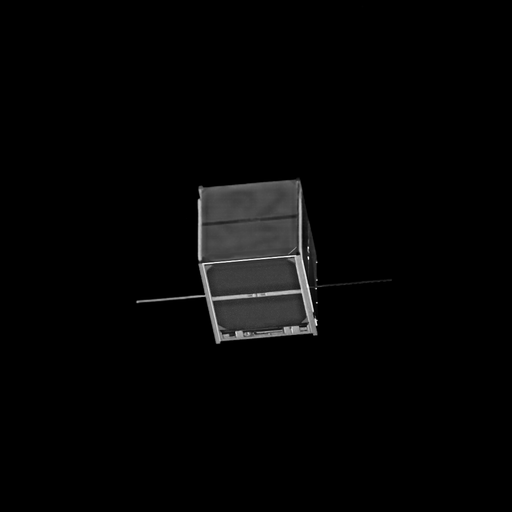} & \visFig{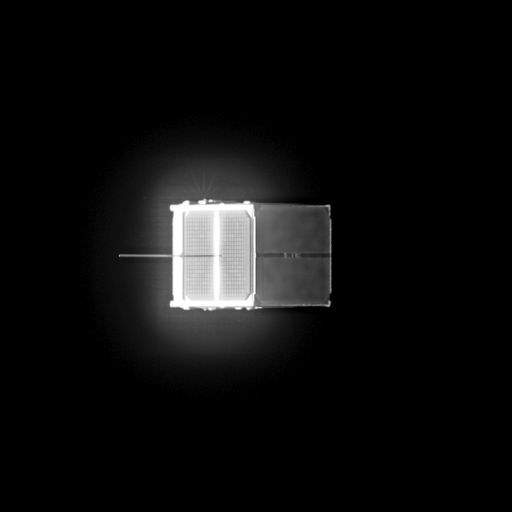} & \visFig{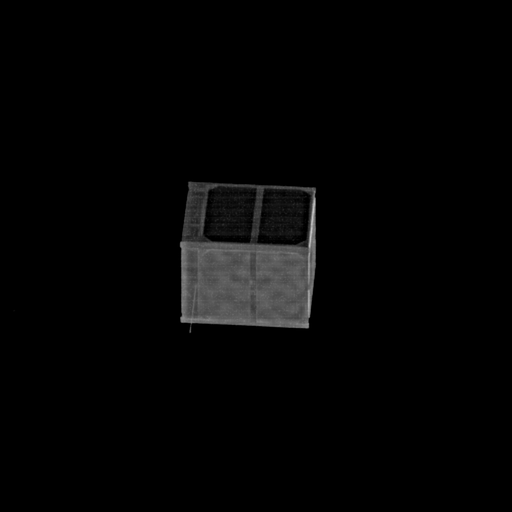} & \visFig{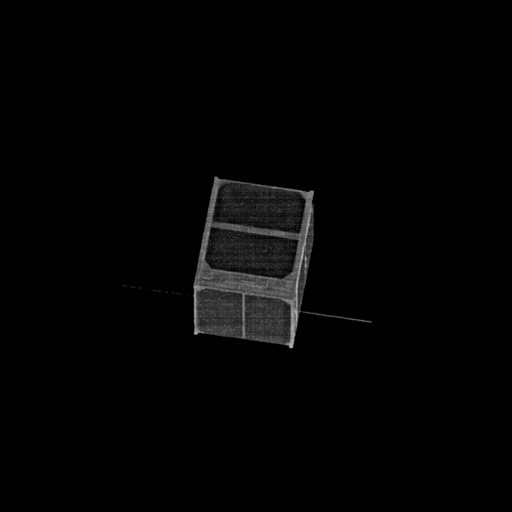} & \visFig{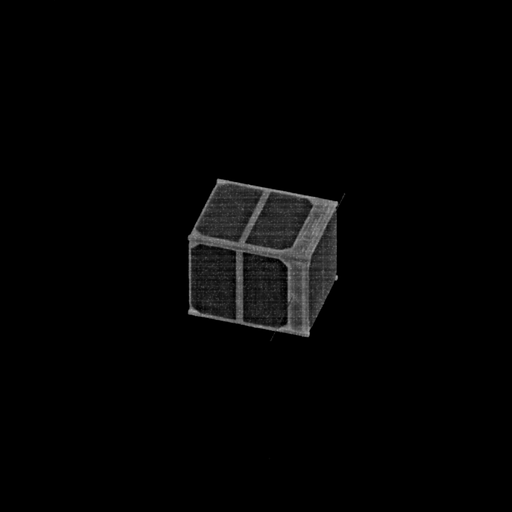} & \visFig{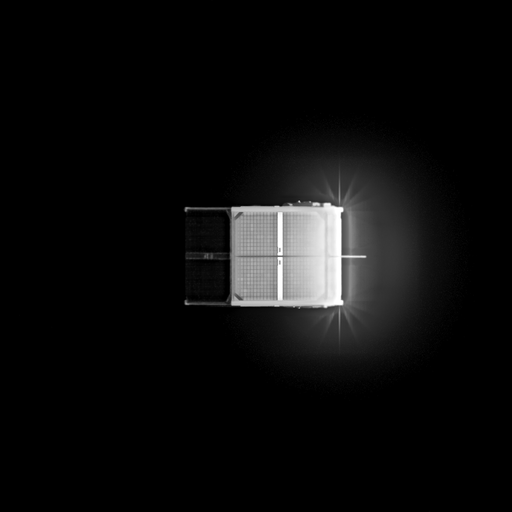} & \visFig{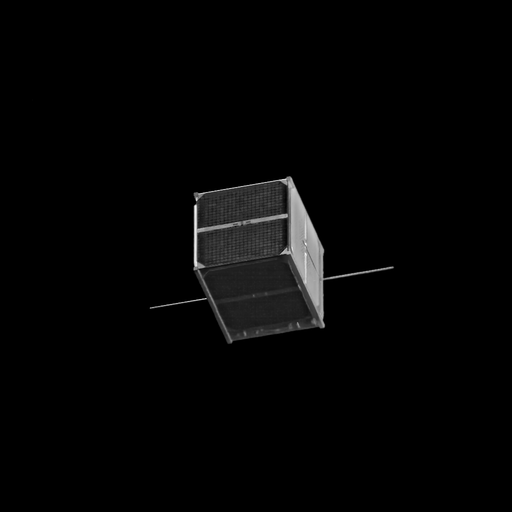} & \visFig{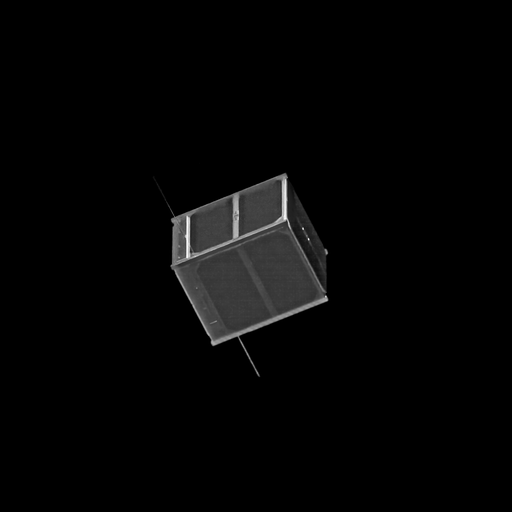} \\
\hline \hline
  & 0 & 11.5 & 23 & 34.5 & 46.17 & 57.67 & 69.17 & 80.67 & 92.17 \\
  & \SetCell[c=9]{c} Time, mins \\
\end{tblr}
\caption{Simulated imagery of the target spacecraft at specified times on the inspector's trajectory for the various sensor types. This sequence shows the first orbit of Scenario 1 with CloudSat as the target. The full scenario has a similar sequence repeated twice more, for a total of three orbits. The first orbit of Scenario 6 is also shown with a 1U CubeSat as the target. Unlike in Scenario 1, the CubeSat is spinning in the inertial frame, causing the images in the following two orbits to be different from this first orbit.}
\label{fig:renders}
\end{figure}

%% file: 5_loopClosure.tex
\section{Silhouette-Based Place Recognition} \label{sec:silhouette_loop_closure}

Accurate, frequent loop closures are extremely important when navigating around an RSO. Localization performance is severely degraded when visual-inertial odometry causes the estimate to drift and few loop closures are detected. In the renderings generated in this work, Scenarios 3 and 6 are particularly difficult because the trajectory in the body frame of the target does not retrace itself with each orbit, resulting in only 12 opportunities for a loop closure as shown in \autoref{fig:rotScenario}.

There are many existing VPR methods that can be applied to uncooperative RSOs; however, they do not perform acceptably in this application for three reasons. First, the loop closure opportunities occur throughout the trajectory, with the matching frames often containing drastically different illumination conditions. \autoref{fig:lc_opportunities} shows the imagery from each camera during each of the loop closure opportunities. VPR and pose recovery methods operate on frames temporally far apart, causing the illumination issues to be even more significant than when performing frame-to-frame feature tracking. The visible imagery is often in eclipse, heavily shadowed, or corrupted by glare/reflections. As a result, images from the same perspective of the RSO often look very different at different points in time. The thermal-infrared camera is more robust to these changes; however, surface temperature changes still create differing thermal imagery from the same perspective.

\input{LCmatchesTable}

The second reason these methods struggle with this scenario is that they were not designed for space applications. They are trained using images from indoor and outdoor environments on the ground. However, the features seen in these environments differ substantially from the features seen in space. Instead of edges of windows, corners of desks, and markings on a road, our imagery includes solar arrays, thrusters, and the spacecraft bus. This creates a situation in which the types of features detected cannot be properly classified because of the domain gap between the training and test datasets, resulting in a large performance loss.

The last reason prior VPR methods perform poorly is that they are trained only with visible imagery. However, to stay robust to eclipse, the thermal camera must be used. The features in visible and thermal cameras look very different, so these methods cannot match the features in the thermal imagery effectively.

One solution is to retrain these methods using datasets containing imagery of RSOs. Thermal images would mostly be relied on due to robustness to eclipse and shadows.  Retraining the Bag of Words vocabulary has been performed by Keil et al.~\cite{keil_towards_2024} to run VPR on thermal imagery on the ground over day-night changes. Training, however, requires a large dataset across a domain representative of the conditions that would be seen in use. For example, Keil et al. retrained the Bag of Words vocabulary with approximately 38,000 pairs of images, including pairs containing both a day and a night image. Images or photo-realistic renderings of RSOs are far more difficult to obtain, especially a diverse thermal image dataset of this size. As a result, an alternative that avoids any training is desired.


\subsection{Algorithm Design}

We propose a simple, efficient, high-performance method to recognize when an inspector revisits the same perspective of an unknown RSO. First, a descriptor must be extracted from each image. In the thermal pairs of loop closure opportunities in \autoref{fig:lc_opportunities}, the most notable similarity (especially in the case of differing heating) between matching images is the silhouette, or boundary shape, of the spacecraft. If the current thermal silhouette of the RSO matches a previous thermal silhouette, it can be assumed that the inspector is viewing the RSO from a similar perspective. As a result, a loop closure can be detected by matching the silhouettes.

Silhouettes have previously been used extensively to navigate around RSOs. Template matching of silhouettes through a visible camera has been used for pose estimation, comparing many pre-rendered silhouettes of a known RSO to the current view~\cite{petit_vision_2012, guo_pose_2021}. Silhouettes of an unknown asteroid seen with a visible camera have also been used to determine the shape of the asteroid and/or the relative pose of an inspector~\cite{bandyopadhyay_silhouette_2019, villa_optical_2020, baker_limb_2020, panicucci_shadow_2023, moon_gaussian_2025}. However, these silhouette-based methods have not been used on thermal imagery and have not been used for place recognition around unknown bodies.

Template matching is a well-studied field that investigates methods to efficiently and accurately match a query shape to similar shapes in a database. Popular existing classical methods match shapes using descriptors derived from the Fourier transforms of the shape boundaries~\cite{wallace_efficient_1980}, the moment invariants of the shapes~\cite{hu_visual_1962}, or the inflection points of the shape boundaries as they are smoothed~\cite{abbasi_curvature_1999}. These approaches allow matching of templates that are far from identical, but that contain a similar overall structure.

We take a simpler approach, given that we would like to match almost identical silhouettes. Inspired by locality-sensitive hashing ~\cite{indyk_approximate_1998, jafari_LSH_2021}, which relies on the intentional collisions of descriptors, effort is made to make our method accurate and fast without the need of a GPU. Our silhouette matching method consists of the following steps:

\begin{enumerate}
  \item The image is thresholded so that a silhouette of the RSO is obtained. This operation is sensitive to glare and bloom, so using the thermal imagery shows the best performance compared to visible or fused.
  \item To make the descriptor invariant to the scale, rotation, and translation of the silhouette in the image, the silhouette is normalized. The silhouette is translated so that the centroid is in the center of the frame, scaled so that the filled area is a specified proportion of the frame, and rotated to align the second moment of area horizontally.
  \item $d$ equally-spaced rays are drawn out from the center of the silhouette. The farthest point along each ray that intersects the silhouette boundary is found.
  \item The position of each ray's intersection is discretized to fall within $b$ equally-spaced annuli.
  \item The index of the annulus that each intersection falls in is vectorized into a descriptor with order preserved. As a result, the descriptor $\mathbf{v}$ takes the form 
    \begin{equation}
        \mathbf{v} = (v_1, v_2, \ldots, v_d), \hspace{1em} v_i \in \{0, 1, \ldots, b-1\} \hspace{0.5em} \forall i \in \{1, 2, \ldots, d\}
    \end{equation}
  \item The descriptor is repeatedly circularly shifted, where the first element is moved to the last, until the lexicographically minimal rotation is found. In other words, the shift that has the lowest digits at the beginning of the descriptor is found. Shifting the descriptor is synonymous with rotating the silhouette, so this step further lowers the sensitivity to any rotations of the silhouette.
  \item To improve performance, $n$ descriptors are found for each image, with each set of $d$ rays rotated by $\frac{2\pi}{dn}$ radians.
\end{enumerate}

\begin{figure}[htpb]
    \centering
    \includegraphics[width=.99\textwidth]{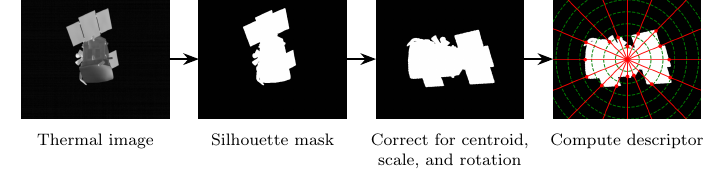}
    \caption[A visualization of the silhouette-based descriptor calculation process]{A visualization of the silhouette-based descriptor calculation process. In the right-most image, there are 16 rays shown, representing $d=16$. The intersection of each ray and the silhouette boundary are found, represented by the circular markers. $b=8$ in this case, representing 8 annuli separated by the dashed green circles. With the center-most circle having an index of 0 and starting from the right-most ray moving counterclockwise, each circular marker falls into the annulus indexed 3232122333321123. We then find the lexicographically minimal rotation of these values to find the final descriptor: 1123323212233332.}
    \label{fig:desc_method}
\end{figure}

\begin{figure}[htp]
    \centering
    \includegraphics[width=0.3\linewidth]{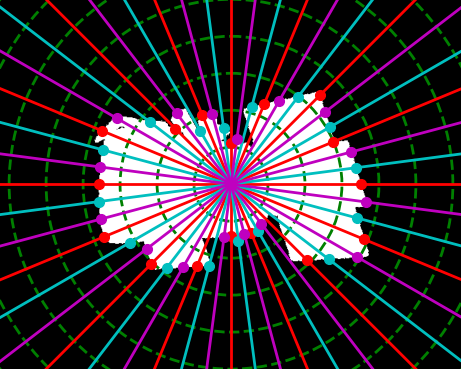}
    \caption[Calculation of descriptors when $n>1$]{The calculation of the descriptors with $d=16$, $b=8$, and $n=3$. Since $n=3$, three descriptors are found for this one silhouette. The red rays and circular markers represent the first descriptor, the turquoise represent the second, and the purple represent the third. This results in three unique descriptors found for this silhouette: 1123323212233332, 1123332211333322, and 1113333211233322.}
    \label{fig:desc_n3}
\end{figure}

A visualization of steps 1 through 6 is shown in \autoref{fig:desc_method}. Step 7 is shown in \autoref{fig:desc_n3}.

\begin{figure}[htpb]
    \vspace{\baselineskip}
    \centering
    \includegraphics[width=0.6\linewidth]{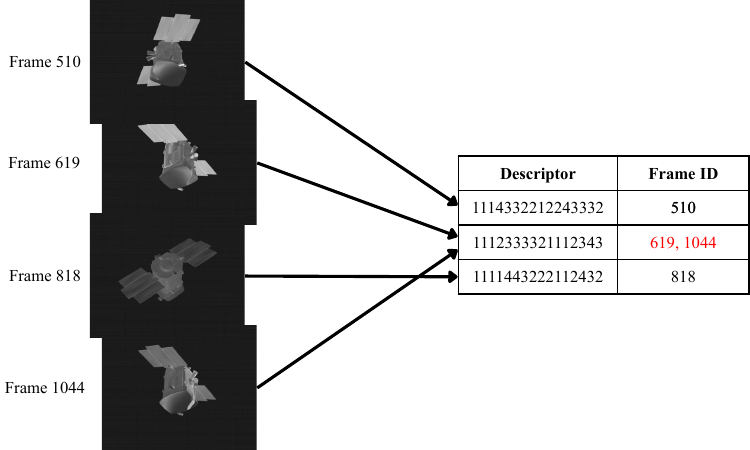}
    \caption[A visualization of how frames are matched using their descriptors]{A visualization of how frames are matched using their descriptors, borrowing ideas from locality-sensitive hashing. Each frame has its frame ID added to a hash table, with the key being its descriptor. When a frame is added to a row with previous frame IDs present, a match is identified as shown in red. This enables efficient matching without requiring a comparison against all previous frames. This example shows the case where $n=1$, meaning each image has one descriptor calculated. When $n>1$, each frame is added to the hash table $n$ times---once for each descriptor calculated.}
    \label{fig:descriptor_matching}
\end{figure}

In order to match descriptors, a hash table is used. Each descriptor found is used as the key, with the corresponding frame number appended to the list of values. Any time a descriptor already has existing frames listed under it, a matching silhouette is found. If the matched frame is more than a set number of frames in the past (to avoid matches to the most recent frames), a loop closure candidate is identified. This process is shown in \autoref{fig:descriptor_matching}.

\subsection{Pose Recovery} \label{sec:pose_recovery}

Now that a pair of images is obtained, the relative camera pose between the two images can be estimated. However, only the translation direction can be estimated, not its magnitude. This is due to the ambiguity that the RSO in the frame would look identical if it is small and close to the camera, as well as if it is large and far from the camera. Ultimately, only the rotation estimate is used in the loop closure constraint added to Kimera.

The accuracy of this pose recovery process is critical. The goal of loop closures is to correct for any accumulated frame-to-frame drift in the VIO solution. If the error in the loop closure pose estimate is too large, the addition of the loop closure in Kimera-RPGO will cause the navigation solution to \emph{increase} in error, even if it is not a clear outlier. The lower the errors in the VIO estimate, the more accurate the loop closures must be for them to be helpful.

The pose recovery (geometric verification) process begins by matching features between the images in each pair detected by the VPR module. As is done in Kimera~\cite{kimera}, ORB~\cite{rublee_orb_2011} features are used because they are computationally efficient and perform well. The five-point Nistér method~\cite{nister_efficient_2004} is then used with RANSAC~\cite{fischler_ransac_1981} to find inlier matches that solve for a relative pose. Instances when a small number of inlier matches are found often result in an inaccurate relative pose estimate due to the dependence on only a few feature matches. To decrease the chances of incorrect pose estimates, we impose a threshold for the minimum number of inlier feature matches. We find that a threshold value of 25 successfully filters out pose estimates with high errors without being too large to significantly diminish the number of loop closures. We also set a threshold for the maximum estimated rotation. We expect the silhouette matching method to only match images with a very similar perspective of the target. As a result, we do not use any loop closure candidates with an estimated rotation of more than 10 degrees (ignoring rotation about the camera boresight).

Pose recovery is performed using the thermal-infrared camera. Using the visible camera only is not an option because no features can be detected or matched during eclipse. Fused imagery offers features throughout the entire scenario with higher resolution and texture when available. However, unlike feature tracking for VIO which tracks frame-to-frame, this pose recovery process compares frames that are temporally far apart. As a result, the intense shadowing from the visible camera makes feature matching using the fused images difficult, performing worse than directly matching thermal imagery.


%% file: LCmatchesTable.tex
\begin{figure}[tb]

    \newcommand{\imgcell}[2]{%
      \begin{tabular}{@{}c@{}}
        \includegraphics[height=1.56cm]{#1} \\[-9pt]
        {\tiny #2}
      \end{tabular}%
    }

  \centering
  \begin{tabular}{c@{\hspace{2pt}}c@{\hspace{10pt}}c@{\hspace{2pt}}c}
    \toprule
    \multicolumn{2}{c}{\textbf{VIS}} & \multicolumn{2}{c}{\textbf{TIR}} \\
    \cmidrule(lr){1-2} \cmidrule(lr){3-4}
    \imgcell{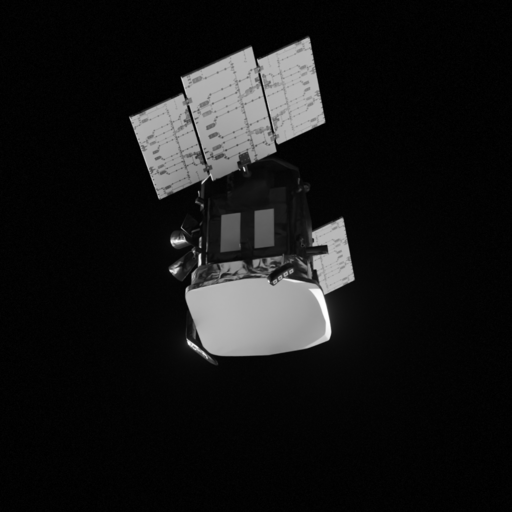}{Frame 66} &
    \imgcell{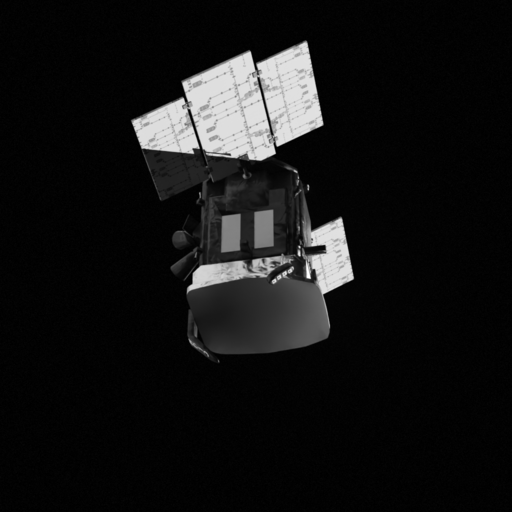}{Frame 490} &
    \imgcell{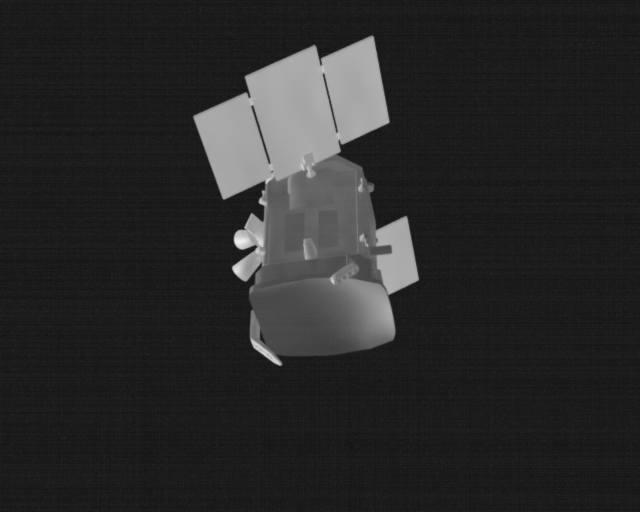}{Frame 66} &
    \imgcell{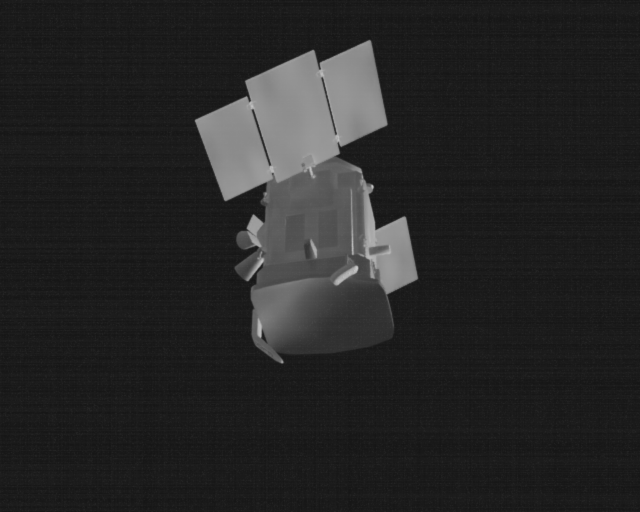}{Frame 490} \\[6pt]
    \imgcell{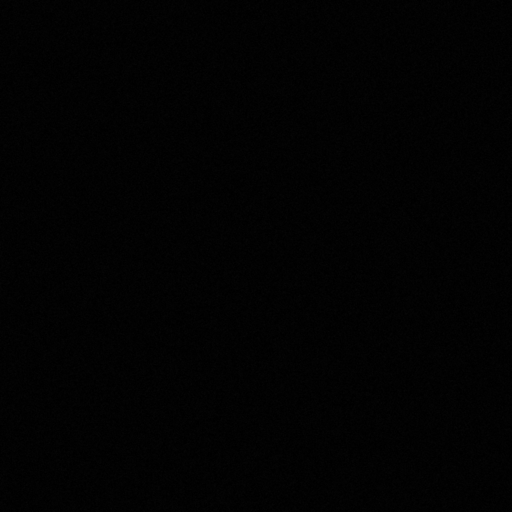}{Frame 341} &
    \imgcell{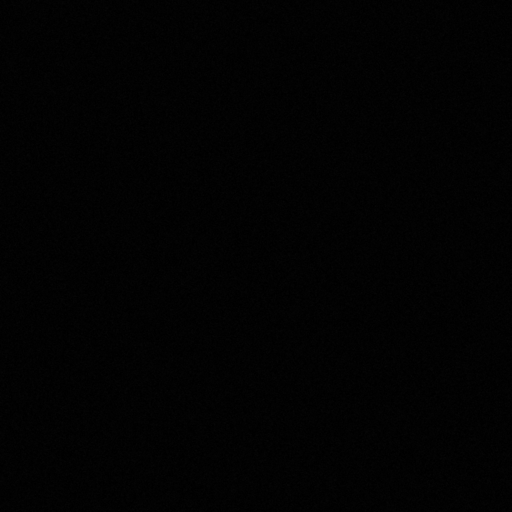}{Frame 767} &
    \imgcell{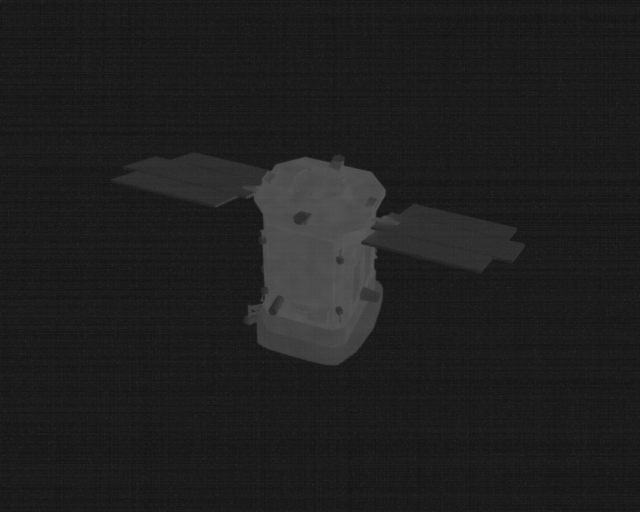}{Frame 341} &
    \imgcell{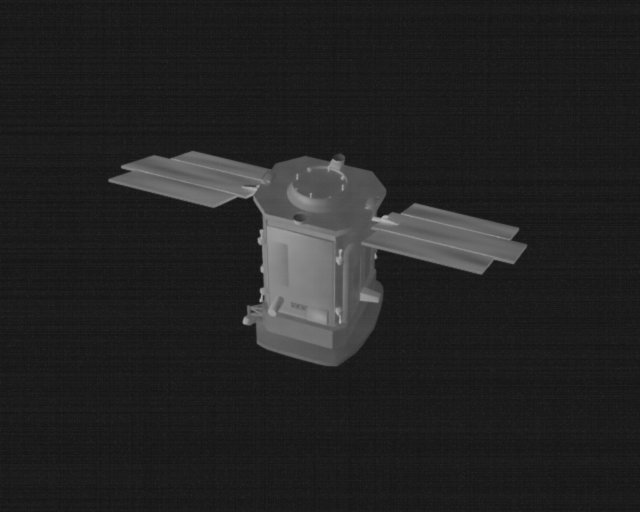}{Frame 767} \\[6pt]
    \imgcell{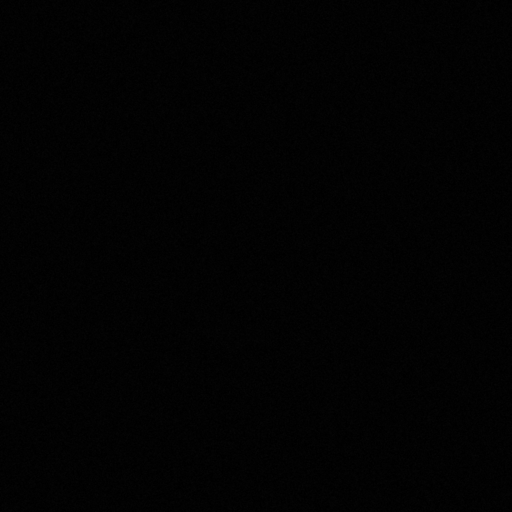}{Frame 184} &
    \imgcell{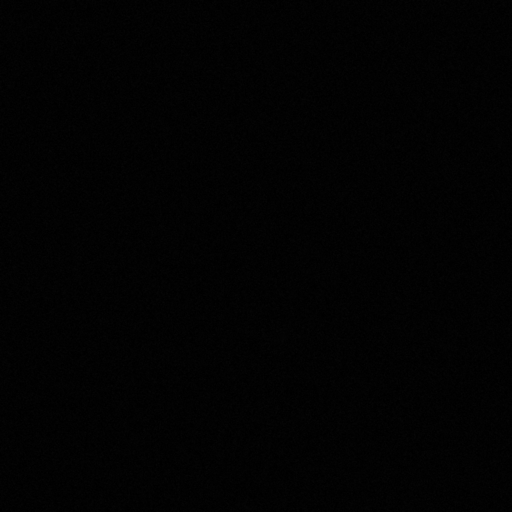}{Frame 924} &
    \imgcell{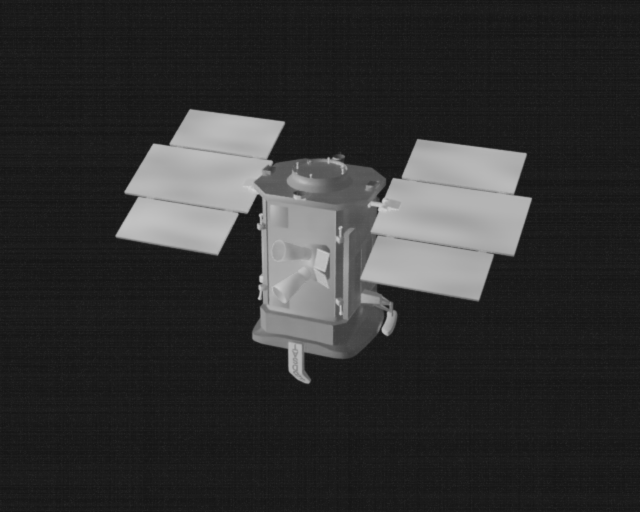}{Frame 184} &
    \imgcell{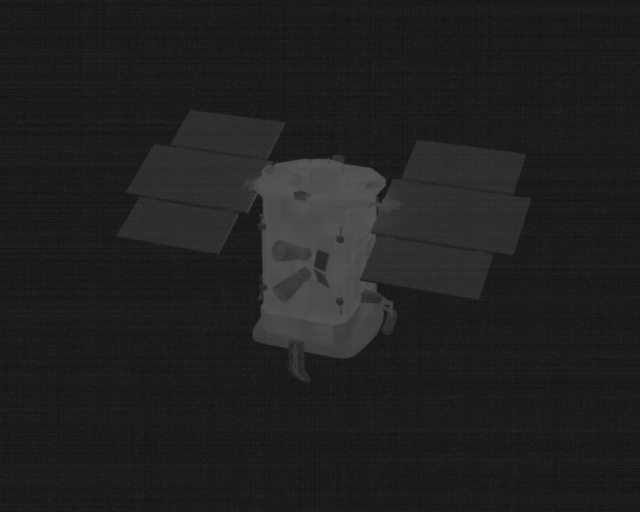}{Frame 924} \\[6pt]
    \imgcell{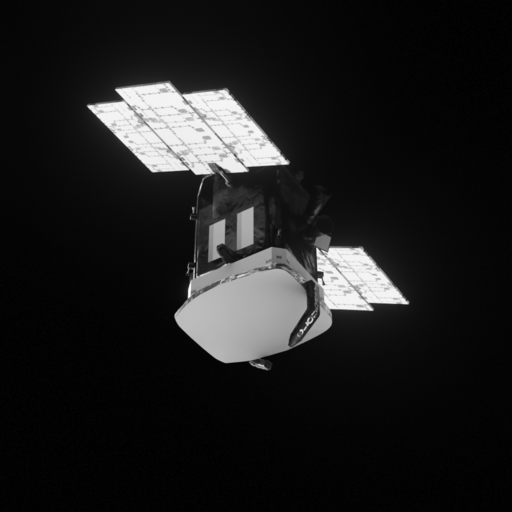}{Frame 619} &
    \imgcell{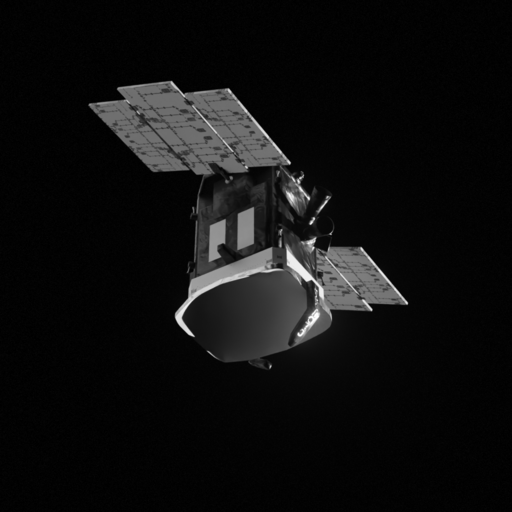}{Frame 1044} &
    \imgcell{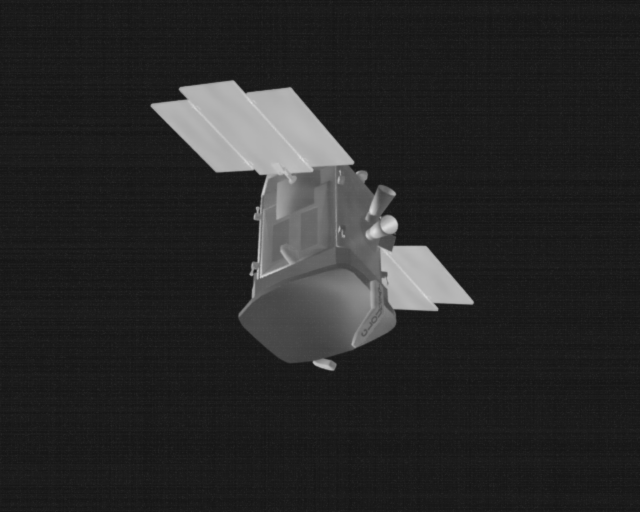}{Frame 619} &
    \imgcell{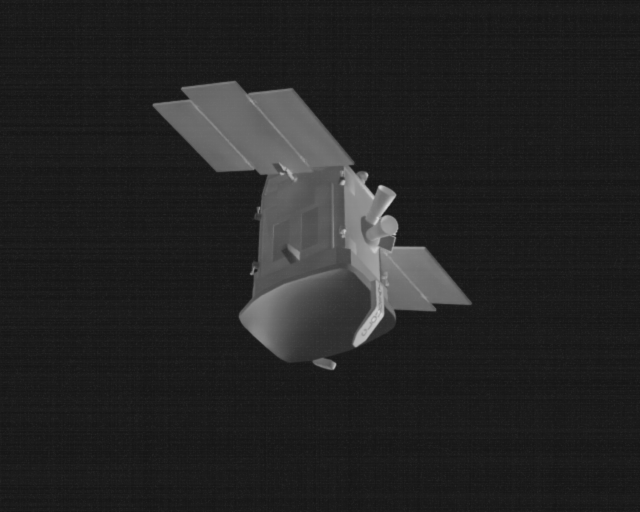}{Frame 1044} \\[6pt]
    \imgcell{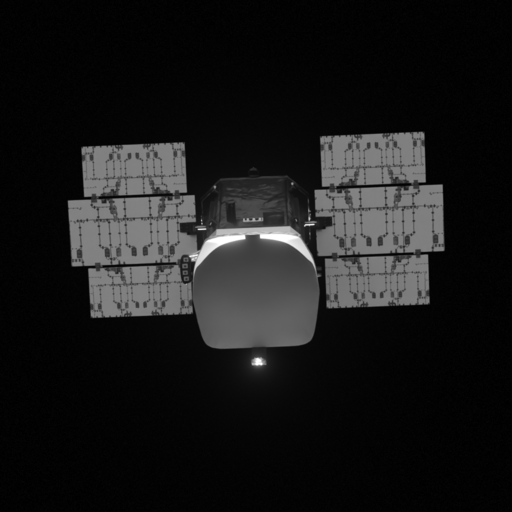}{Frame 0} &
    \imgcell{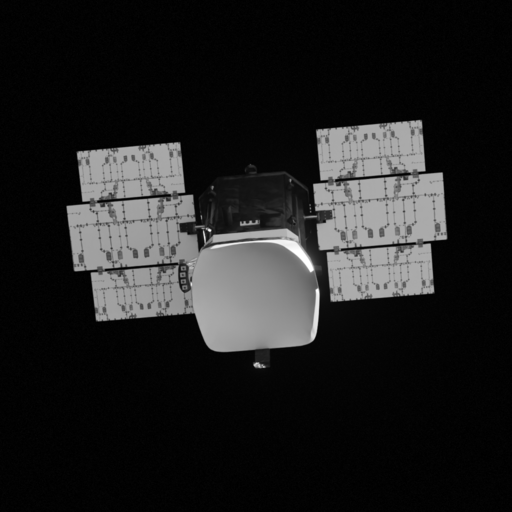}{Frame 1133} &
    \imgcell{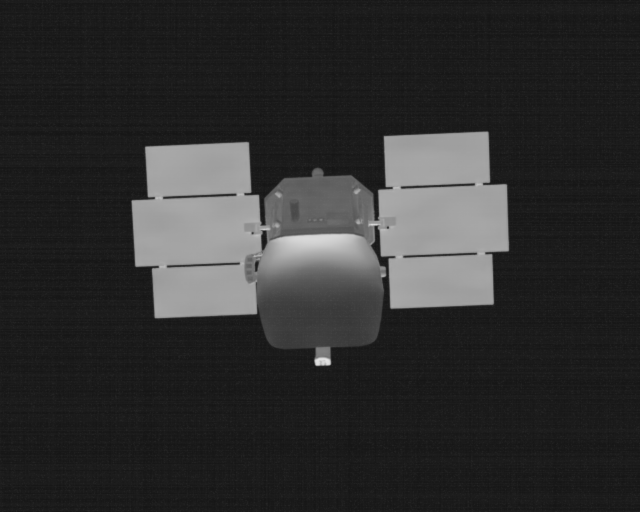}{Frame 0} &
    \imgcell{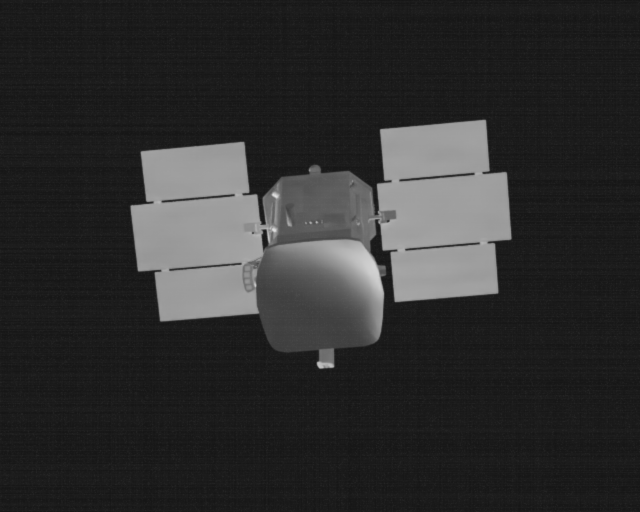}{Frame 1133} \\[6pt]
    \imgcell{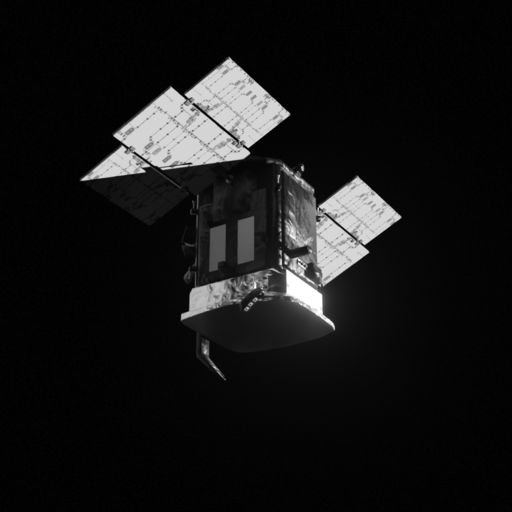}{Frame 462} &
    \imgcell{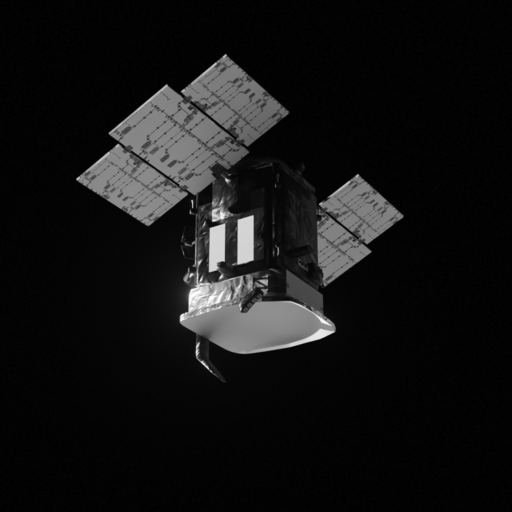}{Frame 1201} &
    \imgcell{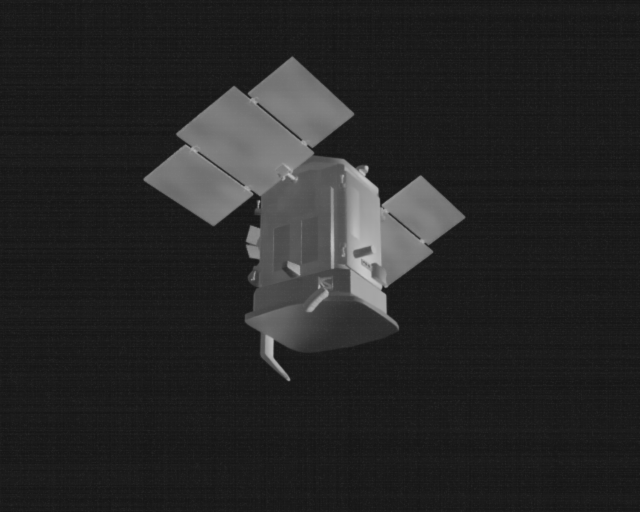}{Frame 462} &
    \imgcell{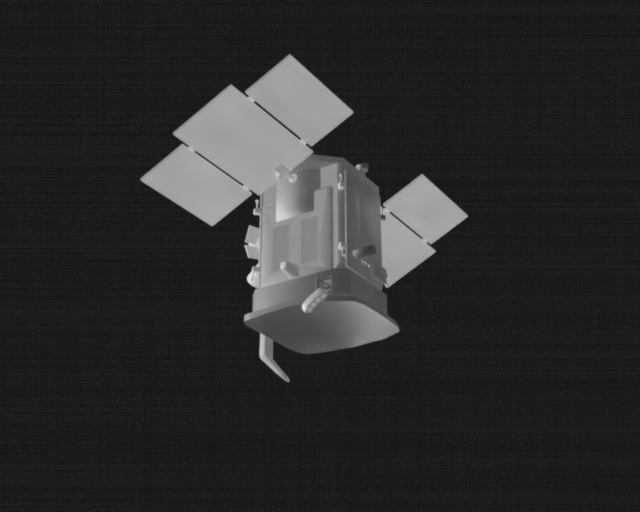}{Frame 1201} \\[6pt]
    \bottomrule
\end{tabular}
\hspace{10pt}%
  \begin{tabular}{c@{\hspace{2pt}}c@{\hspace{10pt}}c@{\hspace{2pt}}c}
    \toprule
    \multicolumn{2}{c}{\textbf{VIS}} & \multicolumn{2}{c}{\textbf{TIR}} \\
    \cmidrule(lr){1-2} \cmidrule(lr){3-4}
    \imgcell{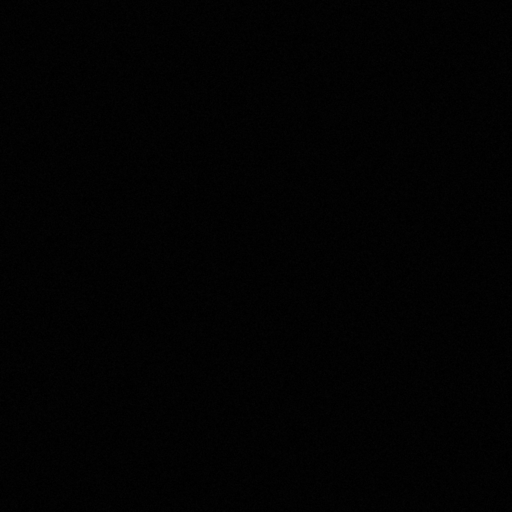}{Frame 896} &
    \imgcell{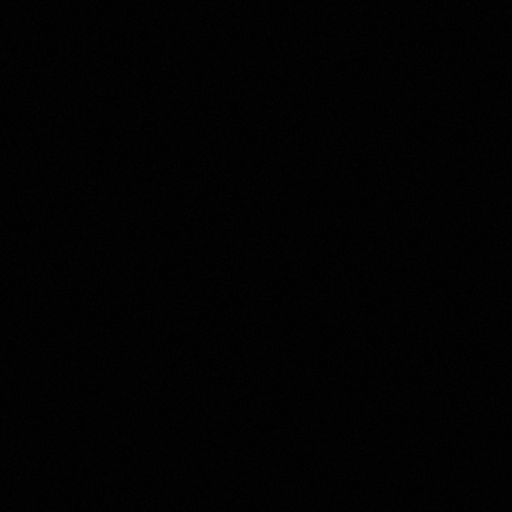}{Frame 1322} &
    \imgcell{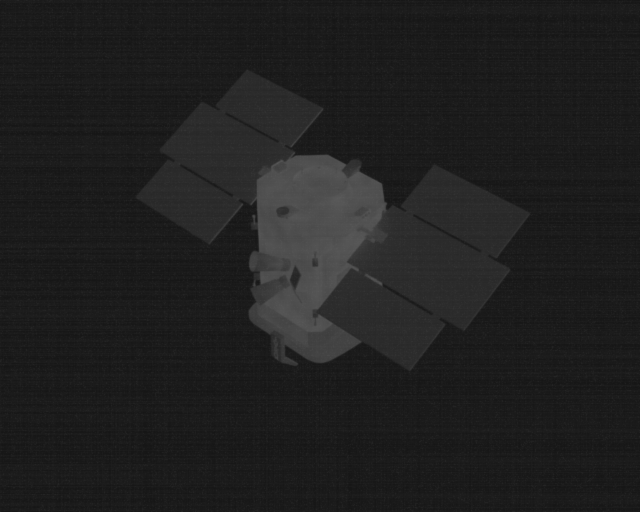}{Frame 896} &
    \imgcell{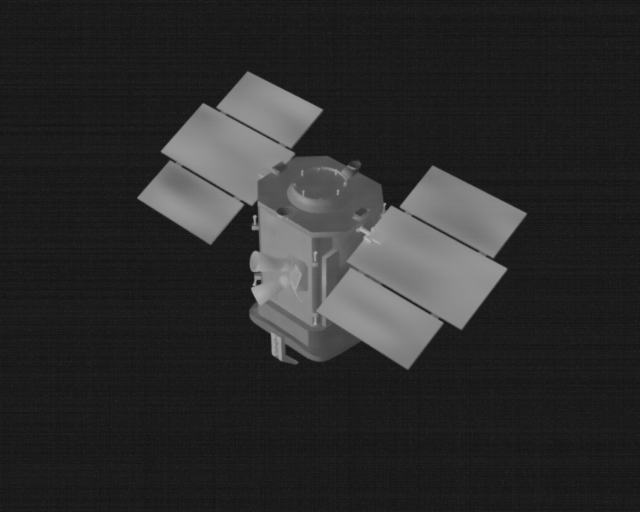}{Frame 1322} \\[6pt]
    \imgcell{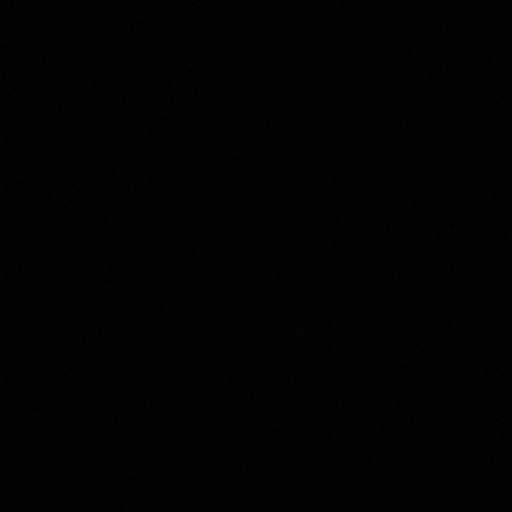}{Frame 263} &
    \imgcell{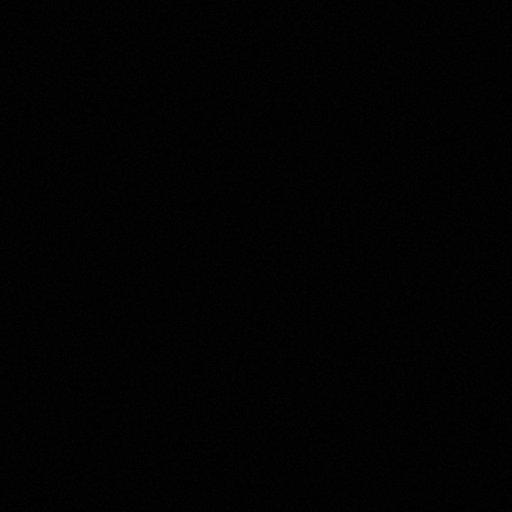}{Frame 1401} &
    \imgcell{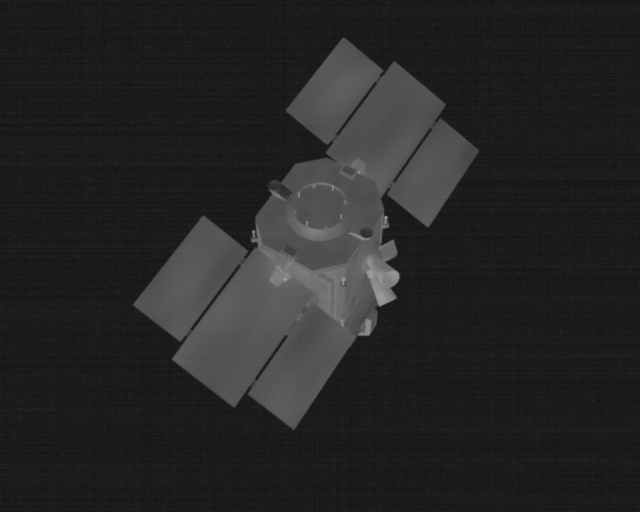}{Frame 263} &
    \imgcell{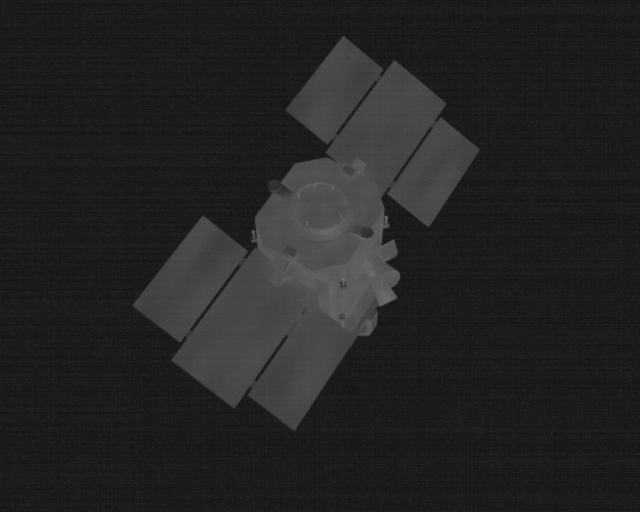}{Frame 1401} \\[6pt]
    \imgcell{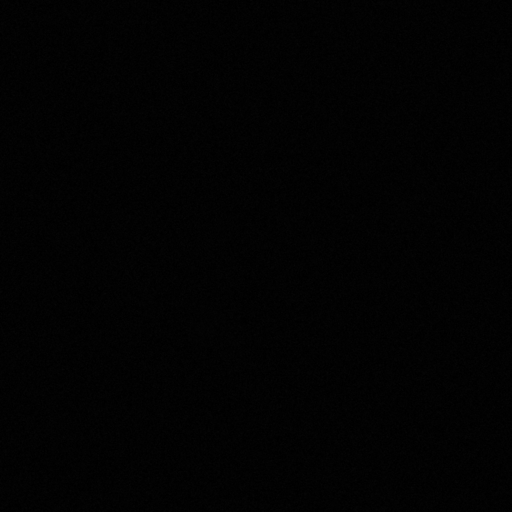}{Frame 739} &
    \imgcell{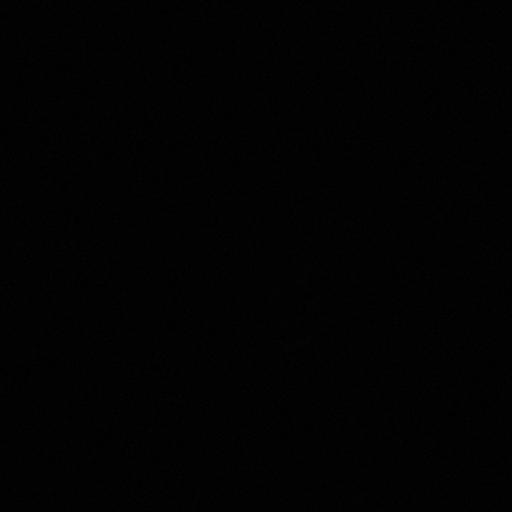}{Frame 1479} &
    \imgcell{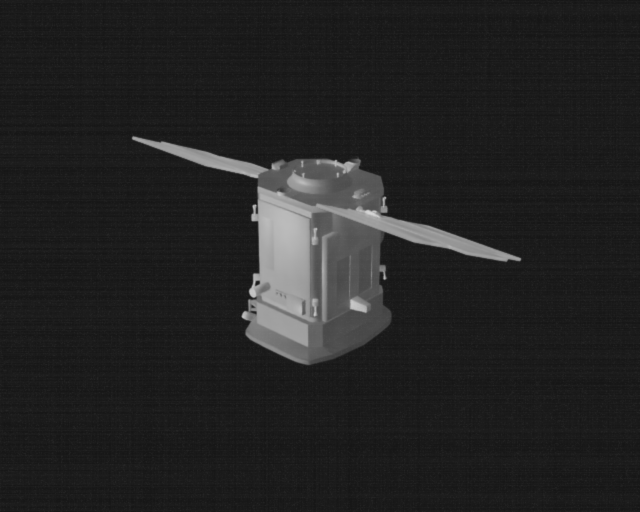}{Frame 739} &
    \imgcell{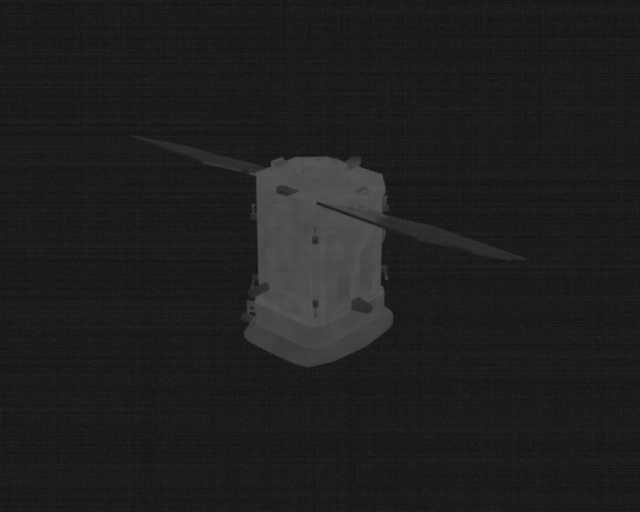}{Frame 1479} \\[6pt]
    \imgcell{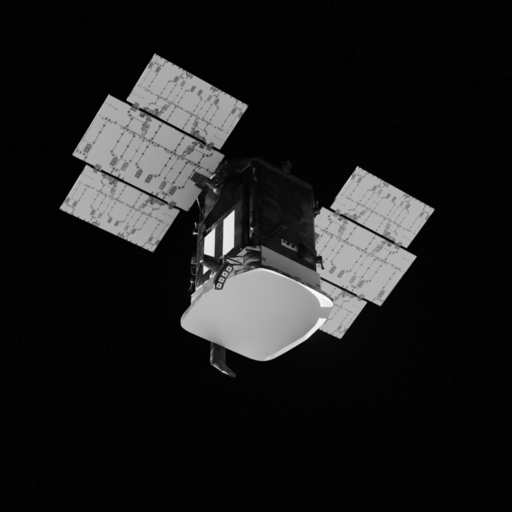}{Frame 1173} &
    \imgcell{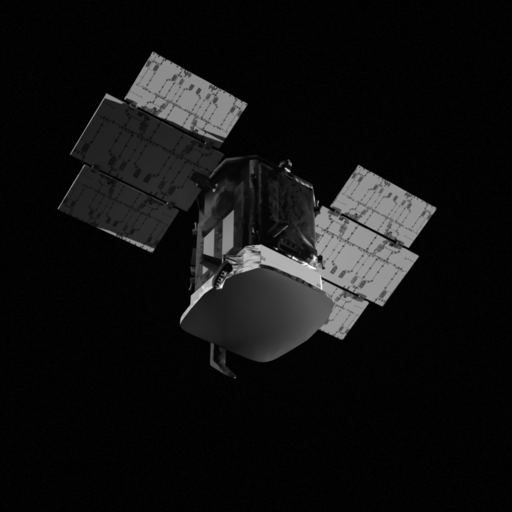}{Frame 1597} &
    \imgcell{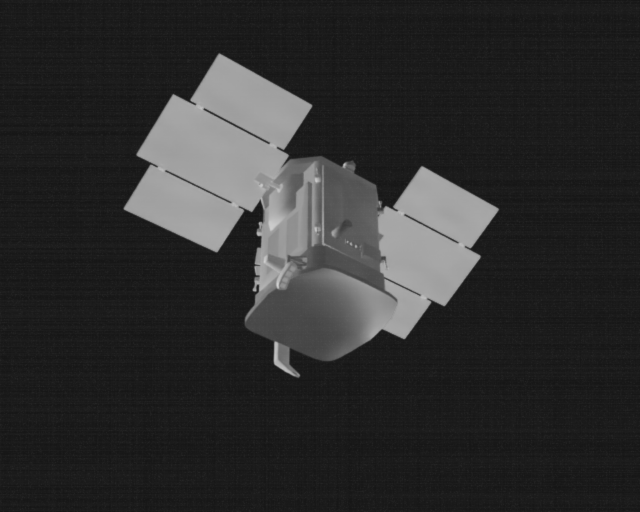}{Frame 1173} &
    \imgcell{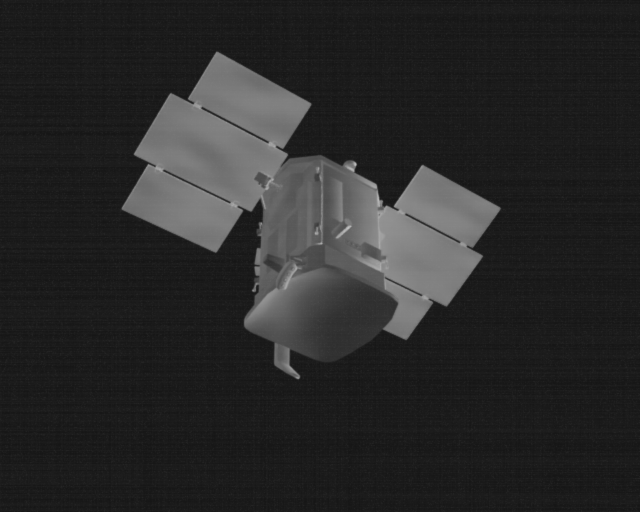}{Frame 1597} \\[6pt]
    \imgcell{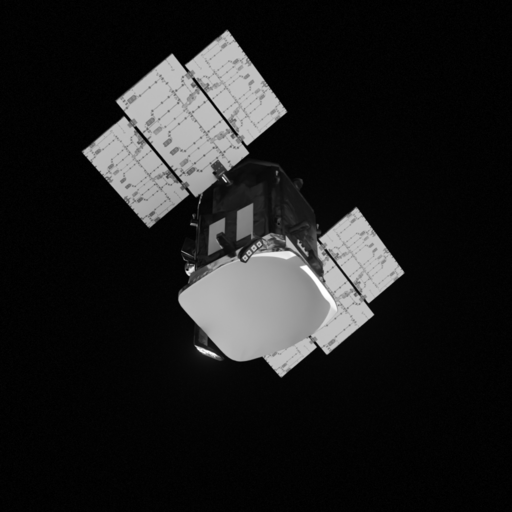}{Frame 43} &
    \imgcell{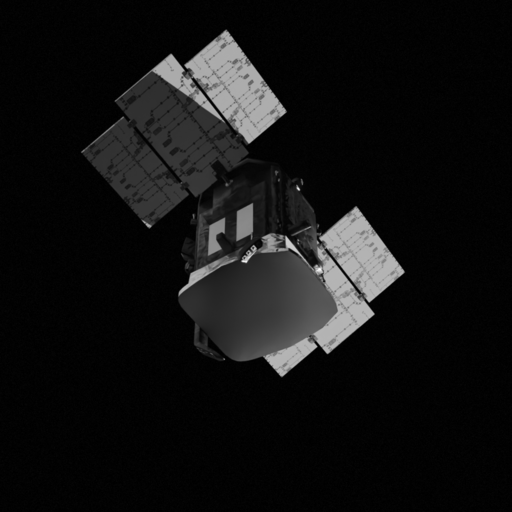}{Frame 1619} &
    \imgcell{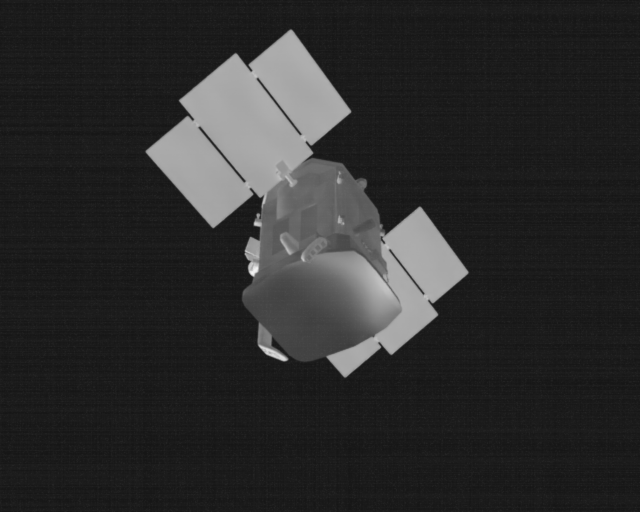}{Frame 43} &
    \imgcell{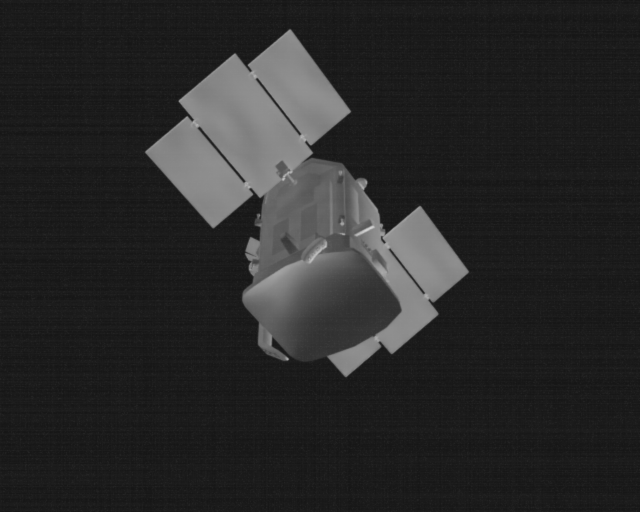}{Frame 1619} \\[6pt]
    \imgcell{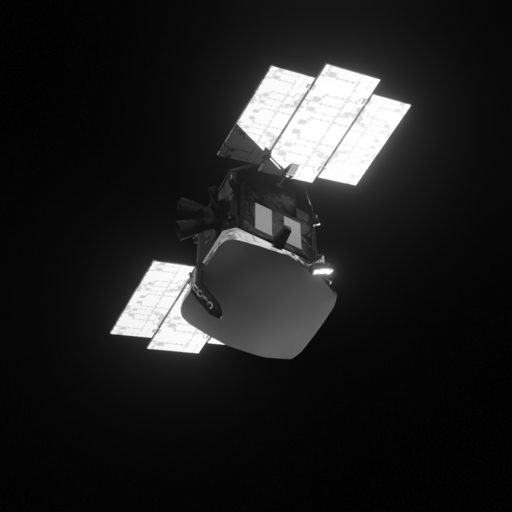}{Frame 528} &
    \imgcell{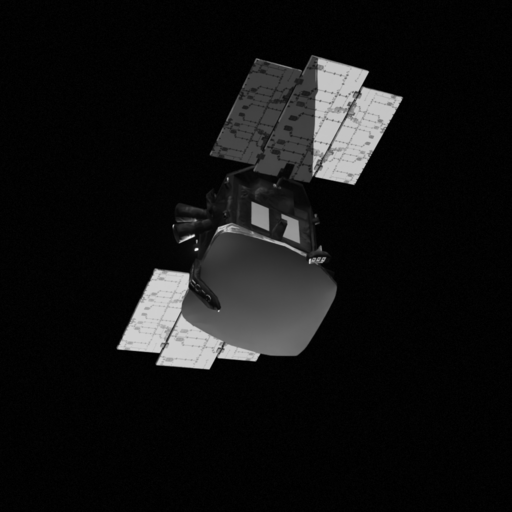}{Frame 1661} &
    \imgcell{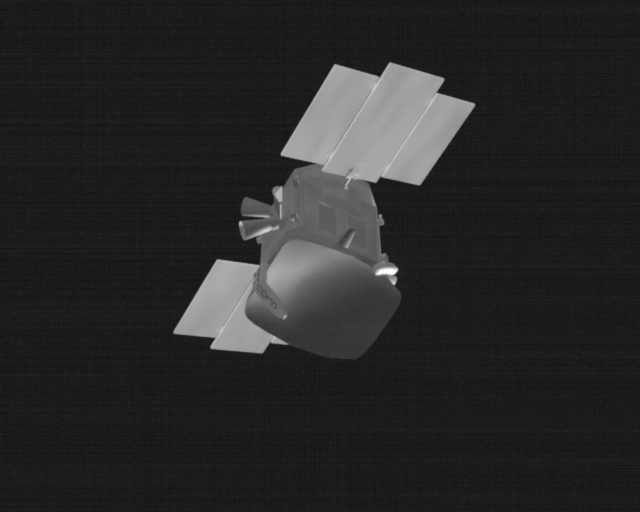}{Frame 528} &
    \imgcell{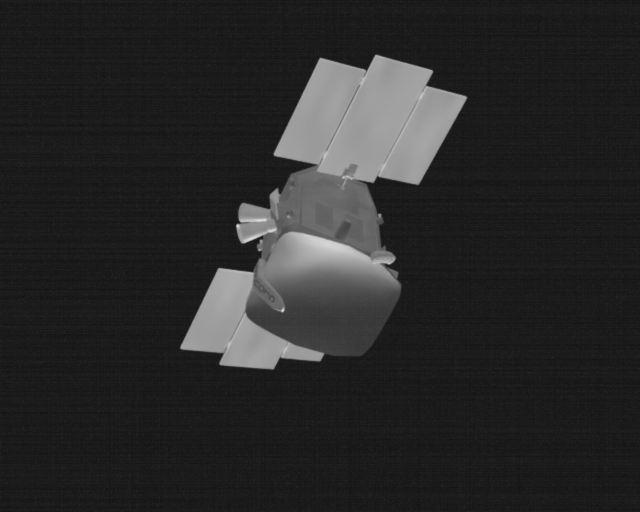}{Frame 1661} \\
    \bottomrule
  \end{tabular}
  \caption{Pairs of frames from each of the 12 loop closure opportunities in both visible and thermal-infrared bands of Scenario 3. The frames from the visible camera are often in eclipse. Drastic illumination/temperature changes are also seen between the frames in each pair in both visible and thermal-infrared.}
  \label{fig:lc_opportunities}
\end{figure}

%% file: 6_results.tex
\section{Results} \label{sec:results}
\subsection{Key Assumptions}
Before presenting the performance of our work, we summarize the key assumptions that are made which prevent this system from being in a flight-ready state.

\paragraph{Inspector Pointing.} The boresight of the camera is assumed to be perfectly aligned with the center of the target, with the target vertically aligned in the image frame throughout every scenario.
\paragraph{Relative Measurements.} Noisy acceleration and angular velocity measurements in the inspector body frame relative to the target body frame are available, substituting orbital dynamics priors.
\paragraph{Background Clutter.} No background clutter, such as Earth, is present in any of the imagery.
\paragraph{Priors.} An initial position, attitude, and velocity of the inspector relative to the target are roughly known to initialize the estimator.
\paragraph{Cameras.} The visible and thermal-infrared cameras are assumed to be perfectly collocated. The camera intrinsics and inspector-to-camera transformation are known exactly.

\subsection{Place Recognition Performance} \label{sec:vpr_performance}
The performance of the VPR method before pose recovery is first investigated on thermal-infrared imagery. The analysis in this section is focused on Scenario 3, as the relatively sparse 12 opportunities for place recognition provide a challenging case. Performance of various parameter configurations of our VPR method is compared, as well as the use of visible, thermal, or fused imagery. Our method is then compared against existing baselines.

\begin{figure}[htp]
    \centering
    \includegraphics[width=0.3\linewidth]{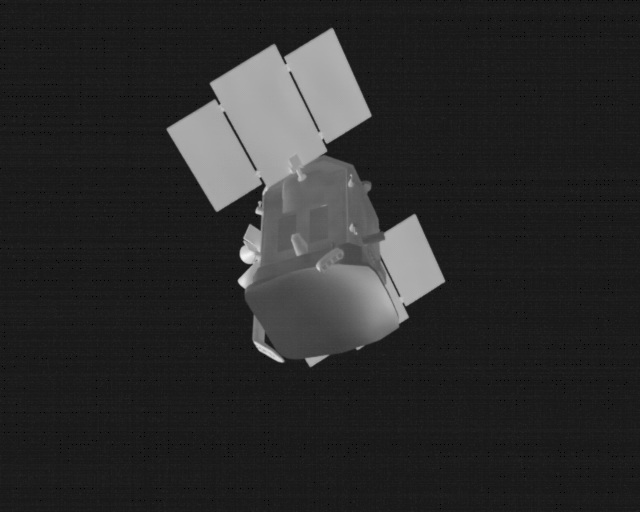}
    \includegraphics[width=0.3\linewidth]{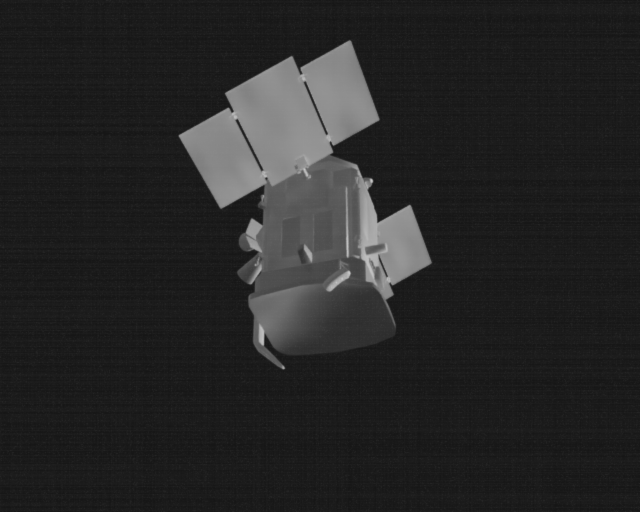}
    \caption{An example of two frames with a 10 degree angle between relative position vectors. These are taken from Scenario 3 in the infrared band. This difference in view is small enough that similar sides and components are visible on the target, but not large enough that unique parts of the target's surface are seen in each image.}
    \label{fig:10degExample}
\end{figure}

To analyze the performance of place recognition for this scenario, we compare the ground truth opportunities to the detected matches. Ground truth place recognition opportunities are identified when the inspector's position vector is within 10 degrees of any past position vector, ignoring any recent positions. This results in ground truth frame pairs, all of which have a perspective falling within 10 degrees of each other. The value of 10 degrees was chosen after testing multiple thresholds. A lower threshold would cause similar views of the target to be assigned as negative matches; any higher and two considerably different views of the target would be assigned as a positive match. The 10 degree threshold we choose is illustrated in \autoref{fig:10degExample}.

True positive matches are pairs of frames detected by the VPR method that are in the ground truth set. False positive matches are those detected that are not in the ground truth. Precision and recall are found according to the equations:
\begin{align}
    \text{Precision} = \frac{TP}{TP + FP} && \text{Recall} = \frac{TP}{GT}
\end{align}
where $TP$ is the number of true positive matches, $FP$ is the number of false positive matches, and $GT$ is the number of ground truth matches. In other words, precision is the proportion of detected matches that are correct. Recall is the proportion of the ground truth matches that are correctly detected.

\begin{figure}[tp]
    \centering
    \subcaptionbox{Precision-recall curves comparing the performance of our silhouette-based VPR method dependent on parameter selection. All curves shown use thermal imagery. Interpolated precision is used to improve readability, where the precision at a certain recall $r$ is the highest precision found for any recall $r'>r$.\label{fig:tir_ours_parameters_pr_curves}}[0.55\textwidth]{
    \includegraphics[width=\linewidth]{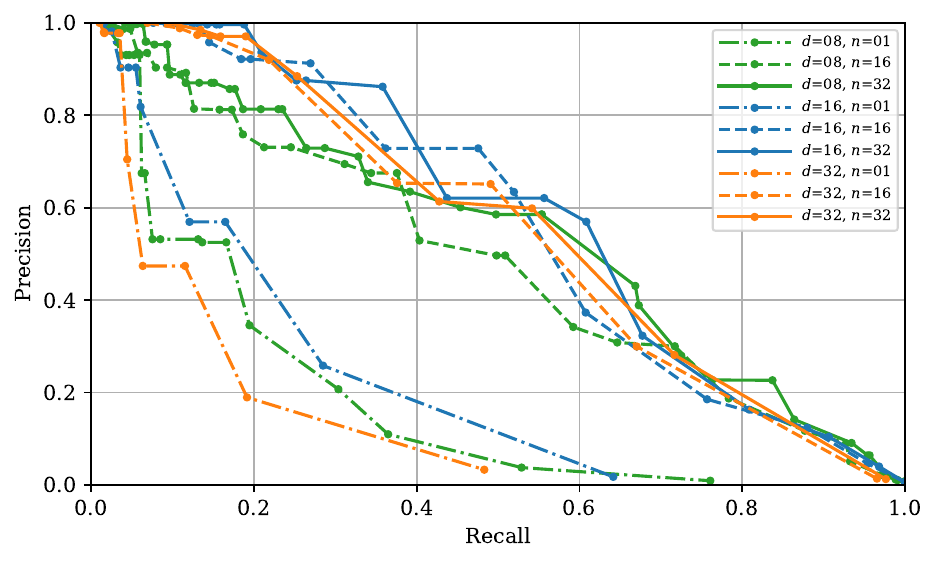}
    } 
    \hfill
    \subcaptionbox{Precision-recall curves comparing the performance of our silhouette-based VPR method dependent on sensor modality, with $d=16$ and $n=16$.\label{fig:sensors_pr_curves}}[0.55\textwidth]{
    \includegraphics[width=\linewidth]{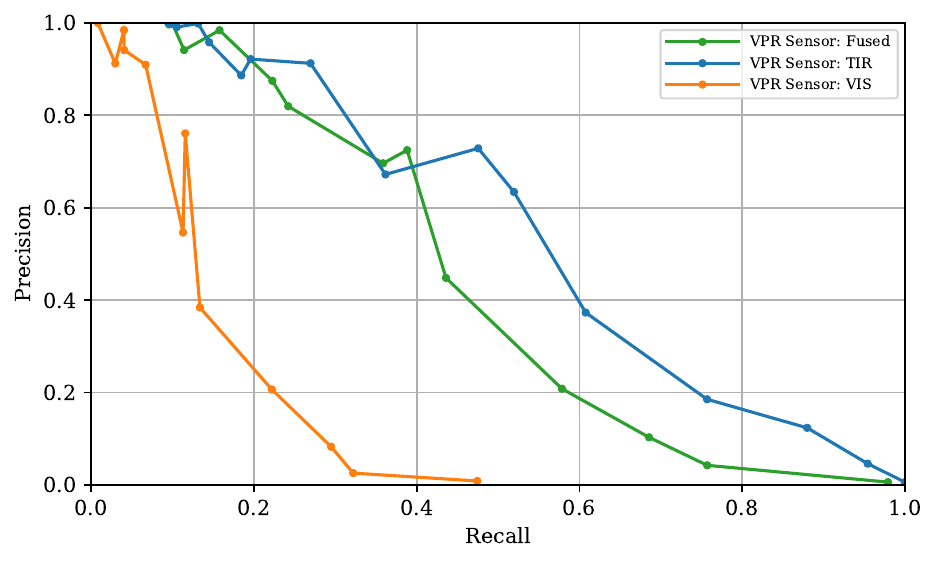}
    }
    \subcaptionbox{Precision-recall curves comparing the performance of our silhouette-based VPR method against existing methods, all run on TIR imagery.\label{fig:oursAndBaseline_pr_curves}}[0.55\textwidth]{
    \includegraphics[width=\linewidth]{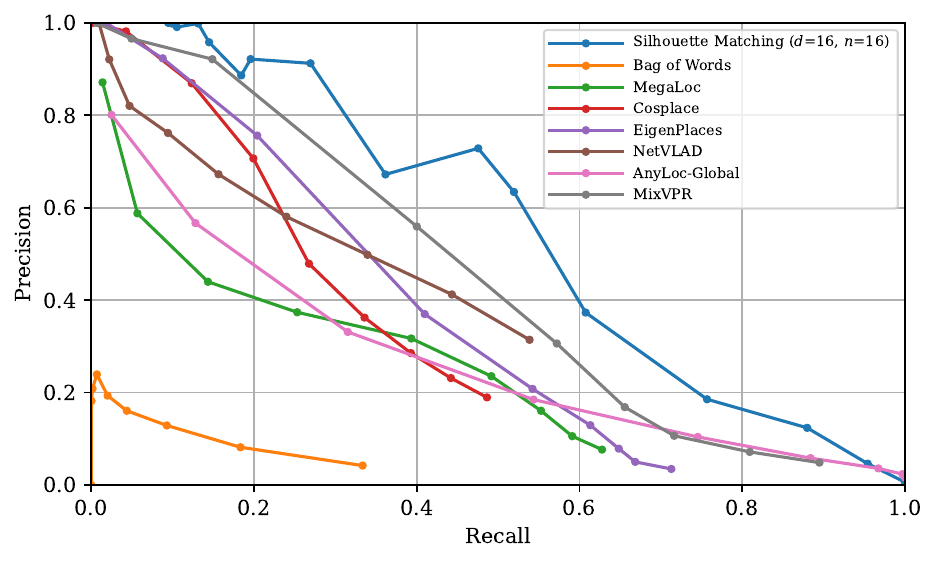}
    }
    \caption{Precision-recall curves comparing the performance of our silhouette-based VPR method run on Scenario 3 using various parameter values (a), sensor modalities (b), and compared to baseline VPR methods (c).}
\end{figure}

First, a range of parameters is run for our silhouette-based VPR method to compare performance. As shown in \autoref{fig:tir_ours_parameters_pr_curves}, varying $b$ (the number of bins for each radial distance) affects the sensitivity of the silhouette matching. A smaller value of $b$ represents fewer bins to fall into, which increases the probability of descriptor collisions, therefore increasing the number of matches. This results in more true matches which increases recall, however this also results in more false matches, which decreases precision. When $b$ is increased, the silhouettes must be more similar to be detected as a match. This decreases recall and increases precision.

To create a precision-recall curve, our method is run with $b$ varied over a range of values and the precision and recall are found for each run, forming a curve. This is repeated for multiple values of $d$ and $n$, where each combination results in a unique precision-recall curve, all shown in \autoref{fig:tir_ours_parameters_pr_curves}. Our method uses thermal imagery for silhouette matching unless otherwise noted.

The first pattern to notice is that when $n$ is increased, the precision-recall curve for any given $d$ shifts up and to the right, which indicates better performance. $n$ represents the number of descriptors found per image, with each descriptor using a slightly rotated set of points on the silhouette. Additional descriptors provide more opportunities for silhouettes to be matched with one another when there are small variations in the silhouette, dramatically improving precision and recall performance. While the notion of finding multiple descriptors per image is helpful, there are quickly diminishing returns. \autoref{fig:tir_ours_parameters_pr_curves} shows that increasing $n$ from 1 to 16 can more than double the precision in many cases; however, the increase from 16 to 32 has a far weaker effect on the algorithm performance. Additionally, a larger $n$ results in longer runtime.

A more nuanced relationship is seen with $d$, the descriptor length representing the number of points on the silhouette boundary used to describe it. Performance improves going from $d=8$ to $d=16$, but then decreases when switching to $d=32$. When $d$ is set too low, the shape of the silhouettes is not able to be described with enough resolution, causing the matching to be imprecise. However if $d$ is set too high, the descriptors become affected by the noise/expected variation of the silhouettes, resulting in a drop in performance. Therefore, $d$ must be chosen high enough to accurately describe the shape of the silhouette, but not so high as to be susceptible to small variations in their shape.

The next analysis compares our method across thermal, visible, and fused imagery, with the results in \autoref{fig:sensors_pr_curves}. The thresholding value used to obtain the silhouette mask is tuned to separate the target spacecraft from the background as well as possible for each type of sensor used. Visible imagery performs the worst by far. This deficiency arises mostly because the spacecraft is not visible at all during eclipse for much of the scenario. However, a less obvious issue is the glare and bloom from solar reflections. These effects cause the silhouette to be inaccurately drawn, corrupting the performance of the matching algorithm.

The fused imagery incorporates information from both the visible and thermal camera. As a result, it contains the same glare and bloom issues as the visible camera. Although this hinders the silhouette matching of the fused imagery, the thermal camera is able to fill in the shadowed and eclipsed regions of the trajectory, substantially improving the performance over the visible camera alone.

The thermal images exhibit the best performance. Compared to the visible camera, the thermal camera is robust to eclipse. However, during the end of the eclipse periods, the brightness of the spacecraft in the thermal imagery approaches the noise floor of the background of space. This creates some noise in the threshold mask. Our method is able to remain robust to this noise, as the descriptor computation method finds the edge of the silhouette as the \emph{farthest} point in the mask. Therefore, most noise within the threshold mask is ignored when calculating the descriptor.

The performance of our silhouette matching method is also compared to baseline methods, as shown in \autoref{fig:oursAndBaseline_pr_curves}. For each method, the threshold for a positive descriptor match is varied. Each threshold level has an associated precision and recall value, which are plotted and connected to form a precision-recall curve. The precision-recall curves for each existing VPR method are calculated, using the simulated thermal-infrared imagery of Scenario 3. Bag of Words performs the worst by far; however, even the learned methods struggle, illustrated by the rapid decrease of precision even at low recall values. Ideally, the precision would stay close to 1 for a large range of small recall values, which would represent a large number of correct frame matches while avoiding false positive matches.

Comparing our VPR method against the baselines, we see a clear improvement. Our method is able to maintain a precision close to 100\% at a recall of 15\%, while the best of the learning-based methods struggle to stay above 90\% precision. It is this top left region of the precision-recall plot that is the most relevant to SLAM, as any false positive loop closure detections should be avoided. Our method is able to avoid these false positives while detecting far more matches than any of the baseline methods.
We also compare the timing of the VPR methods, detailed in \autoref{tab:vpr_timing}. This analysis was run on an Intel i9-10920X CPU and two NVIDIA RTX 3090 GPUs. Our method has CPU runtime faster than that of all VPR methods we compare against. While the learning-based methods are designed to be run on GPU, our method is able to run quickly on a CPU-only system.

\begin{table}[tp]
    \centering
    \caption[Runtime of VPR methods]{Runtime of VPR methods on the thermal-infrared imagery of Scenario 3. Silhouette Matching and Bag of Words are not designed for GPU, so no timing results are available for those cases.}
    \begin{tabular}{l S[table-format=4.1] S[table-format=3.1]}
    \toprule
    VPR Method & \multicolumn{1}{c}{CPU Runtime, ms} & \multicolumn{1}{c}{GPU Runtime, ms} \\
    \midrule
    Silhouette Matching ($d=16, n=16$)          & 23.0      & N/A \\
    Bag of Words*~\cite{lopez_bags_2012}        & 65.0      & N/A \\
    NetVLAD~\cite{arandjelovic_netvlad_2016}    & 432.5     & 12.9 \\
    CosPlace~\cite{Berton_CosPlace_2022}        & 210.6     & 8.3  \\
    MixVPR~\cite{ali_mixvpr_2023}               & 61.1      & 8.7 \\
    EigenPlaces~\cite{Berton_EigenPlaces_2023}  & 164.5     & 54.0 \\
    AnyLoc-Global~\cite{keetha_anyloc_2023}     & 350.8     & 345.9 \\
    MegaLoc~\cite{Berton_MegaLoc_2025}          & 1378.0    & 8.5 \\
     \bottomrule
     \multicolumn{3}{l}{\footnotesize{*All methods run in Python except Bag of Words, which uses C++}}
    \end{tabular}
    \label{tab:vpr_timing}
\end{table}

\subsection{Integrated Navigation Performance} \label{subsec:integrated_performance}
Our final analysis evaluates the integrated navigation performance using our silhouette-based VPR method, comparing a full SLAM solution along three axes: (1) the six simulated scenarios, (2) sensor modality (VIS, TIR, fused), and (3) loop closure method (without loop closures, using Bag of Words, and using our silhouette-based method). For each frame, the place recognition is run to detect any matches. If a match is detected, pose recovery is performed to estimate the transformation between the matching frames. The loop closures from pose recovery are added to a pose graph created using Kimera-RPGO, with Kimera-VIO providing odometry and outlier loop closures removed. Unlike the preceding analyses, we keep only one loop closure per query frame to avoid the accumulation of a significant number of loop closures, which would increase runtime unreasonably. In the case of multiple matches for Bag of Words, the matched frame with the closest descriptor is used. For our silhouette matching method, the matching frame furthest in time from the current frame is used. We compare navigation performance using incremental, real-time navigation estimates.
We use silhouette matching parameters $d=16$, $n=16$, and $b=28$ for Scenarios 1--5. Since the CubeSat is far more symmetric than CloudSat, the parameters are adjusted for Scenario 6 such that the place recognition is more selective. Specifically, $b$ is changed to 64.

The tracking/mapping success rate and the inlier feature match statistics shown in \autoref{tab:trackingSuccessAndAvgFeatures} illustrate the advantages of using fused imagery in the front end of SLAM. The tracking/mapping success rates using the visible camera fail to exceed 64\% (other than when the target is fully illuminated) due to eclipse. Similarly, the minimum inlier match count is 0 for most scenarios; however, the maximum match count is often higher than that of the other sensor modalities due to the high resolution and texture found in the visible camera. The thermal imagery has improved tracking success and has the largest minimum inlier matches due to its illumination-robustness, but struggles to achieve a good upper bound of feature matches due to its lack of feature-richness. The fused imagery incorporates the best aspects of each sensor, reaching 100\% success in nearly every scenario, as well as the highest average inlier match counts.

The performance of the visual place recognition methods is indicated in \autoref{tab:lcc}. Our silhouette-based method using the thermal imagery clearly has superior performance indicated by the larger total and inlier loop closure detections than Bag of Words using any sensor type. In Scenarios 3 and 6 where only 12 loop closure opportunities exist, each opportunity spans many frames. The VPR method then detects each matching pair of frames during these opportunities. All other scenarios have the inspector retracing the same relative trajectory, resulting in many more detected loop closures. Even in Scenario 6, which offers extremely difficult conditions for place recognition due to the symmetric nature of the target, 41 inlier loop closures are detected using our method.

\Cref{tab:avgPosErr,tab:maxPosErr,tab:avgRotErr,tab:maxRotErr} show the average and maximum position and rotation errors over each run. Using only VIO without loop closures, the fused imagery often has substantially lower errors than visible- and thermal-only. The biggest exception is Scenario 2, where using only visible performs better due to the fully illuminated target. Introducing loop closures using Bag of Words for place recognition decreases errors in many runs where inlier loop closures are found. Replacing Bag of Words with our silhouette matching decreases almost all errors even further, caused by the larger number of loop closures. The setup with the lowest navigation errors uses fused imagery for VIO and our silhouette-based place recognition using the thermal imagery. This allows the feature tracking to use the high resolution and texture from the visible camera combined with the illumination-robustness of the thermal camera, as well as robust silhouette matching using the constantly-visible outline of the target from the thermal camera.

\newcommand{\sI}{S1: CloudSat 15 m}
\newcommand{\sII}{S2: CloudSat 15 m, Fully Lit}
\newcommand{\sIII}{S3: CloudSat 15 m, Rotating}
\newcommand{\sIV}{S4: CloudSat 15--25 m}
\newcommand{\sV}{S5: CloudSat 50 m}
\newcommand{\sVI}{S6: CubeSat 15 m, Rotating}

\newcommand{\colSpace}{\hspace{0.5em}}

\newcolumntype{C}{>{\centering\arraybackslash}X}
\newcommand{\ph}[1]{\phantom{#1}}
\newcommand{\cmultiline}[1]{\multicolumn{1}{c}{\addstackgap{\clap{\shortstack{#1}}}}}

\begin{table}[hptb]
\centering
\caption{Tracking/mapping success rate and minimum, average, and maximum inlier feature matches per frame over each run. The success rate is calculated by counting the number of frames in which valid features are detected and mapped as landmarks, divided by the total number of frames.}
\small
\setlength{\tabcolsep}{3pt}

\newcolumntype{Y}{>{\hsize=0.75\hsize\linewidth=\hsize\centering\arraybackslash}X} 
\newcolumntype{Z}{>{\hsize=1.25\hsize\linewidth=\hsize\centering\arraybackslash}X} 

\begin{tabularx}{\linewidth}{l !{\colSpace} *{3}{Y} !{\colSpace}  *{3}{Z}}
\toprule
& \multicolumn{3}{c}{Tracking/Mapping Success Rate} & \multicolumn{3}{c}{Inlier Feature Matches (Min / Avg / Max)} \\
\cmidrule(lr){2-4} \cmidrule(lr){5-7}
Scenario & VIS & TIR & Fused & VIS & TIR & Fused \\
\cmidrule(r){1-1} \cmidrule(lr){2-4} \cmidrule(lr){5-7}
\sI     & \ph{0}\underline{64.0\%} & \textbf{100.0\%} & \textbf{100.0\%}
        & \ph{0}0 / \ph{0}\underline{89} / \underline{272} & \textbf{15} / \ph{0}63 / 133 & \ph{0}\underline{9} / \textbf{120} / \textbf{278} \\
\sII    & \textbf{100.0\%} & \textbf{100.0\%} & \textbf{100.0\%}
        & \textbf{44} / \textbf{169} / \textbf{289} & 15 / \ph{0}\underline{63} / 133 & \underline{36} / \textbf{169} / \underline{288} \\
\sIII   & \ph{0}\underline{63.9\%} & \textbf{100.0\%} & \textbf{100.0\%}
        & \ph{0}0 / \ph{0}\underline{71} / \textbf{269} & \textbf{13} / \ph{0}47 / \ph{0}88 & \ph{0}\underline{9} / \ph{0}\textbf{79} / \underline{249} \\
\sIV    & \ph{0}\underline{63.9\%} & \textbf{100.0\%} & \textbf{100.0\%}
        & \ph{0}0 / \ph{0}\underline{65} / \textbf{285} & \textbf{17} / \ph{0}61 / 169 & \underline{11} / \ph{0}\textbf{77} / \underline{278} \\
\sV     & \ph{0}63.2\% & \ph{0}\underline{99.7\%} & \textbf{100.0\%}
        & \ph{0}0 / \ph{0}93 / \textbf{292} & \textbf{43} / \underline{101} / 271 & \underline{15} / \textbf{112} / \underline{286} \\
\sVI    & \ph{0}\underline{33.3\%} & \ph{0}16.0\% & \ph{0}\textbf{96.7\%}
        & \ph{0}\textbf{0} / \ph{0}\underline{17} / \ph{0}69 & \ph{0}\textbf{0} / \ph{0}10 / \ph{0}\underline{70} & \ph{0}\textbf{0} / \ph{0}\textbf{30} / \textbf{167} \\
\bottomrule
\multicolumn{7}{l}{\footnotesize
    \addstackgap{
        \begin{minipage}{0.6\linewidth}
        \textbf{Bold} values are the highest for each scenario. \\
        \underline{Underlined} values are the second highest for each scenario.
      \end{minipage}
}}
\end{tabularx}
\label{tab:trackingSuccessAndAvgFeatures}
\end{table}

\begin{table}[hptb]
\centering
\caption{Number of loop closures in each run, reported as inliers / total. Each loop closure represents a detected instance when an earlier view of the target spacecraft is revisited. Scenarios 3 and 6 only consist of 12 loop closure opportunities, each made up of many frames that match a previous frame. All other scenarios have the inspector retracing its relative trajectory, resulting in many more loop closure opportunities. These values are after geometric verification. Inlier loop closures are those found to be consistent with VIO estimates using GNC~\cite{yang_graduated_2020}.}
\small

\newcolumntype{Y}{>{\hsize=0.857\hsize\linewidth=\hsize\centering\arraybackslash}X}
\newcolumntype{Z}{>{\hsize=1.429\hsize\linewidth=\hsize\centering\arraybackslash}X}

\begin{tabularx}{0.9\linewidth}{l !{\colSpace} *{3}{Y} !{\colSpace}  Z}
\toprule
& \multicolumn{3}{c}{Bag of Words LCs} & \cmultiline{Silhouette-Based LCs} \\
\cmidrule(lr){2-4} \cmidrule(lr){5-5}
Scenario & VIS & TIR & Fused & TIR \\
\cmidrule(r){1-1} \cmidrule(lr){2-4} \cmidrule(lr){5-5}
\sI     & 116 / 147           & 23 / 40                           & \underline{233} / \underline{270}          & \textbf{1044} / \textbf{1044}              \\
\sII    & 400 / 465           & 23 / 40                           & \underline{401} / \underline{472}          & \textbf{1044} / \textbf{1044}              \\
\sIII   & \ph{00}0 / \ph{0}13 & \ph{0}\underline{3} / \underline{15} & \ph{00}1 / \ph{00}4                     & \ph{00}\textbf{61} / \ph{00}\textbf{61}    \\
\sIV    & \ph{0}\underline{62} / \ph{0}\underline{86} & 19 / 53           & \ph{0}32 / \ph{0}42                     & \textbf{1073} / \textbf{1073}              \\
\sV     & \ph{0}73 / \ph{0}97 & \ph{0}1 / \ph{0}7                 & \underline{187} / \underline{225}          & \ph{0}\textbf{971} / \textbf{1098}         \\
\sVI    & \ph{00}\underline{0} / \ph{00}0 & \ph{0}\underline{0} / \ph{0}0 & \ph{00}\underline{0} / \ph{00}\underline{6} & \ph{00}\textbf{41} / \ph{00}\textbf{93} \\
\bottomrule
\multicolumn{5}{l}{\footnotesize
    \addstackgap{
        \begin{minipage}{0.8\linewidth}
        Counts are reported as inliers / total detections.\\
        \textbf{Bold} counts are the highest inlier count for each scenario. \\
        \underline{Underlined} counts are the second highest inlier count for each scenario.
        \end{minipage}
}}
\end{tabularx}
\label{tab:lcc}
\end{table}

\begin{table}[hptb]
\centering
\caption{Average position errors (meters) over each run. Errors are between Kimera's estimate of the inspector's relative pose and the ground truth relative pose.}
\small
\begin{tabularx}{\linewidth}{l !{\colSpace} *{3}{C} !{\colSpace} *{3}{C} !{\colSpace} *{3}{C}}
\toprule
& \multicolumn{3}{c}{VIO} & \multicolumn{3}{c}{\shortstack{VIO +\\ Bag of Words LCs}} & \multicolumn{3}{c}{\shortstack{VIO +\\ TIR Silhouette-Based LCs}} \\
\cmidrule(lr){2-4} \cmidrule(lr){5-7} \cmidrule(lr){8-10}
Scenario &
VIS & TIR & Fused & VIS & TIR & Fused & VIS & TIR & Fused \\
\cmidrule(r){1-1} \cmidrule(lr){2-4} \cmidrule(lr){5-7} \cmidrule(lr){8-10}
\sI     & \ph{0}7.32            & \ph{0}25.93 & \ph{0}0.86 & \ph{0}4.90            & \ph{0}20.18 & \underline{0.79} & \ph{0}4.55            & \ph{00}8.98 & \textbf{0.65}    \\
\sII    & \ph{0}\underline{0.80} & \ph{0}25.93 & \ph{0}1.80 & \ph{0}0.97            & \ph{0}20.18 & 1.97             & \ph{0}\textbf{0.75}   & \ph{00}8.98 & 0.92             \\
\sIII   & \ph{0}5.78            & \ph{0}12.67 & \ph{0}5.21 & \ph{0}5.78            & \ph{0}12.67 & 5.20             & \ph{0}\textbf{5.03}   & \ph{00}9.45 & \underline{5.14} \\
\sIV    & \ph{0}7.55            & \ph{0}24.79 & \ph{0}3.05 & \ph{0}6.83            & \ph{0}15.63 & \underline{2.68} & \ph{0}6.51            & \ph{00}9.56 & \textbf{1.97}    \\
\sV     & 19.65                 & \ph{0}49.35 & 12.13      & 18.19                 & \ph{0}49.42 & \underline{5.94} & 16.32                 & \ph{0}50.50 & \textbf{5.70}    \\
\sVI    & 17.87                 & 181.09      & \ph{0}\underline{2.71} & 17.87     & 181.09      & \underline{2.71} & 17.68                 & 179.34      & \textbf{2.24}    \\
\bottomrule
\multicolumn{10}{l}{\footnotesize
    \addstackgap{
        \begin{minipage}{0.8\linewidth}
        \textbf{Bold} errors are the lowest for each scenario. \\
        \underline{Underlined} errors are the second lowest for each scenario.
      \end{minipage}
}}
\end{tabularx}
\label{tab:avgPosErr}
\end{table}

\begin{table}[hptb]
\centering
\caption{Maximum position errors (meters) over each run. Errors are between Kimera's estimate of the inspector's relative pose and the ground truth relative pose.}
\small
\begin{tabularx}{\linewidth}{l !{\colSpace} *{3}{C} !{\colSpace} *{3}{C} !{\colSpace} *{3}{C}}
\toprule
& \multicolumn{3}{c}{VIO} & \multicolumn{3}{c}{\shortstack{VIO +\\ Bag of Words LCs}} & \multicolumn{3}{c}{\shortstack{VIO +\\ TIR Silhouette-Based LCs}} \\
\cmidrule(lr){2-4} \cmidrule(lr){5-7} \cmidrule(lr){8-10}
Scenario &
VIS & TIR & Fused & VIS & TIR & Fused & VIS & TIR & Fused \\
\cmidrule(r){1-1} \cmidrule(lr){2-4} \cmidrule(lr){5-7} \cmidrule(lr){8-10}
\sI     & 15.86                 & \ph{0}57.10 & \ph{0}1.76             & 11.19             & \ph{0}48.13 & \ph{0}\underline{1.71} & \ph{0}9.78          & \ph{0}21.29 & \ph{0}\textbf{1.36}    \\
\sII    & \ph{0}\textbf{1.37}   & \ph{0}57.10 & \ph{0}3.34             & \ph{0}4.00        & \ph{0}48.13 & \ph{0}3.62             & \ph{0}1.69          & \ph{0}21.29 & \ph{0}\underline{1.63} \\
\sIII   & \underline{10.33}     & \ph{0}34.59 & 10.64                  & \underline{10.33} & \ph{0}36.16 & 11.53                  & \textbf{10.04}      & \ph{0}21.18 & 10.65                  \\
\sIV    & 23.05                 & \ph{0}53.85 & \ph{0}6.56             & 14.01             & \ph{0}37.76 & \ph{0}\underline{4.71} & 12.71               & \ph{0}20.09 & \ph{0}\textbf{3.46}    \\
\sV     & 59.60                 & 116.50      & 29.10                  & 55.92             & 117.06      & \underline{14.62}      & 49.95               & 106.11      & \textbf{13.41}         \\
\sVI    & 32.45                 & 587.11      & \ph{0}\underline{5.25} & 32.45             & 587.11      & \ph{0}\underline{5.25} & 30.82               & 589.30      & \ph{0}\textbf{3.91}    \\
\bottomrule
\multicolumn{10}{l}{\footnotesize
    \addstackgap{
        \begin{minipage}{0.8\linewidth}
        \textbf{Bold} errors are the lowest for each scenario. \\
        \underline{Underlined} errors are the second lowest for each scenario.
      \end{minipage}
}}
\end{tabularx}
\label{tab:maxPosErr}
\end{table}

\begin{table}[hptb]
\centering
\caption{Average rotation errors (degrees) over each run. Errors are between Kimera's estimate of the inspector's relative pose and the ground truth relative pose.}
\small
\begin{tabularx}{\linewidth}{l !{\colSpace} *{3}{C} !{\colSpace} *{3}{C} !{\colSpace} *{3}{C}}
\toprule
& \multicolumn{3}{c}{VIO} & \multicolumn{3}{c}{\shortstack{VIO +\\ Bag of Words LCs}} & \multicolumn{3}{c}{\shortstack{VIO +\\ TIR Silhouette-Based LCs}} \\
\cmidrule(lr){2-4} \cmidrule(lr){5-7} \cmidrule(lr){8-10}
Scenario &
VIS & TIR & Fused & VIS & TIR & Fused & VIS & TIR & Fused \\
\cmidrule(r){1-1} \cmidrule(lr){2-4} \cmidrule(lr){5-7} \cmidrule(lr){8-10}
\sI     & 22.55                 & 89.50 & \ph{0}3.36 & \ph{0}8.51 & 63.24 & \ph{0}\underline{2.59} & \ph{0}3.81            & 15.92      & \ph{0}\textbf{2.03}    \\
\sII    & \ph{0}\underline{2.29} & 89.50 & \ph{0}5.05 & \ph{0}2.86 & 63.24 & \ph{0}2.81             & \ph{0}\textbf{1.91}   & 15.92      & \ph{0}2.46             \\
\sIII   & 21.08                 & 60.72 & 16.86      & 21.08      & 64.13 & 16.69                  & \underline{11.16}     & 37.25      & \textbf{10.32}         \\
\sIV    & 20.89                 & 72.75 & 11.00      & 12.55      & 37.61 & \ph{0}\underline{4.73} & \ph{0}7.62            & 13.79      & \ph{0}\textbf{3.52}    \\
\sV     & 17.74                 & 21.65 & 15.08      & 12.23      & 22.19 & \ph{0}\underline{4.92} & \ph{0}5.08            & \ph{0}4.99 & \ph{0}\textbf{4.32}    \\
\sVI    & 17.51                 & 50.80 & 12.38      & 17.51      & 50.80 & 12.38                  & \ph{0}\underline{9.66} & 30.00     & \ph{0}\textbf{7.61}    \\
\bottomrule
\multicolumn{10}{l}{\footnotesize
    \addstackgap{
        \begin{minipage}{0.8\linewidth}
        \textbf{Bold} errors are the lowest for each scenario. \\
        \underline{Underlined} errors are the second lowest for each scenario.
      \end{minipage}
}}
\end{tabularx}
\label{tab:avgRotErr}
\end{table}

\begin{table}[hptb]
\centering
\caption{Maximum rotation errors (degrees) over each run. Errors are between Kimera's estimate of the inspector's relative pose and the ground truth relative pose.}
\small
\begin{tabularx}{\linewidth}{l !{\colSpace} *{3}{C} !{\colSpace} *{3}{C} !{\colSpace} *{3}{C}}
\toprule
& \multicolumn{3}{c}{VIO} & \multicolumn{3}{c}{\shortstack{VIO +\\ Bag of Words LCs}} & \multicolumn{3}{c}{\shortstack{VIO +\\ TIR Silhouette-Based LCs}} \\
\cmidrule(lr){2-4} \cmidrule(lr){5-7} \cmidrule(lr){8-10}
Scenario &
VIS & TIR & Fused & VIS & TIR & Fused & VIS & TIR & Fused \\
\cmidrule(r){1-1} \cmidrule(lr){2-4} \cmidrule(lr){5-7} \cmidrule(lr){8-10}
\sI     & 39.39               & 180.00      & \ph{0}7.84        & 19.78 & 131.17      & \ph{0}\underline{6.01} & 10.23                 & 66.82 & \ph{0}\textbf{3.89} \\
\sII    & \ph{0}\textbf{3.50} & 180.00      & \ph{0}9.18        & 15.54 & 131.17      & \ph{0}6.55             & \ph{0}\underline{4.34} & 66.82 & \ph{0}7.04         \\
\sIII   & 45.73               & 129.22      & \underline{30.11} & 45.73 & 147.10      & 49.87                  & 31.95                 & 80.29 & \textbf{27.27}      \\
\sIV    & 51.22               & 171.03      & 26.79             & 39.38 & \ph{0}84.98 & \underline{10.80}      & 16.17                 & 26.87 & \ph{0}\textbf{8.32} \\
\sV     & 48.14               & \ph{0}36.59 & 44.65             & 35.34 & \ph{0}45.30 & 12.19                  & \underline{10.55}     & 18.88 & \ph{0}\textbf{8.09} \\
\sVI    & 26.45               & 109.06      & 23.71             & 26.45 & 109.06      & 23.71                  & \underline{22.54}     & 55.72 & \textbf{16.42}      \\
\bottomrule
\multicolumn{10}{l}{\footnotesize
    \addstackgap{
        \begin{minipage}{0.8\linewidth}
        \textbf{Bold} errors are the lowest for each scenario. \\
        \underline{Underlined} errors are the second lowest for each scenario.
      \end{minipage}
}}
\end{tabularx}
\label{tab:maxRotErr}
\end{table}

%% file: 7_conclusion.tex
\section{Conclusion} \label{sec:conclusion}
We present a relative navigation method for an inspector spacecraft operating around an uncooperative, unknown resident space object (RSO) in LEO, based on image-level fusion of visible (VIS) and thermal-infrared (TIR) imagery. Unlike approaches that keep the two modalities separate until a final fusion step, we fuse at the first stage of the pipeline, so the high resolution, contrast, and feature richness of the visible camera and the illumination-robustness of the TIR camera are jointly exploited during feature tracking and pose estimation.

A high-fidelity Blender simulation of a target spacecraft in both bands—including internal heating, solar heating, Earth albedo, and Earth IR—generates six representative inspection scenarios with distinct relative trajectories, lighting conditions, and target spacecraft. Images are fused and passed to a SLAM algorithm together with noisy relative acceleration and angular-velocity measurements serving as orbital dynamics priors. Averaged over the six scenarios, the use of fused imagery reduces position error in the visual-inertial odometry by 56\% versus visible-only and 92\% versus thermal-only. Rotation errors are reduced by 38\% and 83\%, respectively.

We also introduce a training-free, silhouette-based visual place recognition method, motivated by the failure of both classical and learning-based VPR in the space domain. It extracts the target silhouette from TIR imagery, forms a compact descriptor from radial boundary distances, and matches candidates using locality-sensitive hashing. It outperforms all baseline VPR methods in precision–recall at a faster CPU-only speed. Comparing fused imagery with only VIO to fused imagery with silhouette-based loop closures, average position errors are reduced by an average of 35\% and average rotation errors are reduced by 53\%.

Our multispectral navigation approach exploits the unique information from each imager in every illumination scenario. Thermal imagery clearly has practical uses for improving both front-end tracking and place recognition. Being fully passive, this approach avoids the size, weight, and power burden of lidar while retaining robust tracking, providing a path toward small, low-cost spacecraft to perform complex proximity operations missions such as on-orbit servicing, active debris removal, and asteroid exploration.